\documentclass[10pt,twocolumn,letterpaper]{article}

\usepackage{iccv}
\usepackage{times}
\usepackage{epsfig}
\usepackage{graphicx}
\usepackage{amsmath}
\usepackage{amssymb}
\usepackage{multirow}
\usepackage{makecell}
\usepackage[inline]{enumitem}
\usepackage{booktabs}
\usepackage{xfrac}  
\usepackage{flushend}
\usepackage{subcaption}

\DeclareMathOperator*{\argmin}{arg\,min}

\iccvfinalcopy % *** Uncomment this line for the final submission

\def\iccvPaperID{12} % *** Enter the ICCV Paper ID here

\ificcvfinal\fi
\begin{document}

%%%%%%%%% TITLE
\title{Learning Invariant Riemannian Geometric Representations Using Deep Nets}

\author{Suhas Lohit\\
Arizona State University\\
{\tt\small slohit@asu.edu}
% For a paper whose authors are all at the same institution,
% omit the following lines up until the closing ``}''.
% Additional authors and addresses can be added with ``\and'',
% just like the second author.
% To save space, use either the email address or home page, not both
\and
Pavan Turaga\\
Arizona State University\\
{\tt\small pturaga@asu.edu}
}

\maketitle
%\thispagestyle{empty}

%%%%%%%%% ABSTRACT
\begin{abstract}
Non-Euclidean constraints are inherent in many kinds of data in computer vision and machine learning, typically as a result of specific invariance requirements that need to be respected during high-level inference. Often, these geometric constraints can be expressed in the language of Riemannian geometry, where conventional vector space machine learning does not apply directly. The central question this paper deals with is: {\em How does one train deep neural nets whose final outputs are elements on a Riemannian manifold?} To answer this, we propose a general framework for manifold-aware training of deep neural networks -- we utilize tangent spaces and exponential maps in order to convert the proposed problem into a form that allows us to bring current advances in deep learning to bear upon this problem. We describe two specific applications to demonstrate this approach: prediction of probability distributions for multi-class image classification, and prediction of illumination-invariant subspaces from a single face-image via regression on the Grassmannian. These applications show the generality of the proposed framework, and result in improved performance over baselines that ignore the geometry of the output space. In addition to solving this specific problem, we believe this paper opens new lines of enquiry centered on the implications of Riemannian geometry on deep architectures.
\end{abstract}

%%%%%%%%% BODY TEXT
\section{Introduction}
\label{sec:intro}

Many applications in computer vision employ data that are naturally represented on manifolds \cite{lui2012advances}. Shapes that are invariant to affine transforms \cite{begelfor2006affine} and linear dynamical systems \cite{turaga2011statistical} can be represented as points on the Grassmannian. In diffusion tensor imaging, each "pixel" of the "image" is a symmetric positive definite (SPD) matrix and the space of SPD matrices forms a manifold \cite{pennec2006riemannian}.  Lie groups like $SO(3)$ and $SE(3)$ are used to represent human skeletons \cite{vemulapalli2014human,vemulapalli2016rolling}. Predicting probability density functions is another area of interest, applicable to multi-class classification and bag-of-words models \cite{Lafferty2005}, and saliency prediction \cite{Jetley2016}.

Several years of research has presented us with various tools for statistics, and thereby machine learning approaches to be deployed when the objects of interest have manifold-valued domains (c.f. \cite{RCCV}). In deep learning, it is usually the case that data samples are viewed as elements of vector spaces. Any additional structure that the data may possess is left to be learned through the training examples. However, recently, there has been interest in employing deep learning techniques for non-Euclidean inputs as well: \cite{bronstein2016geometric} including graph-structured data \cite{bruna2014spectral, henaff2015deep, niepert2016learning} and 3D shapes viewed as Riemannian manifolds \cite{masci2015geodesic}. Also, deep networks that preserve the input geometry at each layer have been studied for inference problems, e.g., for symmetric positive definite matrices \cite{huang2017riemannian}, Lie groups \cite{huang2017deep} and points on the Stiefel manifold \cite{huang2016building}. Another recent work considers weight matrices which are constrained to be orthogonal, i.e., points of the Stiefel manifold, and propose a generalized version of backpropagation \cite{harandi2016generalized}. These works do not consider output variables with geometric constraints.

In contrast to the above, instead of enforcing geometry at the inputs, our goal is to design a general framework to extend neural network architectures where output variables (or deeper feature maps) lie on manifolds of known geometry, typically due to certain invariance requirements. We do not assume the inputs themselves have known geometric structure and employ standard back-propagation for training. Equivalently, one may consider this approach as trying to estimate a mapping from an input $\mathbf{x} \in \mathbb{R}^N$ to a manifold-valued point $\mathbf{m} \in \mathcal{M}$ i.e., $f: \mathbb{R}^N \to \mathcal{M}$, using a neural network, where $\mathbf{m}$ is the desired output. 
%and if required can be further composed with standard neural network layers or a classifier that can operate on manifold-valued representations.

That is, this paper provides a framework for regression that is applicable to predicting manifold-valued data and at the same time is able to leverage the power of neural nets for feature learning, using standard backpropagation for unconstrained optimization. In this paper, we focus on two manifolds that are of wide interest in computer vision -- the hypersphere and the Grassmannian. We describe the applications next.

\vspace{-0.1in}
\paragraph{Face $\to$ Illumination Subspace as regression on the Grassmannian:} The \textit{illumination subspace} of a human face is a popular example from computer vision where for a particular subject, the set of all face images of that subject under all illumination conditions can be shown to lie close to a low dimensional subspace \cite{hallinan1994low}. These illumination subspaces are represented as points on the Grassmannian (or Stiefel, depending on application) manifold. Several applications such as robust face recognition have been proposed using this approach. In this work, in order to demonstrate how deep networks can be employed to map to Grassmannian-valued data, we consider the problem of estimating the illumination subspace from a single input image of a subject under unknown illumination. We refer to this application as Face$\to$Illumination Subspace (F2IS).

\vspace{-0.1in}
\paragraph{Multi-class classification as regression on the unit hypersphere:} Classification problems in deep learning use the softmax layer to map arbitrary vectors to the space of probability distributions. However, more formally, probability distributions can be easily mapped to the unit hypersphere, under a square-root parametrization \cite{srivastava2007riemannian} inspired by the Fisher-Rao metric used in information geometry. Thus, multi-class classfication can be posed as regression to a hypersphere. Indeed, there has been work recently that consider \textit{spherical-loss functions} which use the Euclidean loss on unit-norm output vectors of a network \cite{vincent2015efficient,de2016exploration}. In this work, we propose loss-functions for the classification problem based on the geometry of the sphere.

\vspace{-0.1in}
\paragraph{Main contributions:} In this paper, we address the training of neural networks using standard backpropagation to output elements that lie on Riemannian manifolds. To this end, we propose two frameworks
in this paper: \begin{enumerate}[label=(\arabic*)]  
\item We discuss how to map to simpler manifolds like the hypersphere directly using a combination of geodesic loss functions as well as differentiable constraint satisfaction layers such as the normalization layer in the case of the hypersphere.
\item We also propose a more general framework that is applicable to Riemannian manifolds that may not have closed-form expressions for the geodesic distance or when the constraints are hard to encode as a layer in the neural network. In this framework, the network maps to the tangent space of the manifold and then the exponential map is employed to find the desired point on the manifold. 
\end{enumerate}
We carry out experiments for the applications described above in order to evaluate the proposed frameworks and show that geometry-aware frameworks result in improved performance compared to baselines that do not take output geometry into account. 

\section{Related work}
We will now point to some related work that also examine the problem of predicting outputs with geometric structure using neural networks. Byravan and Fox \cite{byravan2016se3} and Clark et al. \cite{clark2017vinet} design deep networks to output $SE(3)$ transformations. The set of transformations $SE(3)$ is a group which also possesses manifold structure, i.e., a Lie group. It is not straightforward to predict elements on $SE(3)$ since it involves predicting a matrix constrained to be orthogonal. Instead, the authors map to the Lie algebra $se(3)$ which is a linear space. We note that the Lie algebra is nothing but the tangent space of $SE(3)$ at the identity transformation and can be considered a particular case of the general formulation presented in this paper.  Huang et al. \cite{huang2017deep} use the logarithm map to map \textit{feature maps} on $SE(3)$ to $se(3)$ before using regular layers for action recognition. However, the logarithm map is implemented within the network, since for $SE(3)$, this function is simple and differentiable. In contrast, in this work, we require the network \textit{output} to be manifold-valued and thus do not impose any geometry requirements at the input or for the feature maps. This also means that a suitable loss function needs to be defined, taking into account, the structure of the manifold of interest.  

In a more traditional learning setting, there has been work using \textit{geodesic regression}, a generalization of linear regression, on Riemannian manifolds \cite{fletcher2011geodesic,fletcher2013geodesic,shi2009intrinsic, hong2014geodesic, kim2014multivariate}, where a geodesic curve is computed such that the average distance (on the manifold) from the data points to the curve is minimized. This involves computing gradients on the manifold. Recent work has also included non-linear regression on Riemannian manifolds \cite{banerjee2015nonlinear, Banerjee_2016_CVPR}. Here, the non-linearity is provided by a pre-defined kernel function and the mapping algorithm solves an optimization problem iteratively. Our work is a non-iterative deep-learning based approach to the problem described in Banerjee et al. \cite{banerjee2015nonlinear} as regression with the independent variable in $\mathbb{R}^N$ and the dependent variable lying on a manifold $\mathcal{M}$. That is, unlike these works, the mapping $f: \mathbb{R}^N \to \mathcal{M}$ in our case is a hierarchical non-linear function, learned directly from data without any hand-crafted feature extraction, and the required mapping is achieved by a simple feed forward pass through the trained network.  

All neural nets in the paper are trained and tested using Tensorflow \cite{tensorflow2015-whitepaper}, making use of its automatic differentiation capability. Before we discuss the contributions of this work, if required, please refer supplementary material for definitions of some important terms from differential geometry. 

%$$$$$$$$$$$$$$$$$$$$$$$$$$$$$$$$$$$$$$$$$$$$$$$$$$$$$$$$$$$$$$$$$$$$$$$$$$$$$$$$$$$$$$$$$$$$
\begin{figure}
\centering
    \includegraphics[trim = {0cm, 0cm, 0cm, .25cm}, clip, width=0.35\textwidth]{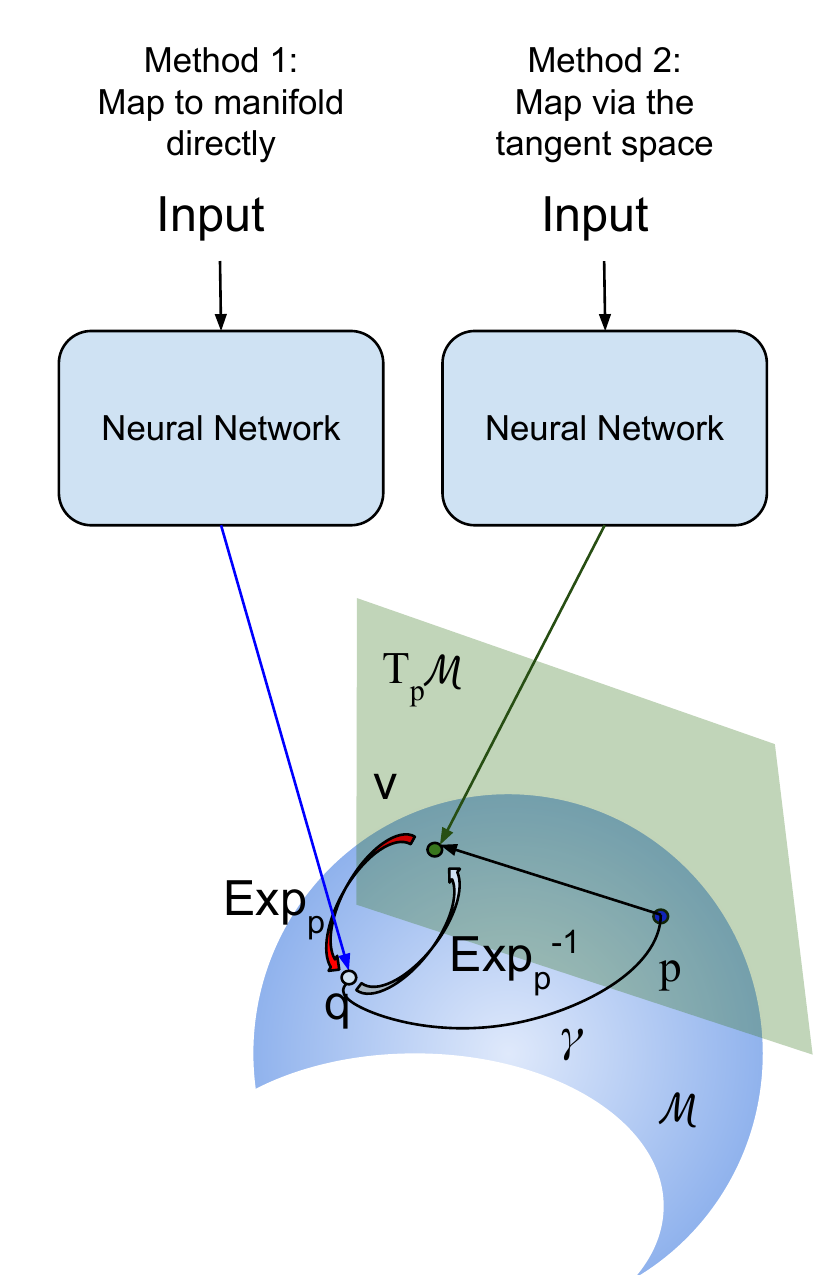}
    \caption{\small{This figure illustrates the two approaches presented in this paper for training neural networks to predict manifold-valued data. It also explains some basic concepts from differential geometry visually. $\mathcal{M}$ is a manifold, $T_P\mathcal{M}$ is the tangent space at $p\in\mathcal{M}$. $p$ is called the pole of the tangent space. The curve connected $p$ and $q\in\mathcal{M}$ is the geodesic curve $\gamma$. $v$ is a point on $T_p\mathcal{M}$ such that the exponential map $\exp_p(v) = \gamma(1)$ and the logarithm map $\exp_p^{-1}(q) = v$.}}
\label{fig:manifold}
\end{figure}

\section{Two approaches for deep manifold-aware prediction}
We propose two ways of predicting manifold-valued data using neural networks with standard backpropagation. See Figure \ref{fig:manifold} for visual illustration and important notation.

\paragraph{Mapping to the manifold via geodesic-loss functions:} In this case, the network directly maps input vectors to elements on the manifold $\mathcal{M}$ and is required to learn the manifold constraints from the data. If we represent the neural network as a mapping \texttt{NN}, we have \texttt{NN}$: \mathbb{R}^N \to \mathcal{M}$. Firstly, unlike simple manifolds like the sphere, manifolds in general do not have a differentiable closed-form expression, that are also efficiently computable, for the geodesic distance function that can be used as a loss function for the neural network. Although one can still resort to using a differentiable loss function such as the Euclidean distance, this approach is not mathematically correct and does not yield the right estimate for distance on the manifold. Secondly, the network output has to satisfy the manifold constraints. In the case of the sphere, it is simple to enforce the unit-norm constraint at the output layer of the neural network using a differentiable normalization layer. It is however less clear how to map to more complicated manifolds such as the Stiefel and Grassmann manifolds where the points are usually represented by tall-thin orthonormal matrices. That is, in addition to the unit-norm constraints, orthogonality constraints between all pairs of columns in the matrix need to be enforced. The Grassmann manifold, presents a more difficult challenge, since each point in this space is an equivalence class of points on the Stiefel manifold that are orthogonal transforms of each other. As we will see later, the data representation that respects this equivalence (projection matrix) does not admit a feasible way for a neural network to map to this manifold directly. 

\paragraph{Mapping via the tangent space -- toward a general framework:} This is a more general formulation that is applicable to all the manifolds of interest. Here, the network first maps to a vector on the tangent space constructed at a suitable pole $p \in \mathcal{M}$, which forms the intermediate output. Once the network outputs the required tangent, the exponential map ($\exp_p$) is employed to find the corresponding point on the manifold. Mathematically, we decompose the desired function $f: \mathbb{R}^N \to \mathcal{M}$ as $f = \exp_p\circ\texttt{ NN}$ and  \texttt{NN}$: \mathbb{R}^N \to T_p\mathcal{M}$. Intuitively, since the tangent space is a vector space that encodes geometric constraints implicitly, it is attractive here, as neural networks have been shown to be effective for estimating vector-valued data.  We note that an assumption is implicit in this framework: all the data points of interest on the manifold are much closer to $p$ than the cut-locus of the manifold and in this case, the distance on the tangent space serves as a good approximation to the geodesic distance. This is the same assumption that goes into currently successful approaches for statistical computing methods on manifolds \cite{pennec2006riemannian}. In practice, we find this assumption is respected in our applications as well.
    
%This framework offers different advantages specific to the manifold. For the sphere and the Grassmannian, we use as the Riemannian metric the Euclidean metric inherited on the tangent-space, hence the Euclidean distance is appropriate on the tangent-space. For the Stiefel manifold, the distance between two points is not available in closed form. However, since we know the Riemannian metric, we can construct a closed-form expression for the loss function on the tangent-space. In the case of Grassmann manifold, the algebraic representation that respects the geometry of the manifold, has a very large dimensionality and makes direct mapping to the Grassmann infeasible, whereas the tangent space representation has a much smaller intrinsic dimensionality. In the rest of the paper, we implement these frameworks for two specific manifolds -- the sphere and Grassmann.
    
%$$$$$$$$$$$$$$$$$$$$$$$$$$$$$$$$$$$$$$$$$$$$$$$$$$$$$$$$$$$$$$$$$$$$$

%$$$$$$$$$$$$$$$$$$$$$$$$$$$$$$$$$$$$$$$$$$$$$$$$$$$$$$$$$$$$$$$$$$$$$

\section{Deep regression on the Grassmannian for F2IS}
\label{sec:grassmann}

\paragraph{Face $\to$ Illumination Subspace:} 
We will now describe an ill-posed inverse problem from computer vision that serves as our canonical application to illustrate prediction on the Grassmann manifold using a neural network. It is well known that the set of images of a human face in frontal pose under all illuminations lies close to a low-dimensional subspace, known as the illumination subspace \cite{hallinan1994low,epstein1995}. If we compute the eigenvectors of this set of images for different subjects using PCA, we observe that the top 5 principal components (PCs) capture nearly 90\% of the variance. More importantly, for this paper, an obvious pattern can be observed between the subject under consideration and the PCs of the illumination subspace. Firstly, the identity of the subject can be easily determined from the PCs. Secondly, as noted by Hallinan \cite{hallinan1994low}, we observe that the illumination patterns of top 5 principal components are the same across subjects only up to certain permutations and sign flips. \footnote{It is clear that for an eigenvector $\mathbf{e}$, $-\mathbf{e}$ is also an eigenvector.} Using the terminology in \cite{hallinan1994low}, we can interpret the visualizations of the top 5 PC's as a face under the following respective illuminations: \textit{frontal lighting, side lighting, lighting from above/below, extreme side lighting and lighting from a corner}. The $1^{st}$ and $2^{nd}$ PC's have eigenvalues in a similar range and sometimes exchange places depending on the subject. The $3^{rd}$ PC, corresponding to  eigenvalue is always at the same place. The $4^{th}$ and $5^{th}$ PCs have eigenvalues in a similar range and can interchange places for a few subjects.

The illumination subspace refers to the linear {\em span} of these eigenvectors, and is a point on the Grassmannian. \textbf{When we represent the subspace by its projection matrix representation, the representation becomes invariant to both sign flips and permutations} (in fact, invariant to the full set of right orthogonal transforms).

In this paper, as an example of predicting points on Grassmann manifold, we define the following ill-posed inverse problem: given a human face in frontal pose under an unknown illumination, output the corresponding illumination subspace. We will refer to this problem as the "Face $\to$ Illumination Subspace" problem or F2IS. In our experiments, we consider the illumination subspace to be of dimension $d = 3,4$,or $5$.

\paragraph{Geometry of the Grassmannian:} The Grassmann manifold, denoted by $\mathcal{G}_{n,d}$, is a matrix manifold and is the set of $d$-dimensional subspaces in $\mathbb{R}^n$. To represent a point on $\mathcal{G}_{n,d}$, we can use an orthonormal matrix, $\mathbf{U} \in \mathbb{R}^{n \times d}$ ($\mathbf{U}^T\mathbf{U} = \mathbf{I}_n$), to represent the equivalence class of points in $\mathbb{R}^{n \times d}$, such that, two points are equivalent if their columns span the same $d$-dimensional subspace. That is, $\mathcal{G}_{n,d} = \{[\mathbf{U}]\}$, where $[\mathbf{U}] = \{\mathbf{UQ} | \mathbf{U}^T\mathbf{U} = \mathbf{I}, \mathbf{Q} \text{ is orthogonal}\}$. In order to uniquely represent the equivalence class $[\mathbf{U}] \in \mathcal{G}_{n,d}$, we use its projection matrix representation $\mathbf{P} = \mathbf{UU}^T \in \mathbb{R}^{n \times n}$, where $\mathbf{U}$ is some point in the equivalence class. $\mathbf{UU}^T$ contains $\frac{n(n+1)}{2}$ unique entries as it is a symmetric matrix. Clearly, for any other point in the same equivalence class $\mathbf{UQ}$, its projection matrix representation is $(\mathbf{UQ})(\mathbf{UQ})^T = \mathbf{UU}^T$, as required. \textbf{Thus, the space of all rank $d$ projection matrices of size $n \times n$, $\mathcal{P}_n$ is diffeomorphic to $\mathcal{G}_{n,d}$.} The identity element of $\mathcal{P}_n$ is given by $\mathbf{I}_{\mathcal{P}_n} = diag(\mathbf{I}_d, \mathbf{0}_{n-d})$, where $\mathbf{0}_{n-d}$ is the matrix of zeros of size $(n-d) \times (n-d)$. In order to find the exponential and logarithm maps for $\mathcal{G}_{n,d}$, we will view $\mathcal{G}_{n,d}$ as a quotient space of the orthogonal group, $\mathcal{G}_{n,d} = \mathcal{O}_n/(\mathcal{O}_{n-d}\times\mathcal{O}_d)$. The Riemannian metric in this case is the standard inner product \cite{edelman1998geometry} and thus, the distance function induced on the tangent space is the Euclidean distance function. Using this formulation, given any point $\mathbf{P} = \mathbf{UU}^T \in \mathcal{P}_n$, a geodesic of $\mathcal{P}_n$ at $\mathbf{I}_{\mathcal{P}_n}$ passing through $\mathbf{P}$ at $t = 0$, is a particular geodesic $\alpha(t)$ of $\mathcal{O}(n)$ completely specified by a skew-symmetric $\mathbf{X} \in \mathbb{R}^{n \times n}$: $\alpha(t) = \exp_m(t\mathbf{X})I_P\exp_m(-t\mathbf{X})$, where $\exp_m(.)$ is the matrix exponential and $\mathbf{P} = \alpha(1)$, such that $\mathbf{X}$ belongs to the set $M$ given by $M = \bigg\{\begin{bmatrix} \mathbf{0}_d & \mathbf{A} \\ -\mathbf{A}^T & \mathbf{0}_{n-d} \end{bmatrix}| \quad \mathbf{A} \in \mathbb{R}^{d \times (n-d)} \bigg\}$. $\mathbf{X}$ serves as the tangent vector to $\mathcal{G}_{n,d}$ at the identity and is completely determined by $\mathbf{A}$. The geodesic between two points $\mathbf{P}_1,\mathbf{P}_2 \in \mathcal{P}_n$, is computed by rotating $\mathbf{P}_1$ and $\mathbf{P}_2$ to $\mathbf{I}_{\mathcal{P}_n}$ and some $\mathbf{P} \in \mathcal{P}_n$ respectively. The exponential map, takes as inputs, the pole and the tangent vector and returns the subspace $span(\mathbf{U})$, represented by some point $\mathbf{UQ}$. Refer Srivastava and Klassen \cite{srivastava2004bayesian} and Taheri et al. \cite{taheri2011towards} for algorithms to compute exponential and log maps for the Grassmannian.

\paragraph{Synthetic dataset for F2IS:} \label{sec:f2is_dataset}
We use the Basel Face Model dataset \cite{paysan20093d} in order to generate 250 random 3D models of human faces $\{S_i\}, i = 1\dots250$ (200 for training and 50 for testing chosen randomly).\footnote{We use a synthetic dataset because we were unable to find any large publicly available real database that would enable training of neural networks without overfitting.} We then generate a set of 64 faces for each subject where each face is obtained by varying the direction of the point source illumination for the frontal pose i.e., $S_i = \{\mathbf{F}_i^{j}\}, j = 1\dots64$. The directions of illumination are the same as the ones used in the Extended Yale Face Database B \cite{georghiades2001few}. Each face image is converted to grayscale and resized to $28 \times 28$. Once we have the 250 sets of faces under the 64 illumination conditions, we calculate the illumination subspace for each subject as follows. For each subject, we first subtract the mean face image of that subject under all illumination conditions and then calculate the principal components (PCs). For a subject $i$ and an illumination condition $j$, we will denote the input face image by $\mathbf{F}_i^j$ and the desired $d$ top PCs by $\mathbf{E}_i^1, \mathbf{E}_i^2, \dots, \mathbf{E}_i^d$ (note that the PCs do not depend on the input illumination condition). It is clear that every $\mathbf{E}_i^k, k = 1,2,\dots,d$ is of size $28 \times 28$ and $\langle \mathbf{E}_i^k, \mathbf{E}_i^l \rangle = 1$, if $k=l$ and 0 otherwise.

If we lexicographically order each $\mathbf{E}_i^k$ to form a vector $vec(\mathbf{E}_i^k)$ of size $784 \times 1$ and for each subject, arrange the $\mathbf{E}_i^k$'s to form a matrix $\mathbf{U}_i = [vec(\mathbf{E}_i^1) \text{ } vec(\mathbf{E}_i^2) \dots vec(\mathbf{E}_i^d)]$, then the orthonormality constraint can be rewritten as $\mathbf{U}_i^T\mathbf{U}_i = \mathbf{I}_d$, where $\mathbf{I}_d$ is the identity matrix of size $d \times d$. As we argued earlier, due to the nature of the problem, $\mathbf{U}_i$ should be represented as a point on the Grassmann $\mathcal{G}_{784,d}$ using the projection matrix representation. With this notation, the desired mapping is $f: \mathbb{R}^{28 \times 28} \to \mathcal{G}_{784,d}$ such that $f(\mathbf{F}_i^j) = \mathbf{U}_i\mathbf{Q} \in [\mathbf{U}_i]$, the required equivalence class or equivalently, $\mathbf{U}_i\mathbf{U}_i^T$.

For the inputs $\mathbf{F}_i^j$ during training and testing, we do not use all the illumination directions ($j's$). We only use illumination directions that light at least half of the face. This is because for extreme illumination directions, most of the image is unlit and does not contain information about the identity of the subject, which is an important factor for determining the output subspaces. We select the same $33$ illumination directions for all subjects to form the inputs for the network. We randomly split the dataset into 200 subjects for training and 50 subjects for testing. Therefore there are $33 \times 200 = 6600$ and $33 \times 50 = 1650$ different input-output pairs for training and testing respectively. The 33 illumination directions used for creating inputs for both the training and test sets are given in the supplementary and are a subset of the illumination directions used in the Extended Yale Face Database B \cite{georghiades2001few}.

\subsection{Proposed frameworks for solving F2IS} \label{sec:frameworks_f2is}
We propose two frameworks which employ networks with nearly the same architecture: The network consists of 3 \texttt{conv} layers and two \texttt{fc} layers. ReLU non-linearity is employed. Each \texttt{conv} layer produces 16 feature maps. All the filters in the \texttt{conv} layers are of size $11 \times 11$. The first \texttt{fc} layer produces a vector of size 512. Size of the second \texttt{fc} layer depends on the framework. Both networks are trained using the Adam optimizer \cite{kingma2015adam} using a learning rate of $10^{-3}$ for 50000 iterations with a mini-batch size of 30. Euclidean loss between the desired output and ground truth is employed in both cases. We show that the choice of representation of the desired output is crucial in this application. We carry out three sets of experiments using subspace dimension $d = 3,4$ and $5$.

\vspace{-0.07in}
\paragraph{Baseline:} The first framework is a baseline that attempts to directly map to the desired PCs represented as a matrix $\mathbf{U}_i$, given $\mathbf{F}_i^j$, i.e., \texttt{NN}$(\mathbf{F}_i^j) = \mathbf{U}_i$. We use the Euclidean loss function between the ground-truth $\mathbf{U}_i$ and the network output $\hat{\mathbf{U}}_i$ for training: $L_b = ||\mathbf{U}_i - \hat{\mathbf{U}}_i||^2_F$ That is, instead of regressing to the desired subspace, the network attempts to map to its basis vectors (PCs). However, the mapping from $\mathbf{F}_i^j$ to $\mathbf{U}_i$ is consistent across subjects \textbf{only up to certain permutations and sign flips in the PCs}. Hence, without correcting these inconsistencies ad hoc, the problem is rendered too complicated, since during the training phase, the network receives conflicting ground-truth vectors depending on the subject. Thus, this framework performs poorly as expected. It is important to note that mapping to the correct representation $\mathbf{UU}^T$ (which respects Grassmann geometry and is invariant to these inconsistencies) directly is not feasible because the size of $\mathbf{UU}^T$ is too large ($\frac{784 \times 785}{2}$) and has rank constraints. This necessitates mapping via the tangent space, discussed next. 

\vspace{-0.07in}
\paragraph{Mapping via the Grassmann tangent space -- GrassmannNet-TS:}\label{sec:regress_grassmann_tangent}
The second framework represents the output subspaces as points on the Grassmann manifold and first maps to the Grassmann tangent space and then computes the required subspace using the Grassmann exponential map. This circumvents the problem of very large dimensionality encountered in the first approach since the tangent vector has a much smaller intrinsic dimensionality. This representation has a one-to-one mapping with the projection matrix representation and thus, is naturally invariant to the permutations of the PCs and all combinations of sign flips present in the data. And the mapping we intend to learn becomes feasible in a data-driven framework. Mathematically, \texttt{NN}$: \mathbb{R}^{784 \times d} \to T_{p}\mathcal{G}_{784, d}$.

\begin{table*}[]
    \centering
    \begin{tabular}{cccc}
     
        \hline
        Input & Ground-truth PCs & Output of baseline n/w & Output of GrassmannNet-TS \\
        \hline
       \includegraphics[trim = {18cm, 13cm, 18cm, 6cm}, clip, width=0.11\textwidth]{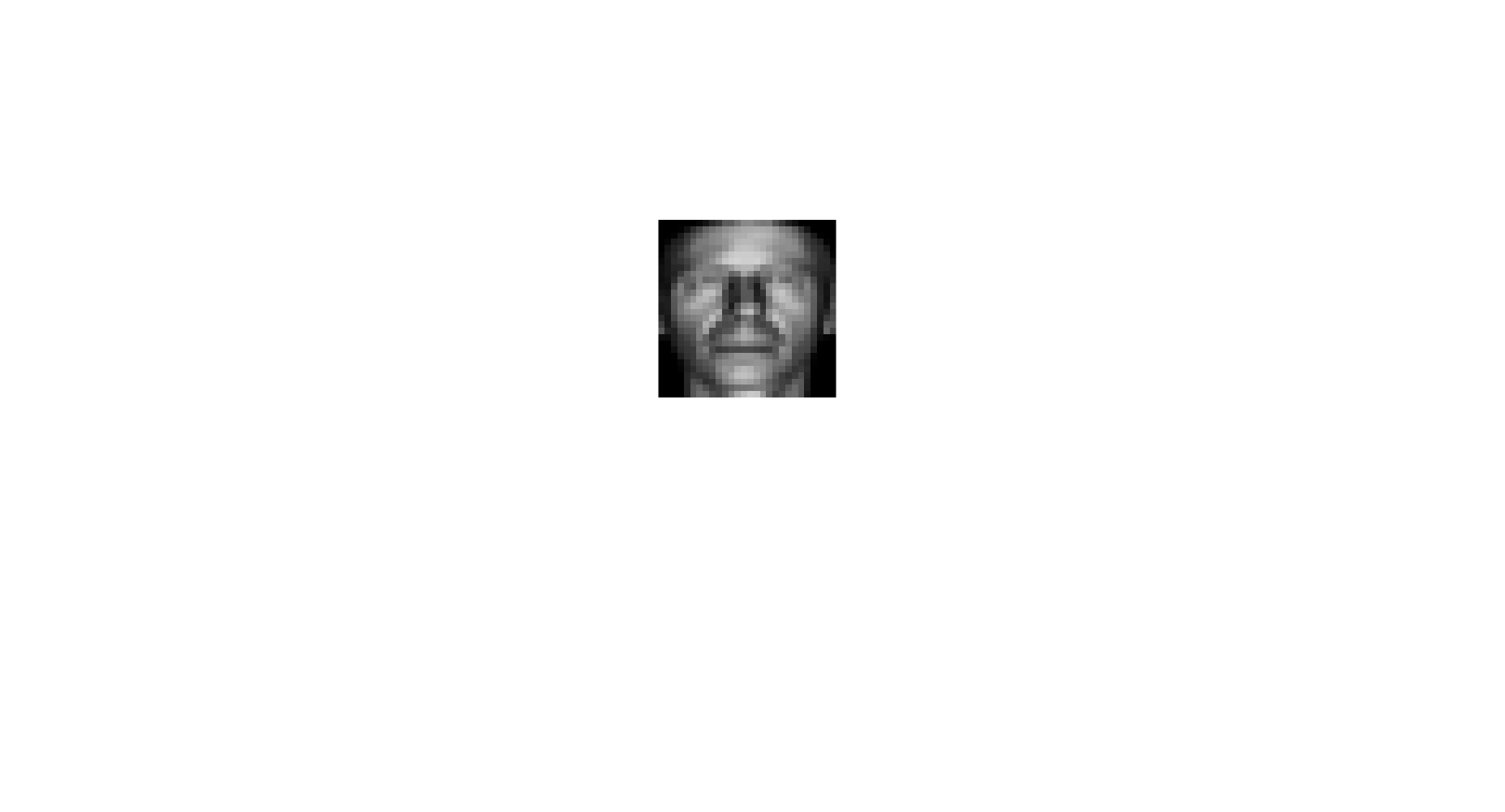} 
         & \includegraphics[trim = {14cm, 14cm, 14cm, 6cm}, clip, width=0.22\textwidth]{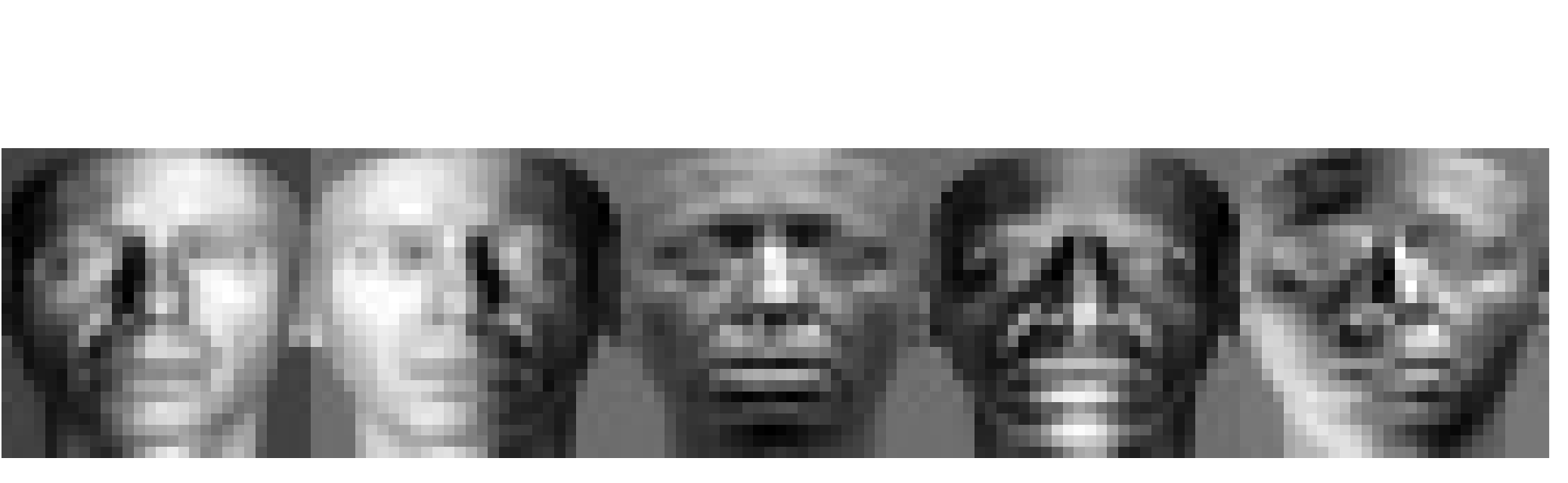} 
         & \includegraphics[trim = {14cm, 14cm, 14cm, 8cm}, clip, width=0.22\textwidth]{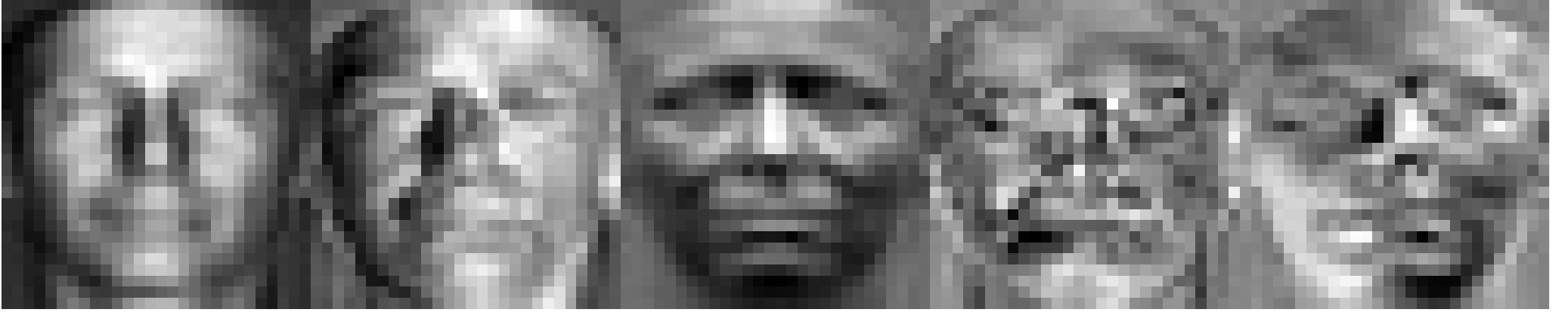}
         & \includegraphics[trim = {14cm, 14cm, 14cm, 8cm}, clip, width=0.22\textwidth]{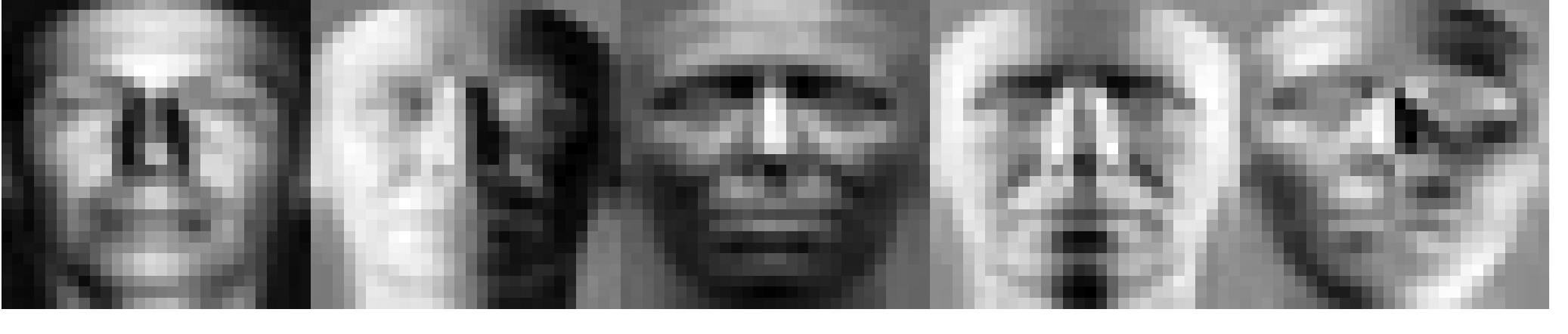}\\
         
         & & $D_G = 1.6694$ & $D_G = 0.7006$ \\
         \midrule
         
        \includegraphics[trim = {18cm, 13cm, 18cm, 6cm}, clip, width=0.11\textwidth]{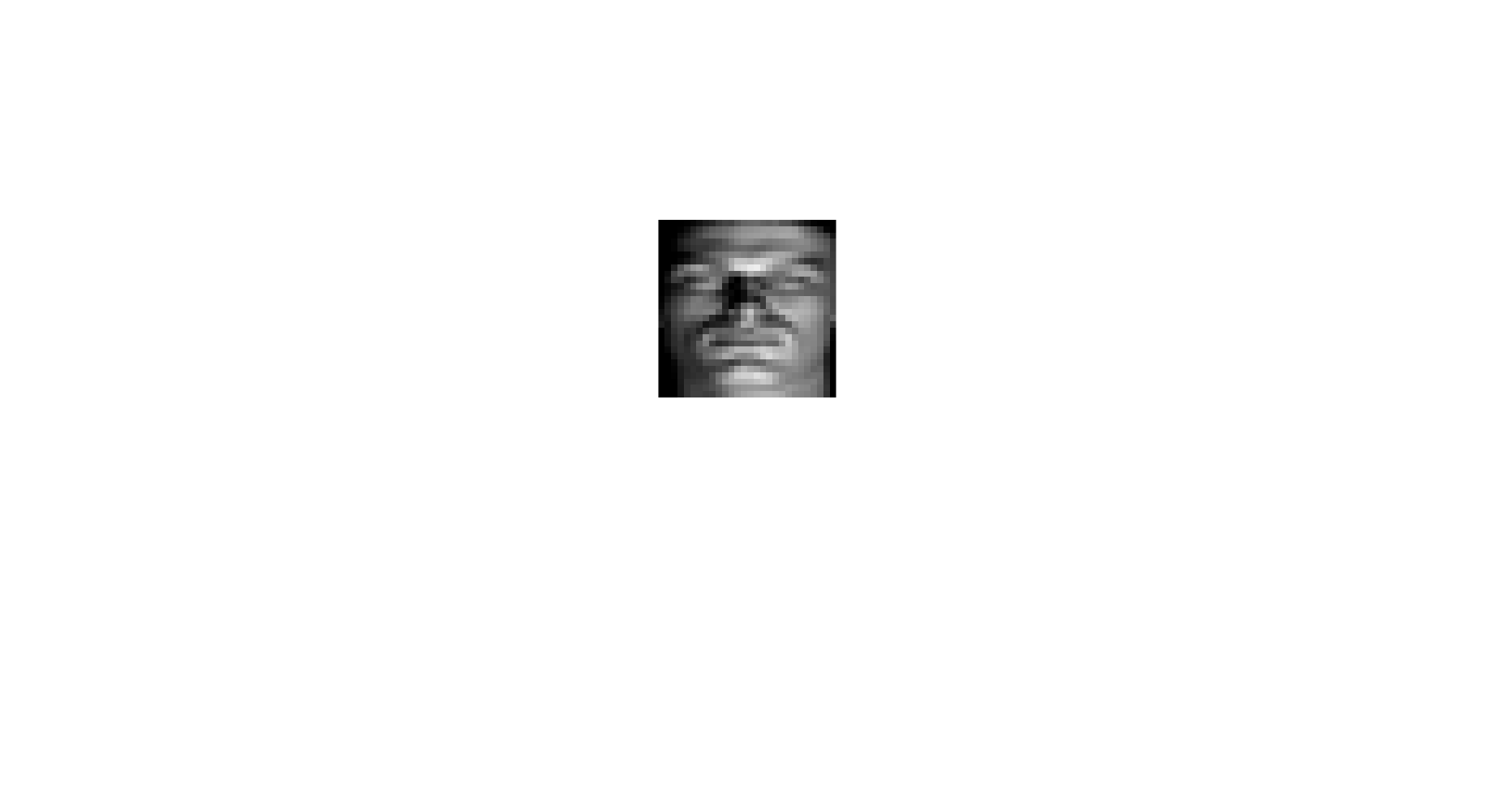} 
         & \includegraphics[trim = {14cm, 14cm, 14cm, 6cm}, clip, width=0.22\textwidth]{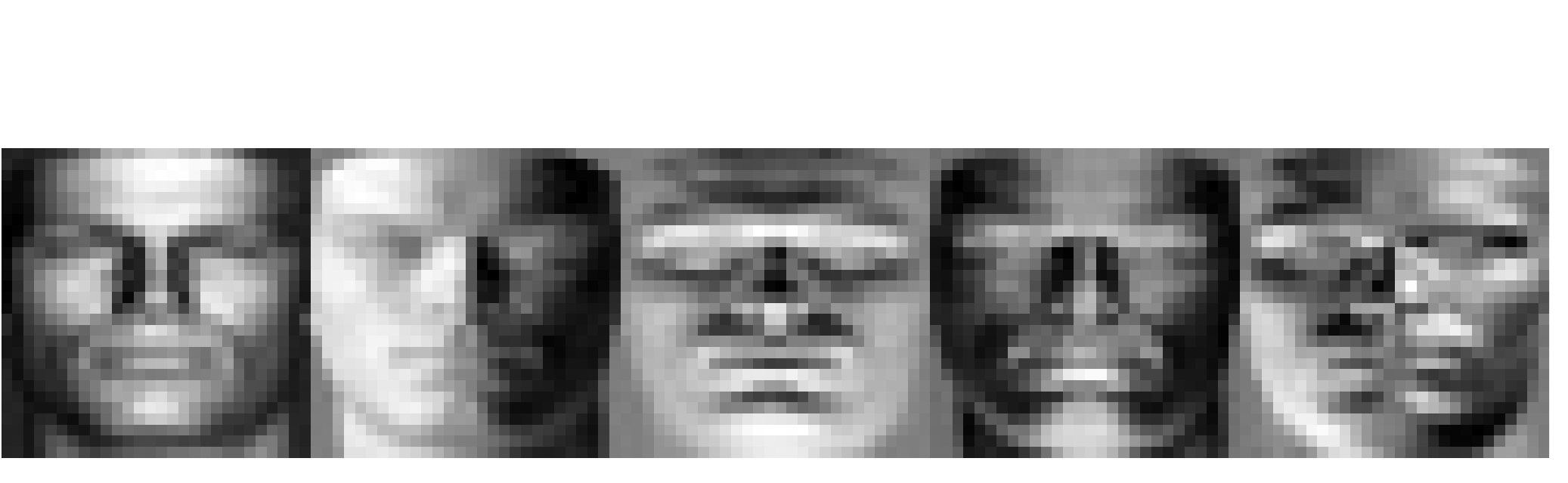} 
         & \includegraphics[trim = {14cm, 14cm, 14cm, 8cm}, clip, width=0.22\textwidth]{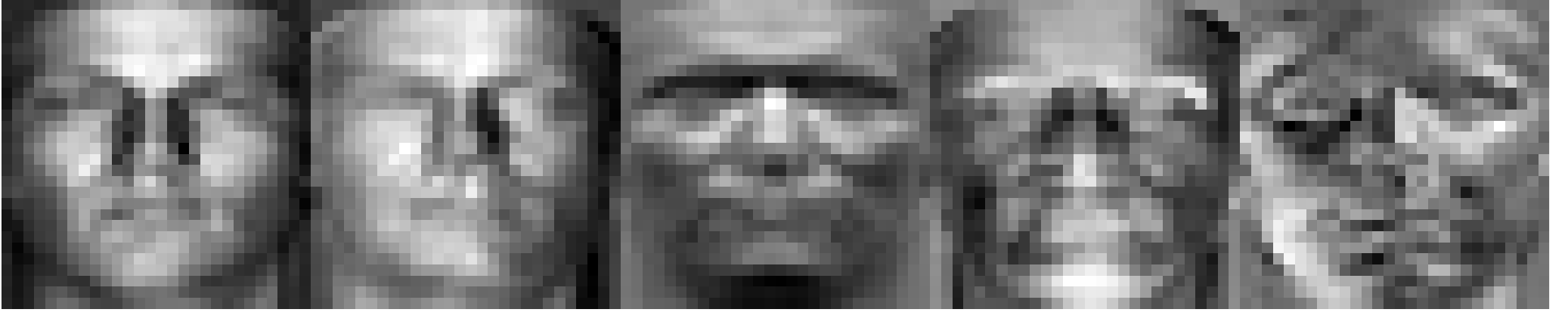}
         & \includegraphics[trim = {14cm, 14cm, 14cm, 8cm}, clip, width=0.22\textwidth]{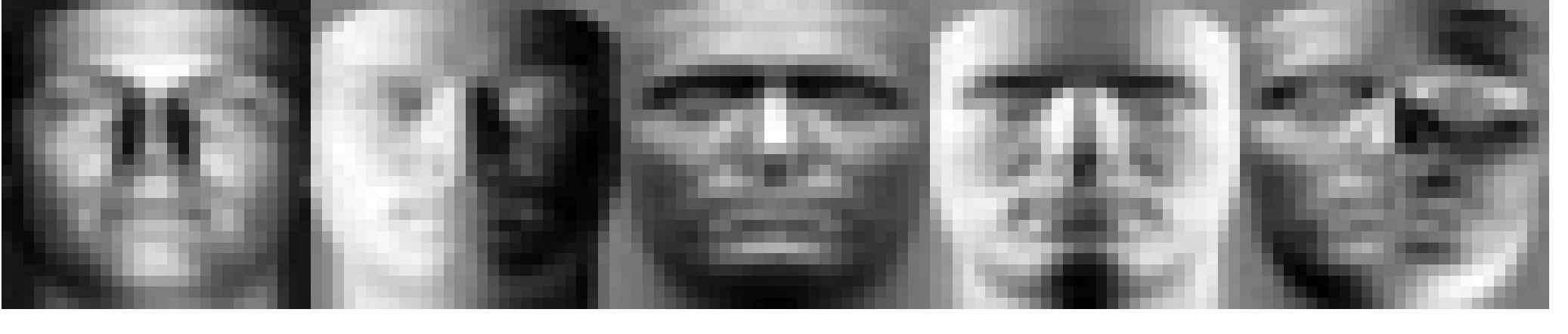}\\
         
         & & $D_G = 1.2998$ & $D_G = 0.7238$ \\
         \hline
         
	      \includegraphics[trim = {18cm, 13cm, 18cm, 6cm}, clip, width=0.11\textwidth]{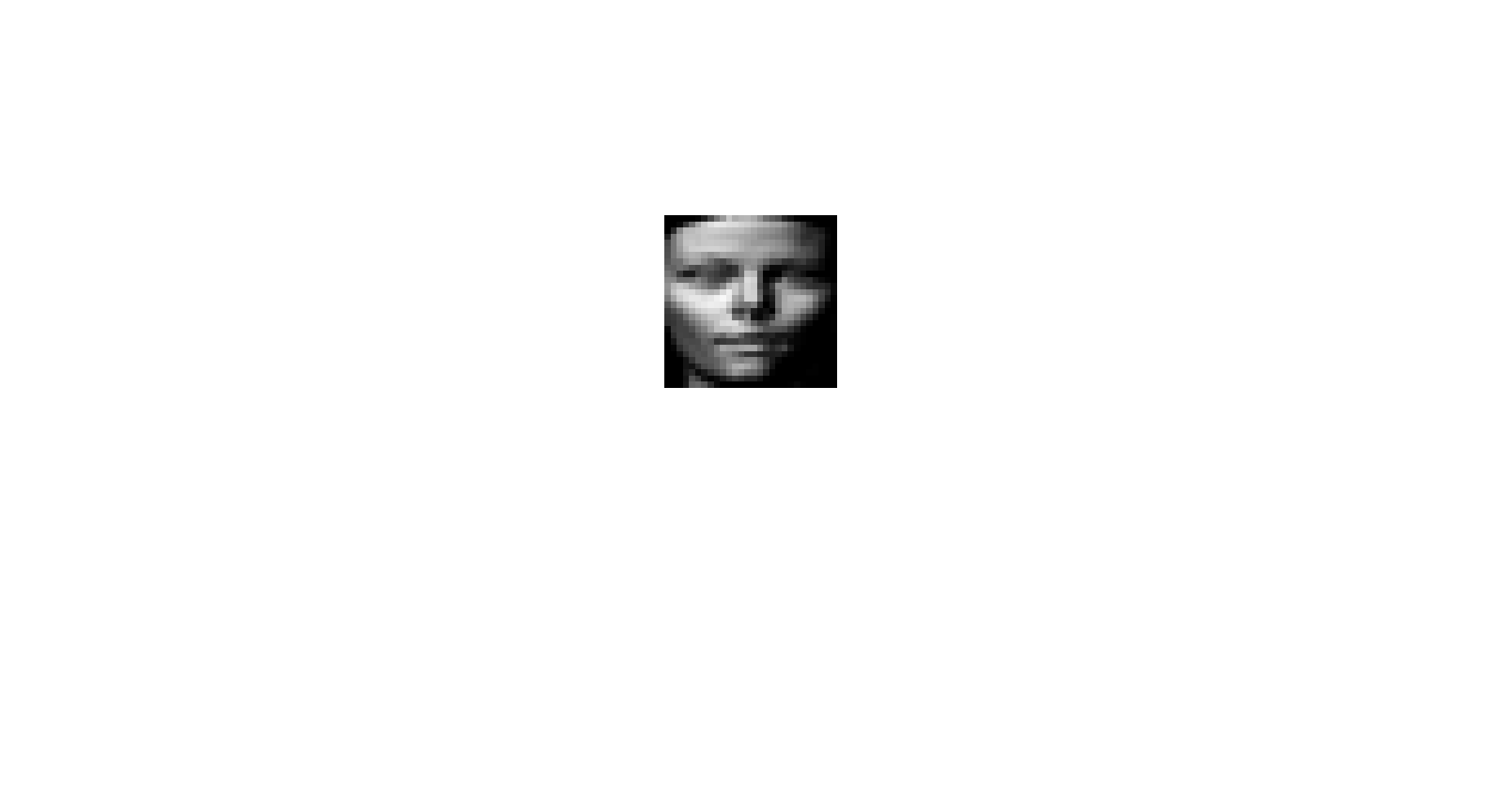} 
	      & \includegraphics[trim = {14cm, 14cm, 14cm, 6cm}, clip, width=0.22\textwidth]{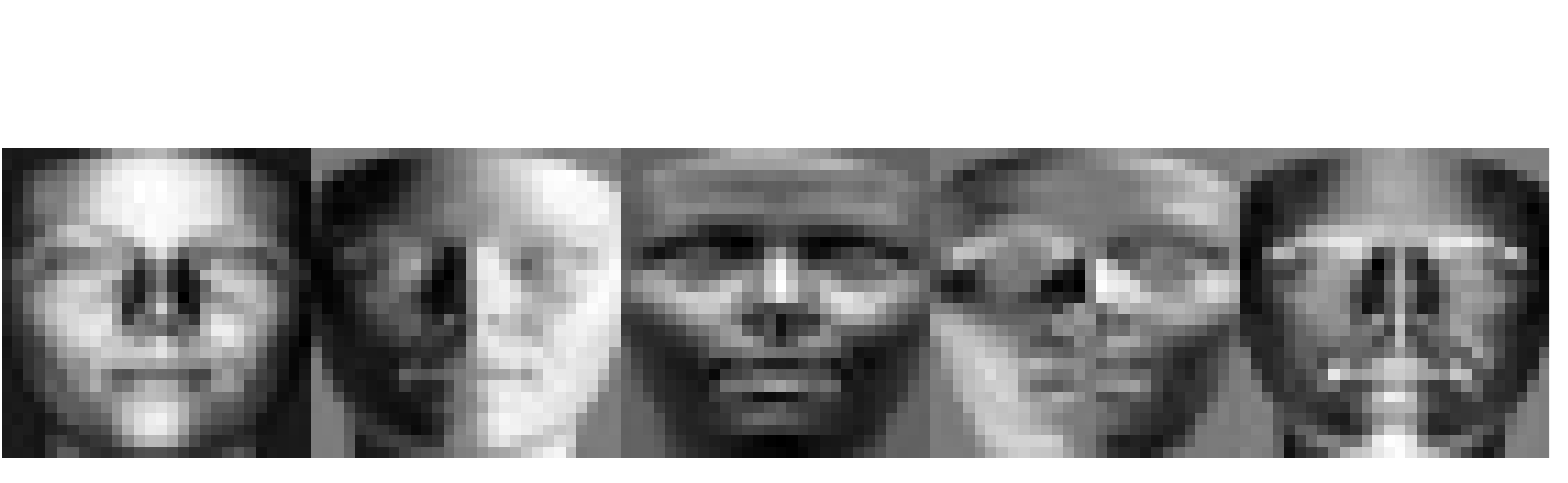} 
	      & \includegraphics[trim = {14cm, 14cm, 14cm, 8cm}, clip, width=0.22\textwidth]{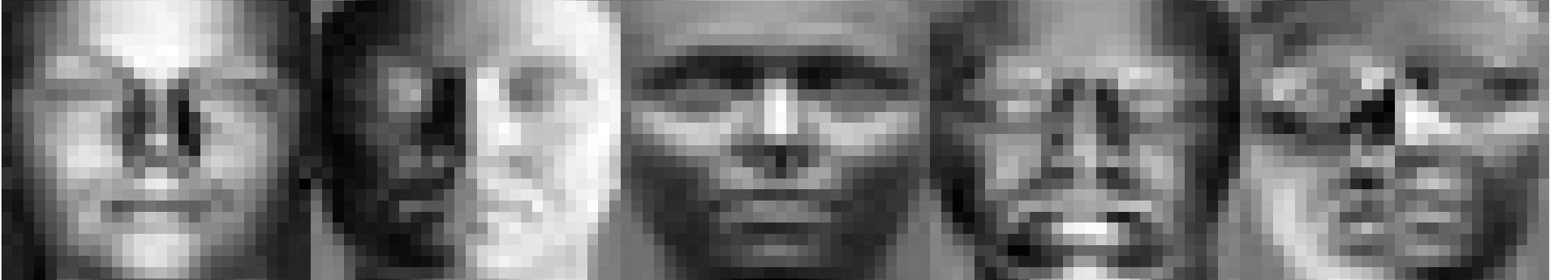}
	      & \includegraphics[trim = {14cm, 14cm, 14cm, 8cm}, clip, width=0.22\textwidth]{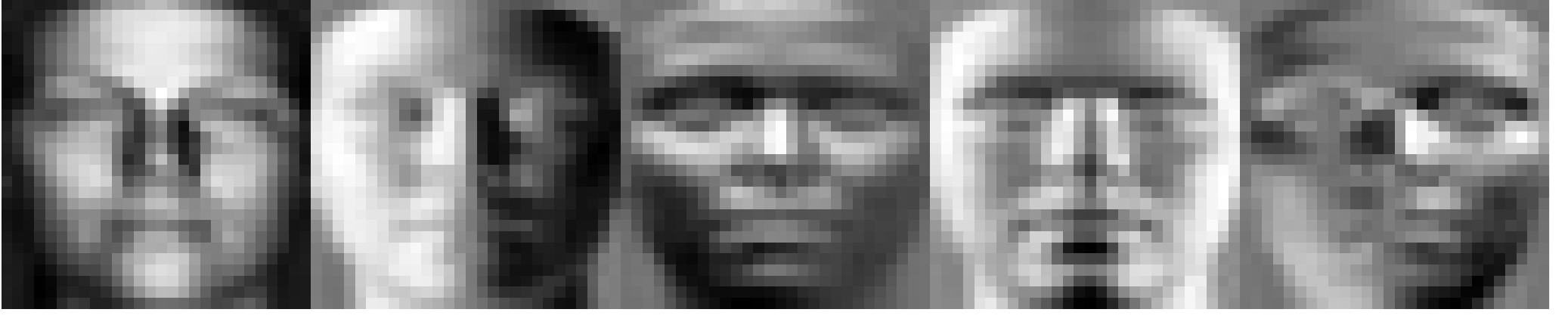}\\
	      
	      & & $D_G = 0.7797$ & $D_G = 0.5966$ \\
	      \midrule
	      
	      \includegraphics[trim = {18cm, 13cm, 18cm, 6cm}, clip, width=0.11\textwidth]{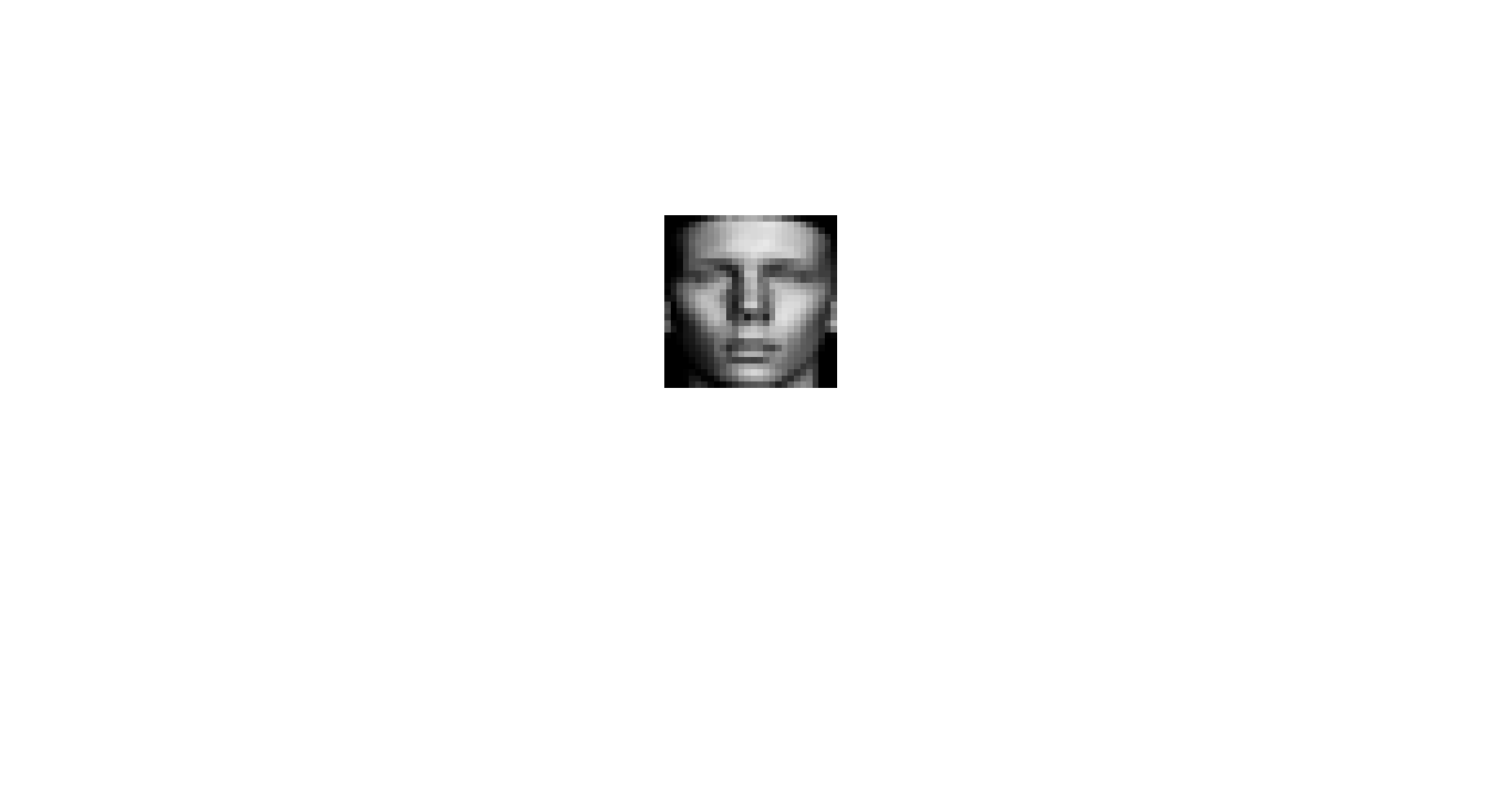} 
	      & \includegraphics[trim = {14cm, 14cm, 14cm, 6cm}, clip, width=0.22\textwidth]{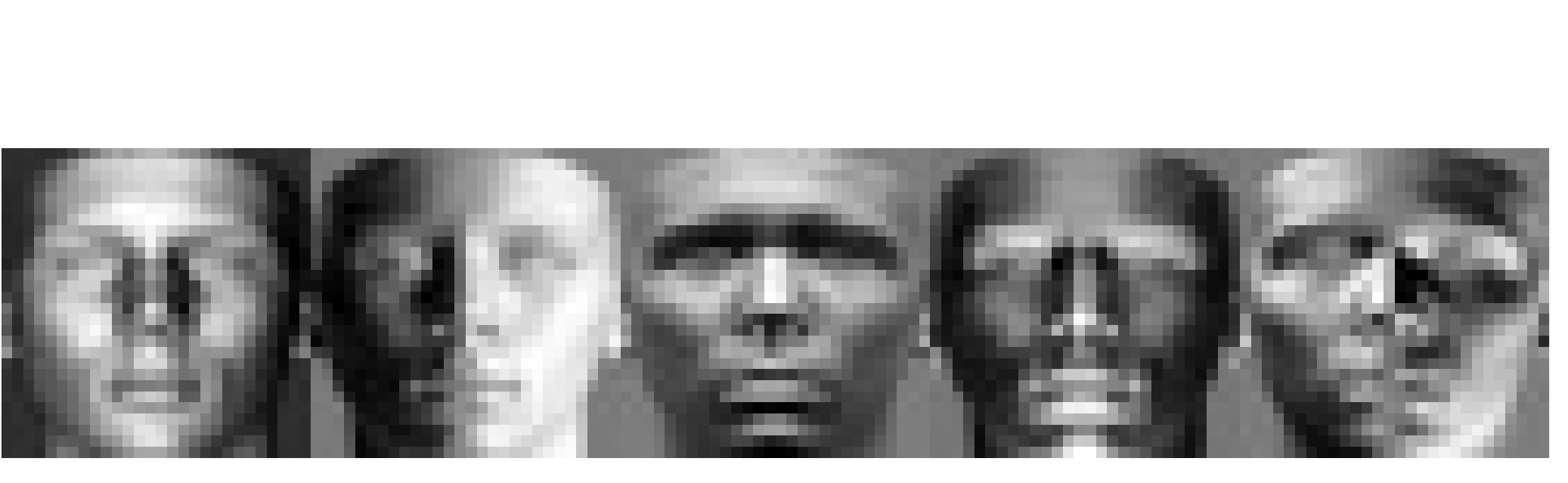} 
	      & \includegraphics[trim = {14cm, 14cm, 14cm, 8cm}, clip, width=0.22\textwidth]{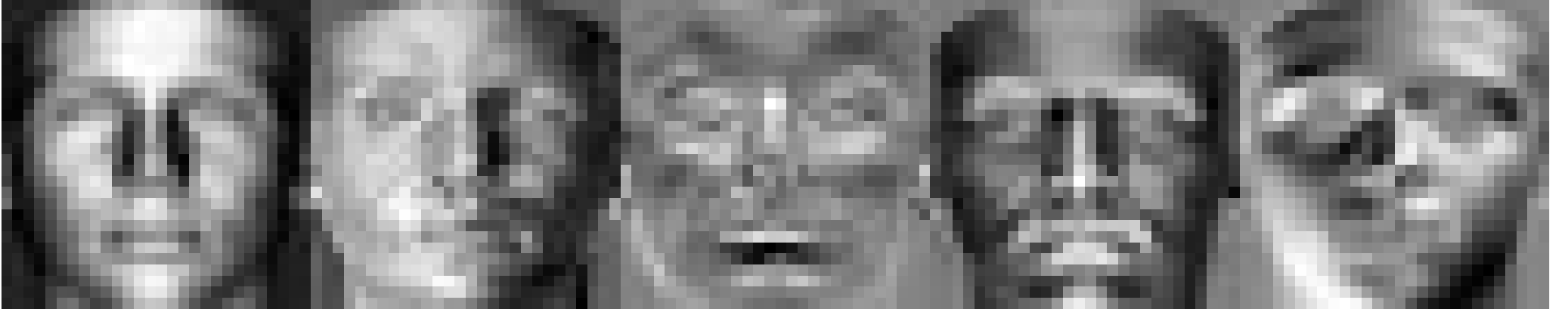}
	      & \includegraphics[trim = {14cm, 14cm, 14cm, 8cm}, clip, width=0.22\textwidth]{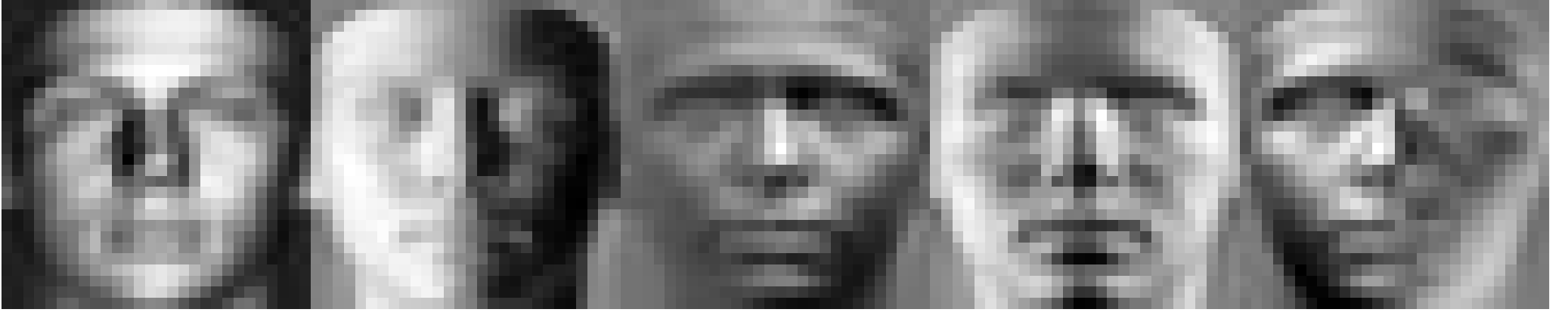}\\
	      
	      & & $D_G = 1.5355$ & $D_G = 0.6170$ \\
	      \midrule
	      
	      \includegraphics[trim = {18cm, 13cm, 18cm, 6cm}, clip, width=0.11\textwidth]{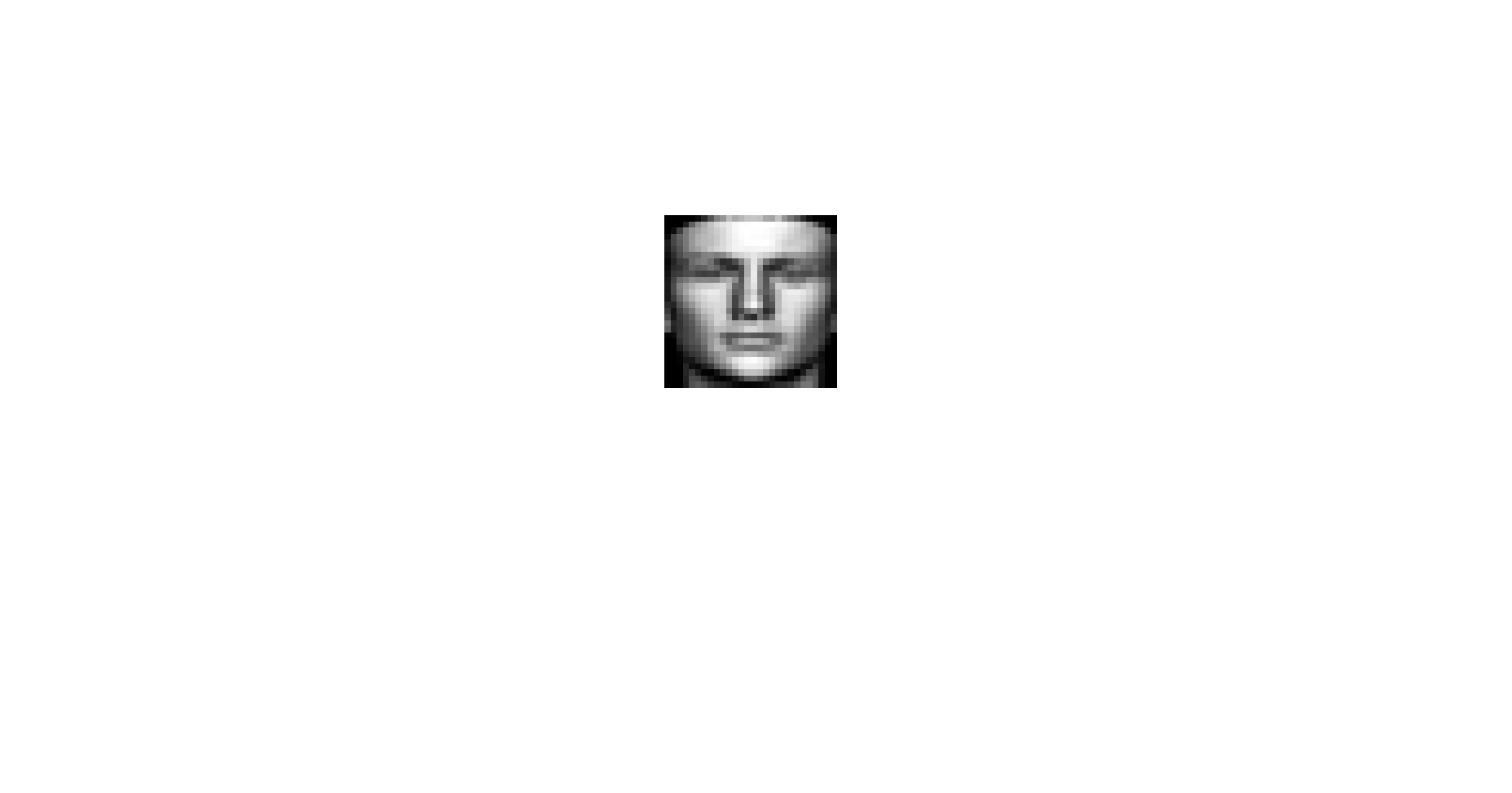} 
	      & \includegraphics[trim = {14cm, 14cm, 14cm, 6cm}, clip, width=0.22\textwidth]{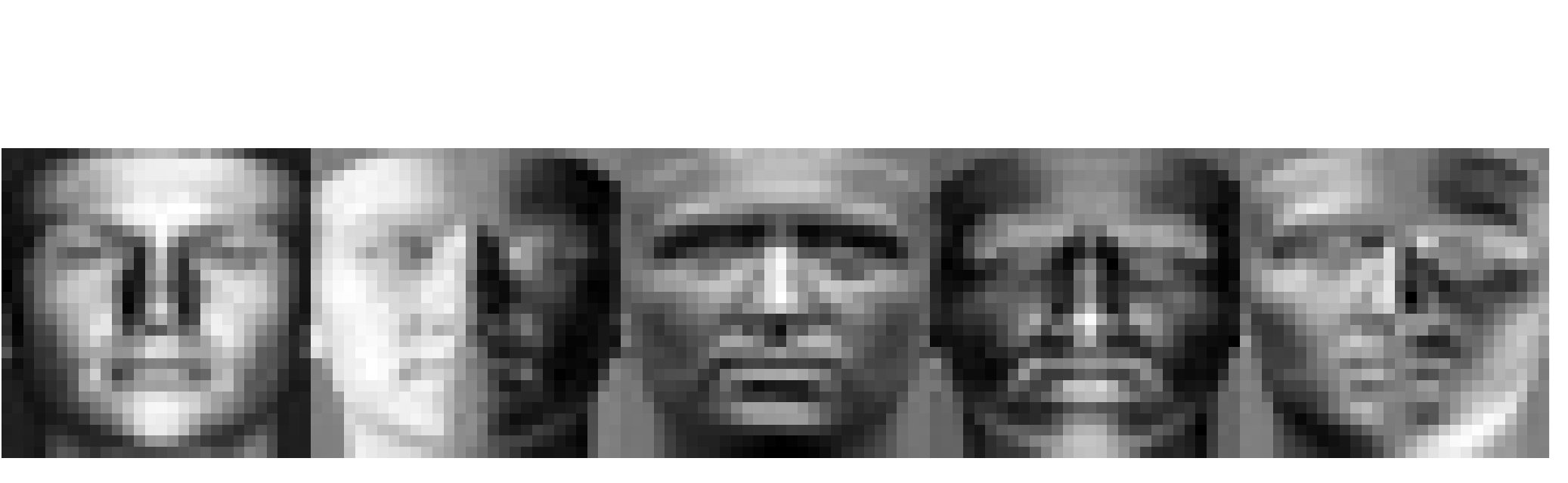} 
	      & \includegraphics[trim = {14cm, 14cm, 14cm, 8cm}, clip, width=0.22\textwidth]{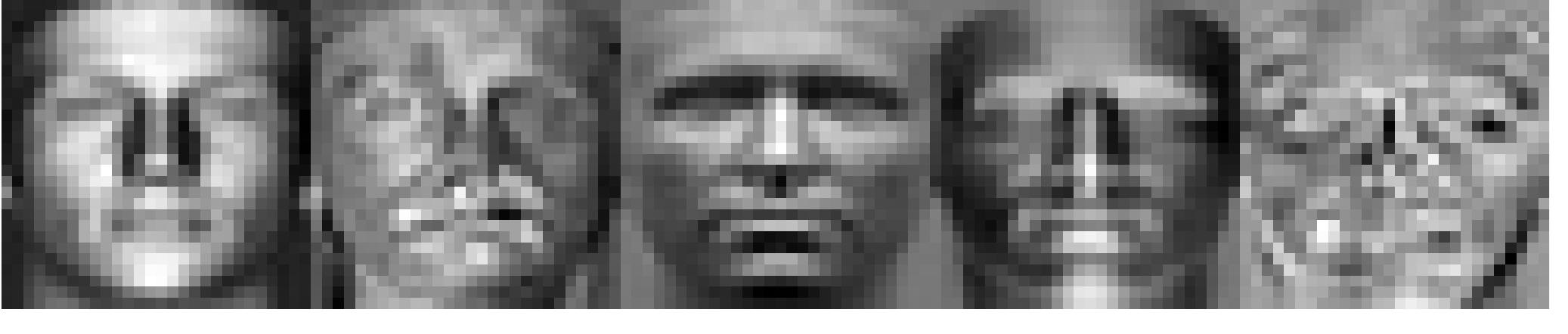}
	      & \includegraphics[trim = {14cm, 14cm, 14cm, 8cm}, clip, width=0.22\textwidth]{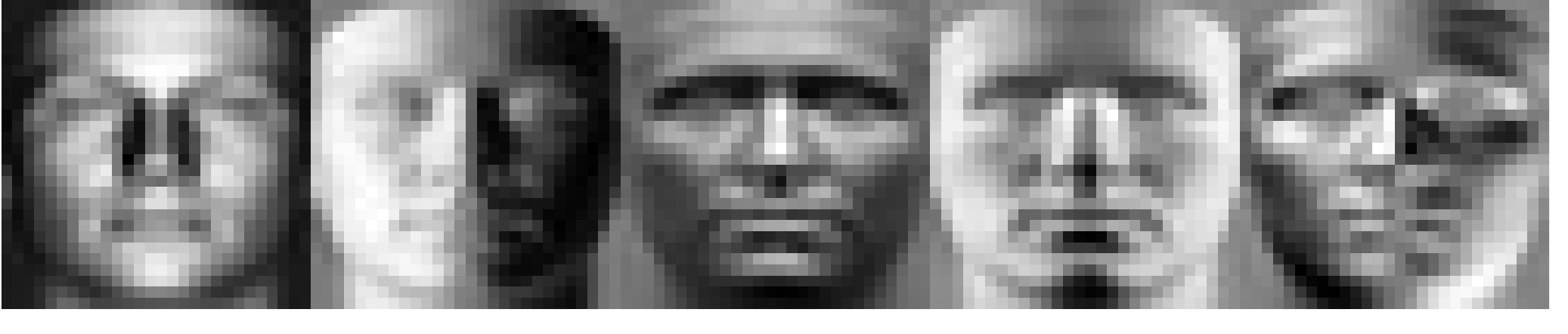}\\
	      
	      & & $D_G = 1.6760$ & $D_G = 0.4420$ \\
	      \midrule
	      
	      \includegraphics[trim = {18cm, 13cm, 18cm, 6cm}, clip, width=0.11\textwidth]{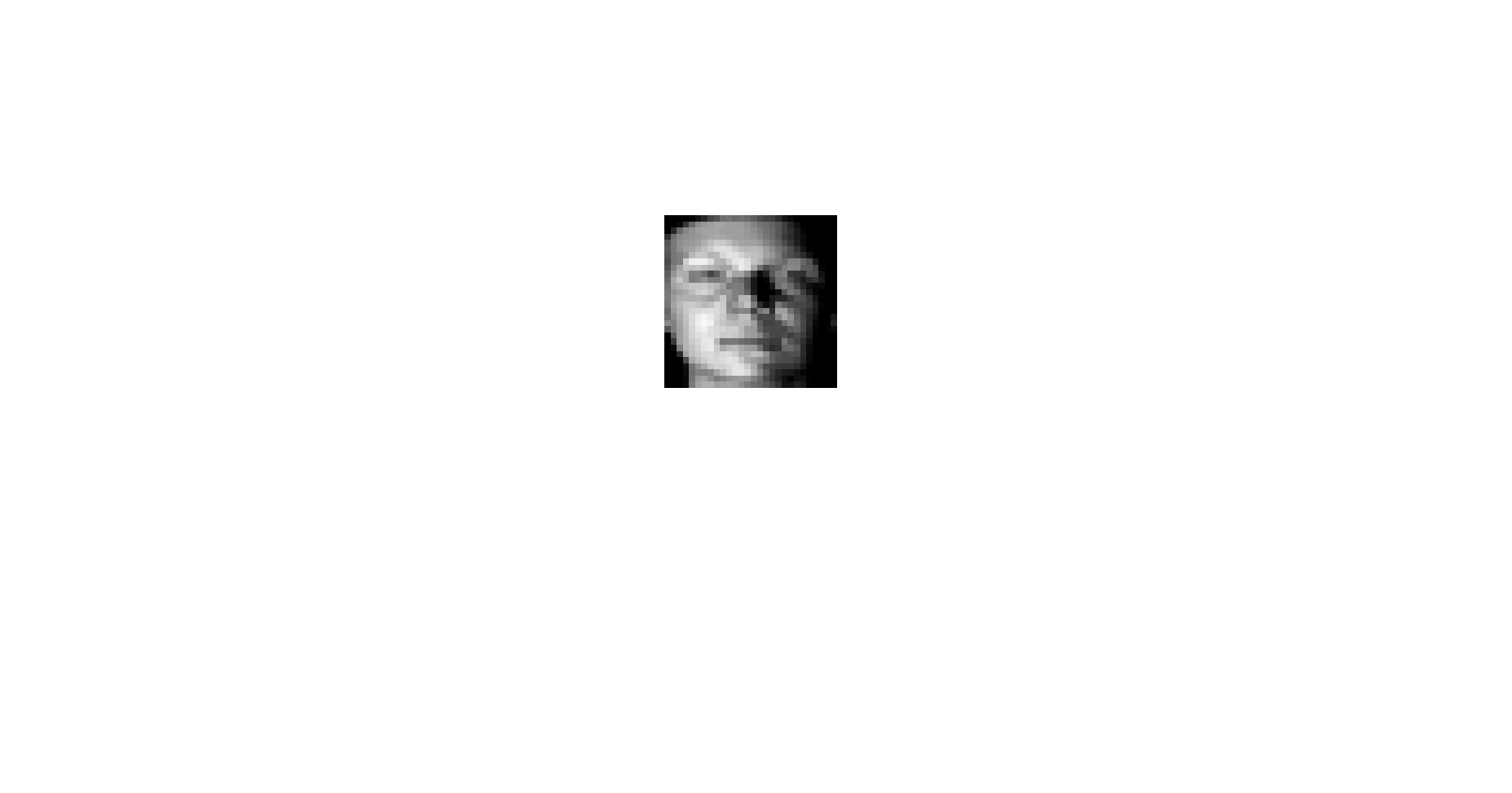} 
	      & \includegraphics[trim = {14cm, 14cm, 14cm, 6cm}, clip, width=0.22\textwidth]{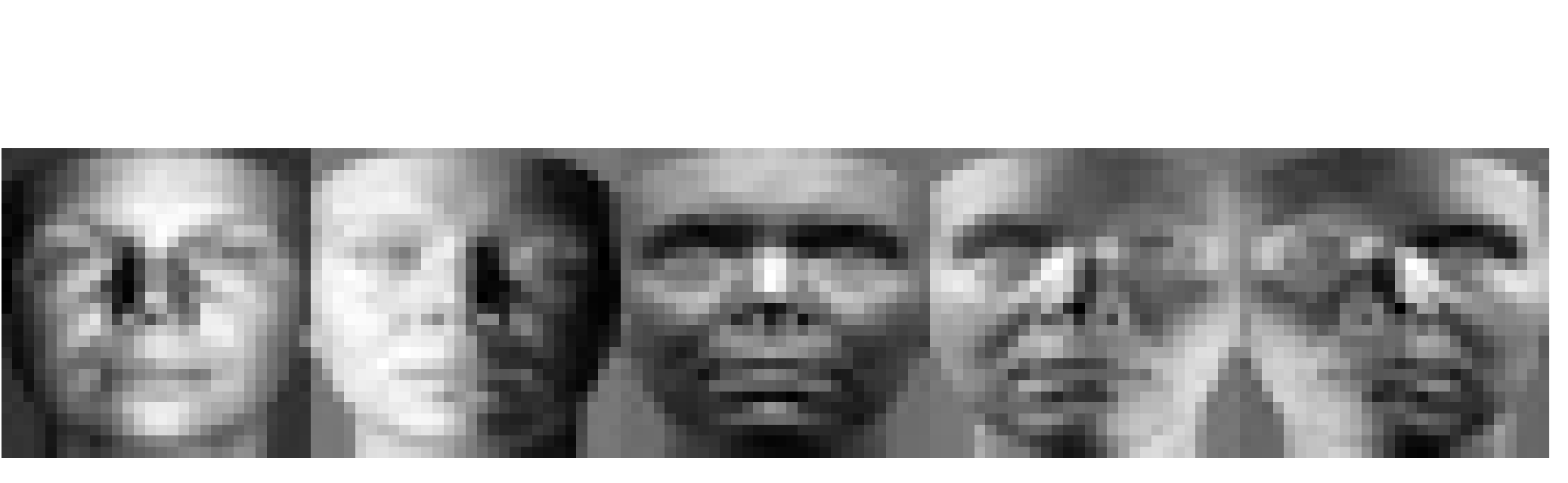} 
	      & \includegraphics[trim = {14cm, 14cm, 14cm, 8cm}, clip, width=0.22\textwidth]{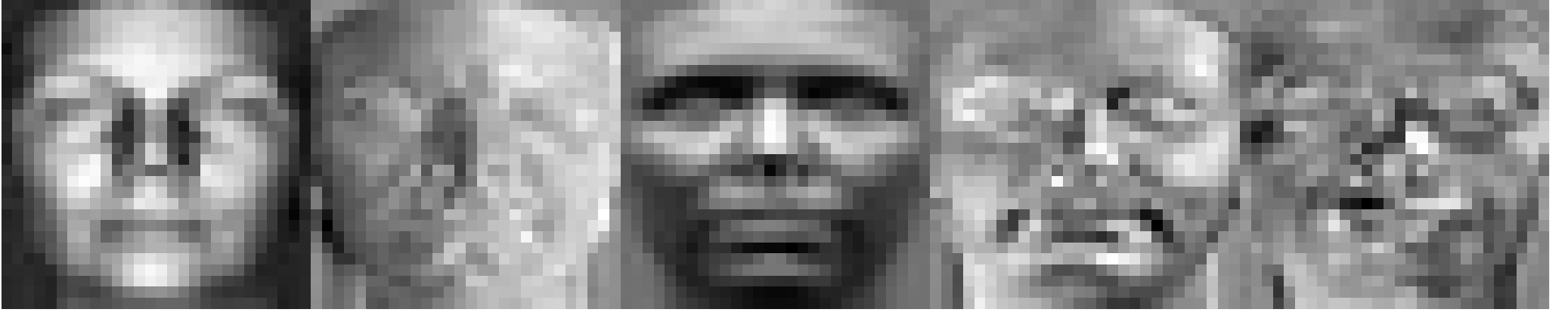}
	      & \includegraphics[trim = {14cm, 14cm, 14cm, 8cm}, clip, width=0.22\textwidth]{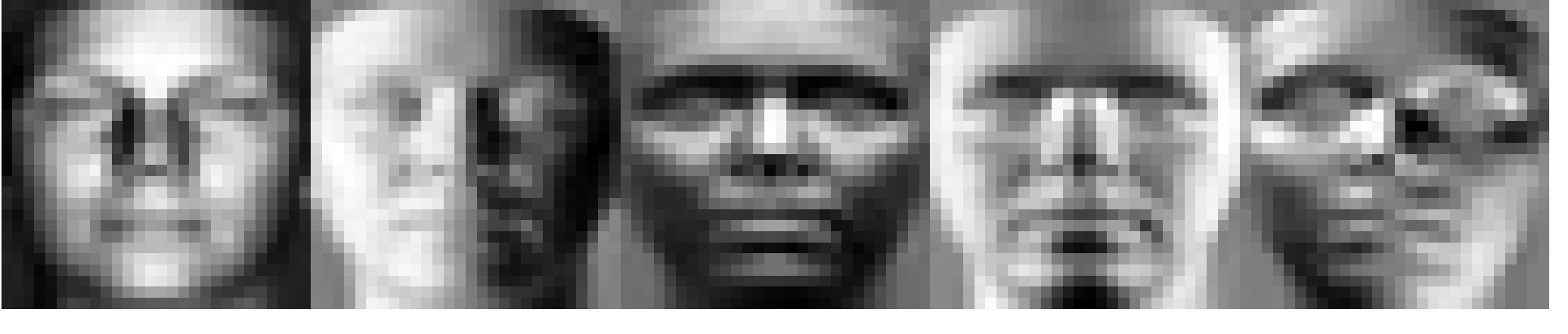}\\
	      
	      & & $D_G = 1.6703$ & $D_G = 0.4939$ \\
    \end{tabular}
    \caption{Test results for two input images using $d=5$. We can clearly observe that the GrassmannNet-TS (with pole $\mathbf{U}_{Fr}^d$) framework performs much better than the baseline that attempts to regress directly to the PCs. The numbers below the output images indicate the subspace distance from the ground truth (lower the better). Note that the outputs need not be exactly the same as the groundtruth PCs since the quantity of interest is the subspace spanned by the groundtruth PCs. See Supplementary Material for more results.}\label{table:f2is_results_5}
\end{table*}

\begin{table*}[]
    \centering
    \begin{tabular}{cccc}
     
        \hline
        Input & Ground-truth PCs & Output of baseline n/w & Output of GrassmannNet-TS \\
        \hline
       \includegraphics[trim = {18cm, 13cm, 18cm, 6cm}, clip, width=0.11\textwidth]{figures/input_image_1.png} 
         & \includegraphics[trim = {14cm, 14cm, 14cm, 6cm}, clip, width=0.22\textwidth]{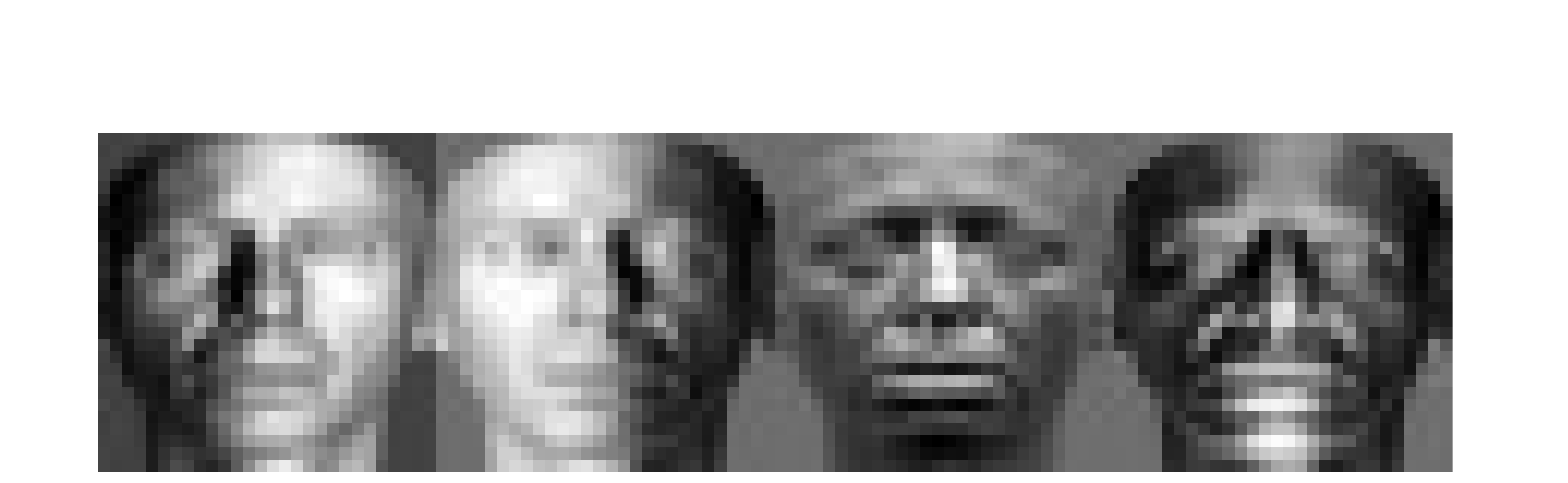} 
         & \includegraphics[trim = {14cm, 14cm, 14cm, 8cm}, clip, width=0.22\textwidth]{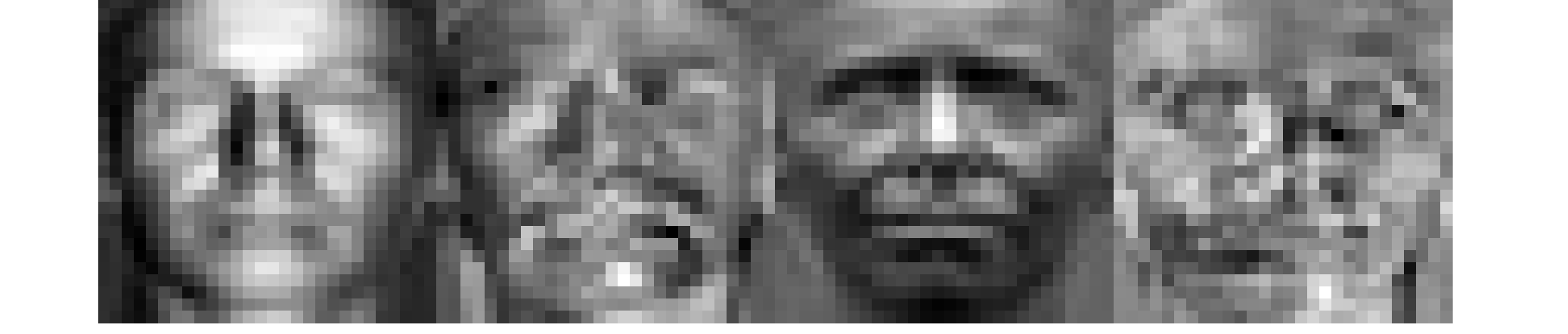}
         & \includegraphics[trim = {14cm, 14cm, 14cm, 8cm}, clip, width=0.22\textwidth]{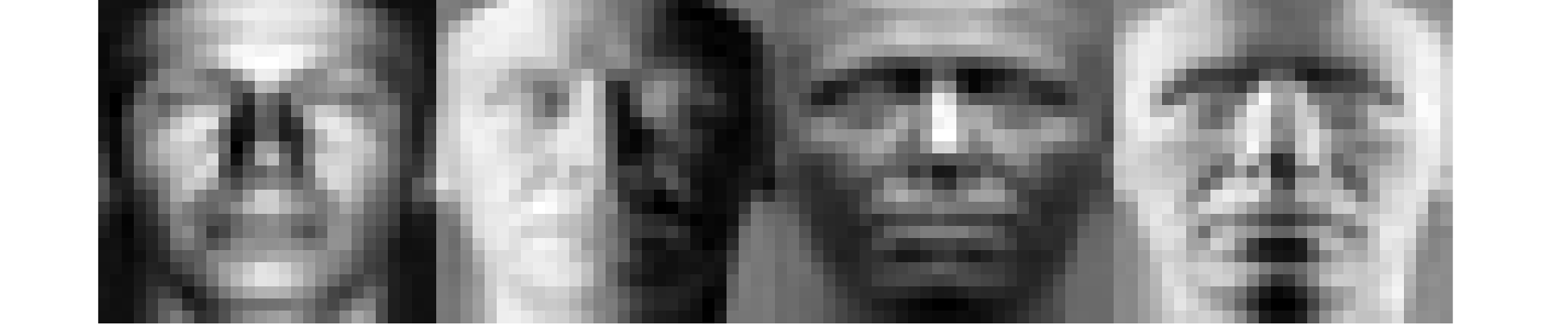}\\
         
         & & $D_G = 1.9400$ & $D_G = 0.5489$ \\
         \midrule
         
        \includegraphics[trim = {18cm, 13cm, 18cm, 6cm}, clip, width=0.11\textwidth]{figures/input_image_2.png} 
         & \includegraphics[trim = {14cm, 14cm, 14cm, 6cm}, clip, width=0.22\textwidth]{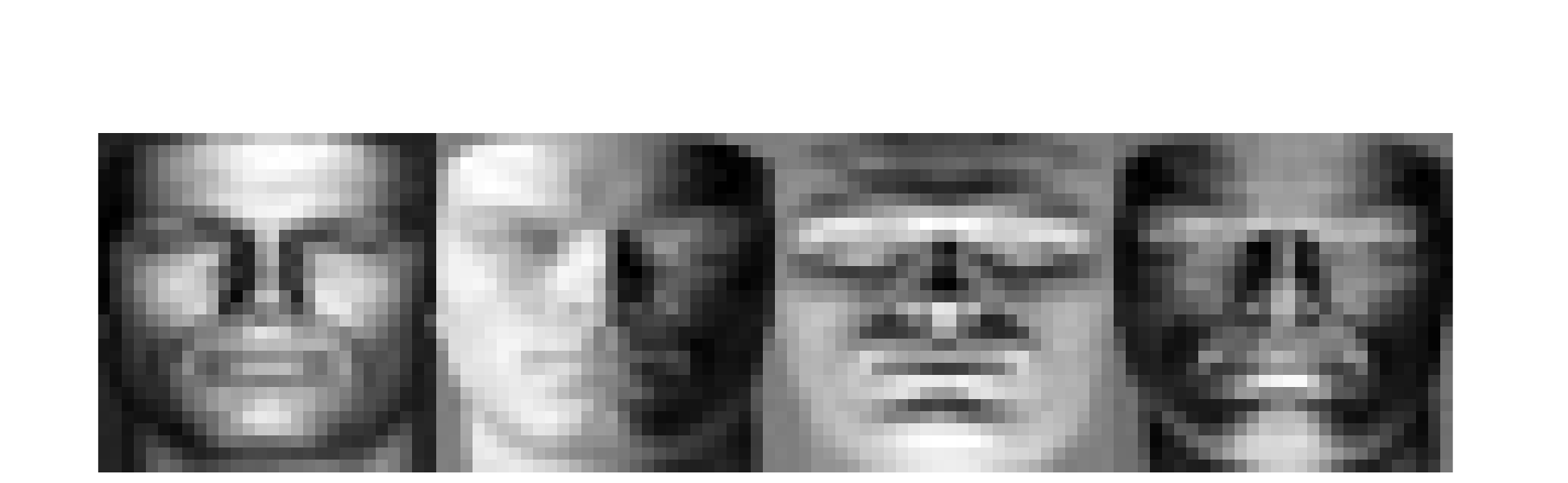} 
         & \includegraphics[trim = {14cm, 14cm, 14cm, 8cm}, clip, width=0.22\textwidth]{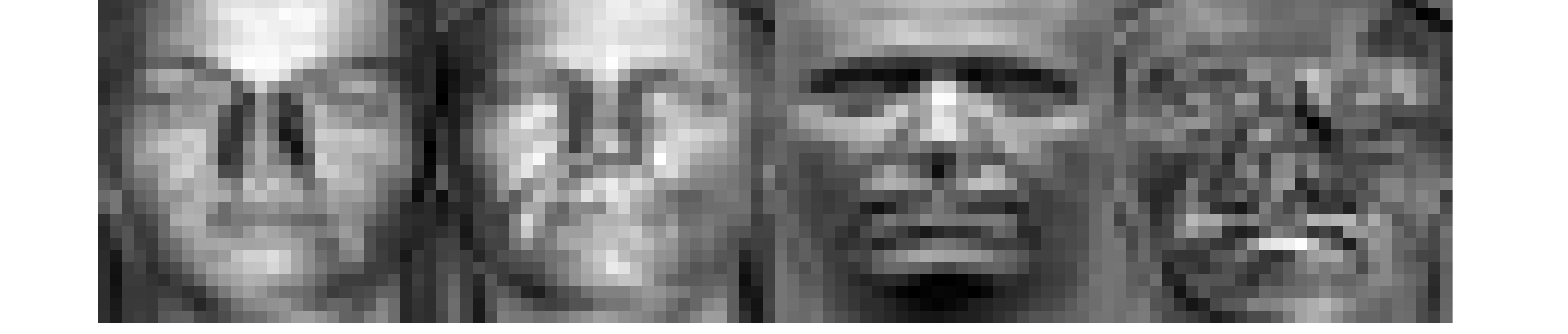}
         & \includegraphics[trim = {14cm, 14cm, 14cm, 8cm}, clip, width=0.22\textwidth]{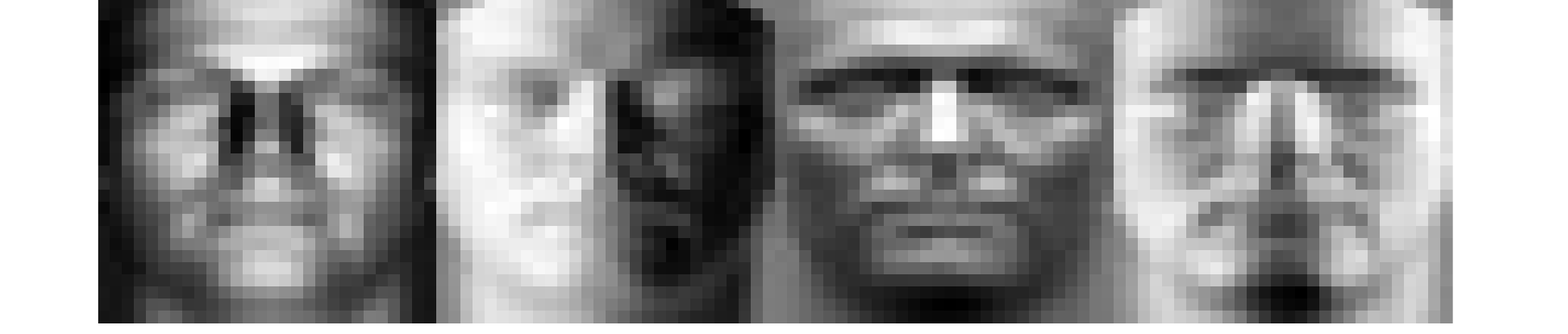}\\
         
         & & $D_G = 1.2735$ & $D_G = 0.6245$ \\
         
         \includegraphics[trim = {18cm, 13cm, 18cm, 6cm}, clip, width=0.11\textwidth]{figures/input_image_3.png} 
          & \includegraphics[trim = {14cm, 14cm, 14cm, 6cm}, clip, width=0.22\textwidth]{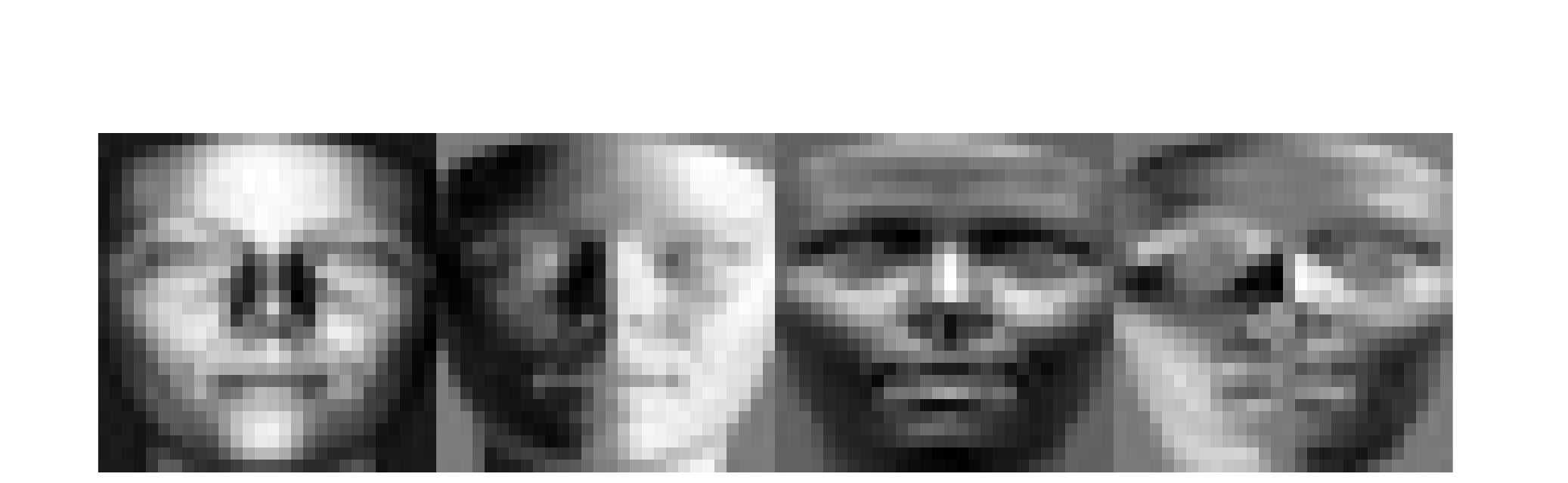} 
          & \includegraphics[trim = {14cm, 14cm, 14cm, 8cm}, clip, width=0.22\textwidth]{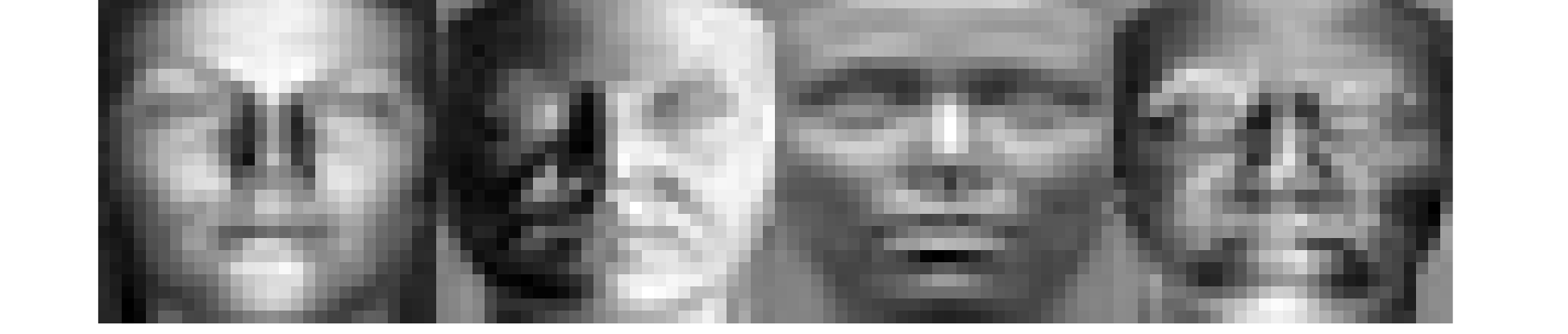}
          & \includegraphics[trim = {14cm, 14cm, 14cm, 8cm}, clip, width=0.22\textwidth]{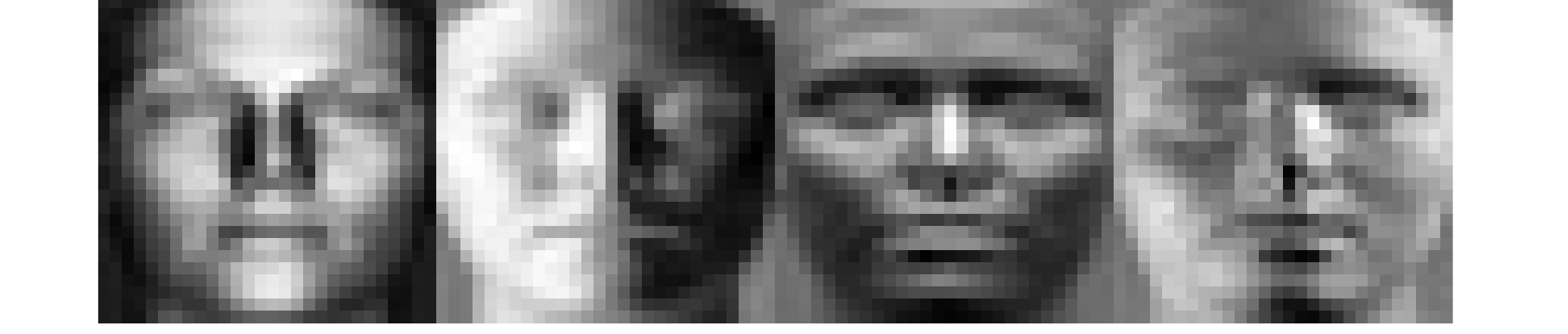}\\
          
          & & $D_G = 0.6750$ & $D_G = 1.1580$ \\
          \midrule
          
         \includegraphics[trim = {18cm, 13cm, 18cm, 6cm}, clip, width=0.11\textwidth]{figures/input_image_4.png} 
          & \includegraphics[trim = {14cm, 14cm, 14cm, 6cm}, clip, width=0.22\textwidth]{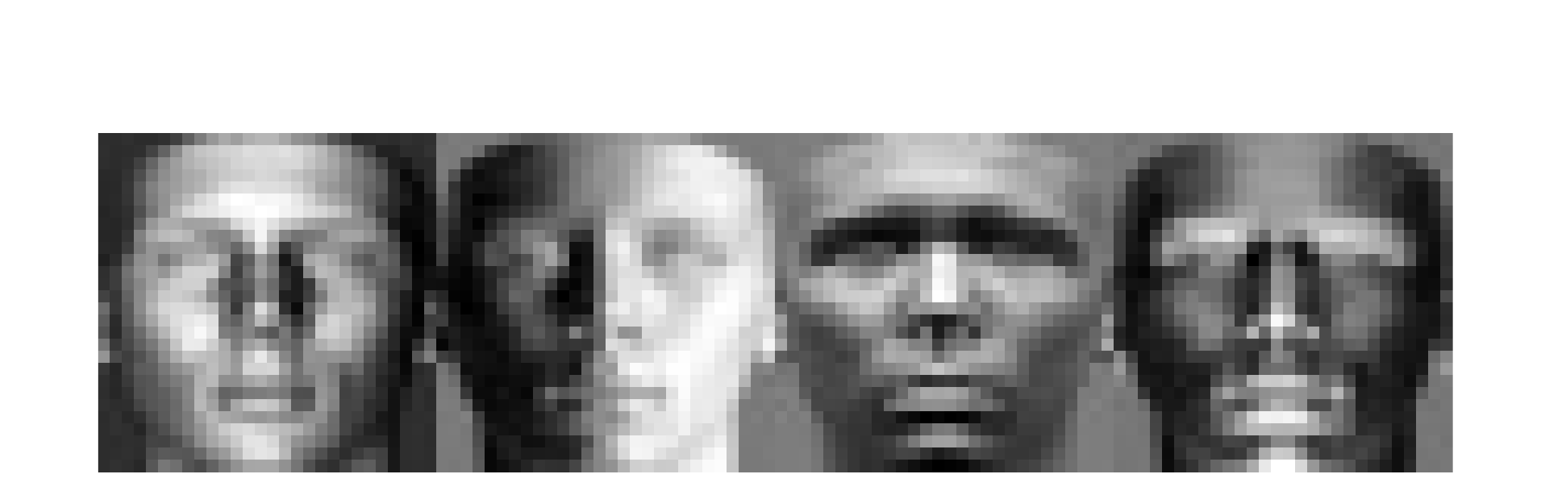} 
          & \includegraphics[trim = {14cm, 14cm, 14cm, 8cm}, clip, width=0.22\textwidth]{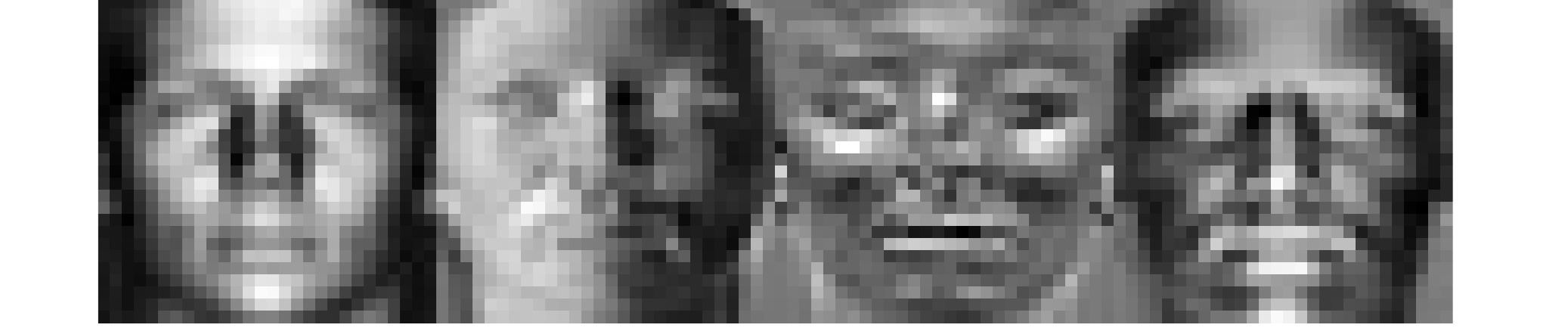}
          & \includegraphics[trim = {14cm, 14cm, 14cm, 8cm}, clip, width=0.22\textwidth]{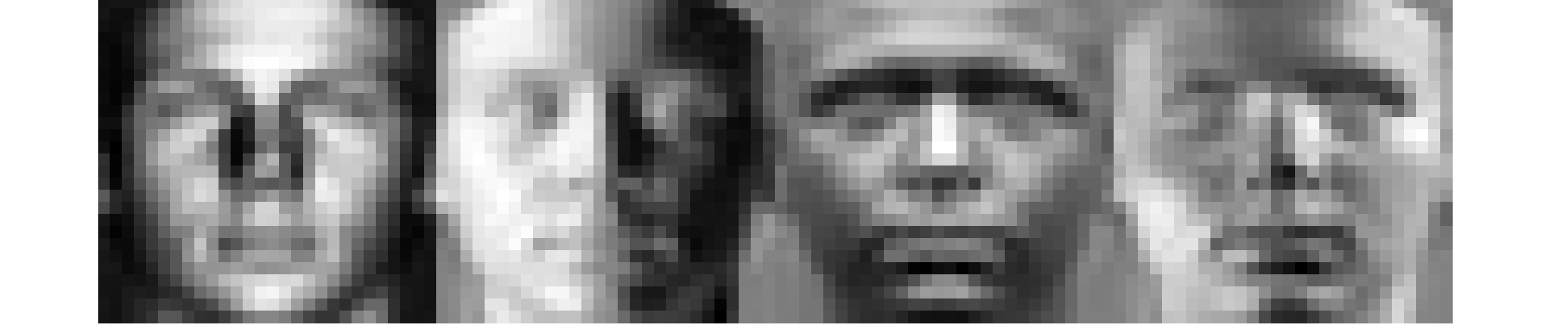}\\
          
          & & $D_G = 1.2047$ & $D_G = 0.6674$ \\
          \midrule
 
          \includegraphics[trim = {18cm, 13cm, 18cm, 6cm}, clip, width=0.11\textwidth]{figures/input_image_5.png} 
          & \includegraphics[trim = {14cm, 14cm, 14cm, 6cm}, clip, width=0.22\textwidth]{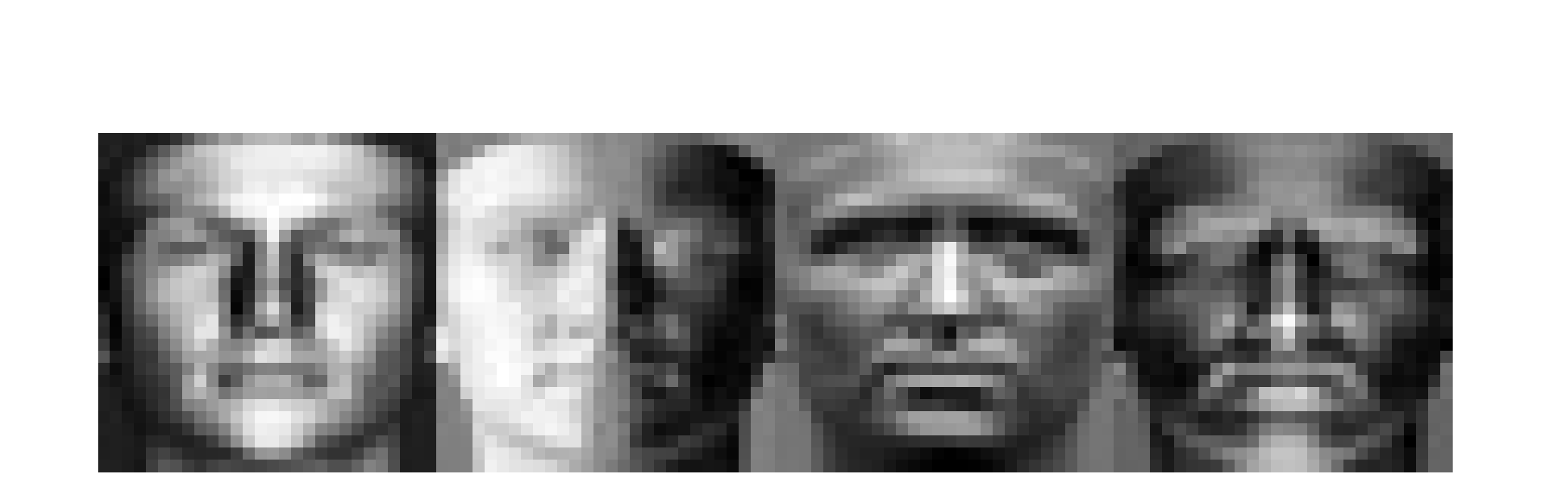} 
          & \includegraphics[trim = {14cm, 14cm, 14cm, 8cm}, clip, width=0.22\textwidth]{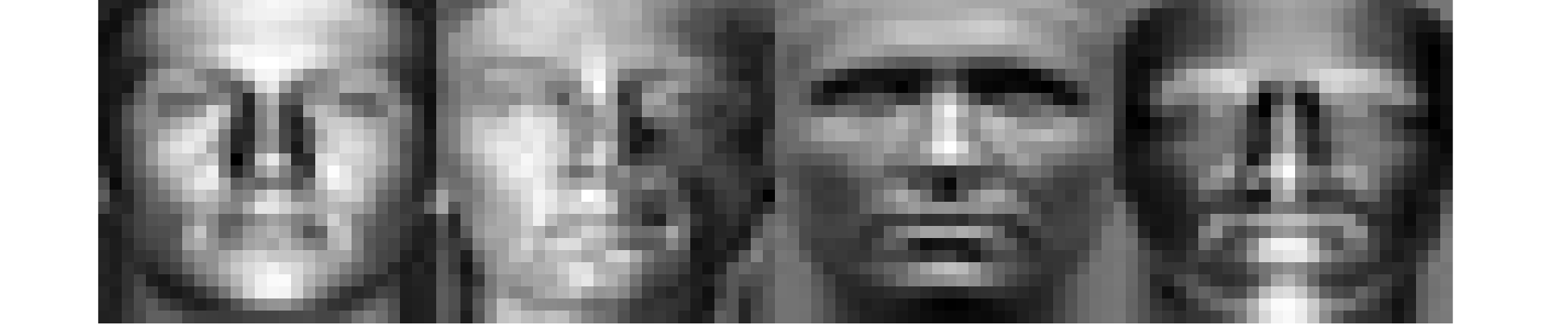}
          & \includegraphics[trim = {14cm, 14cm, 14cm, 8cm}, clip, width=0.22\textwidth]{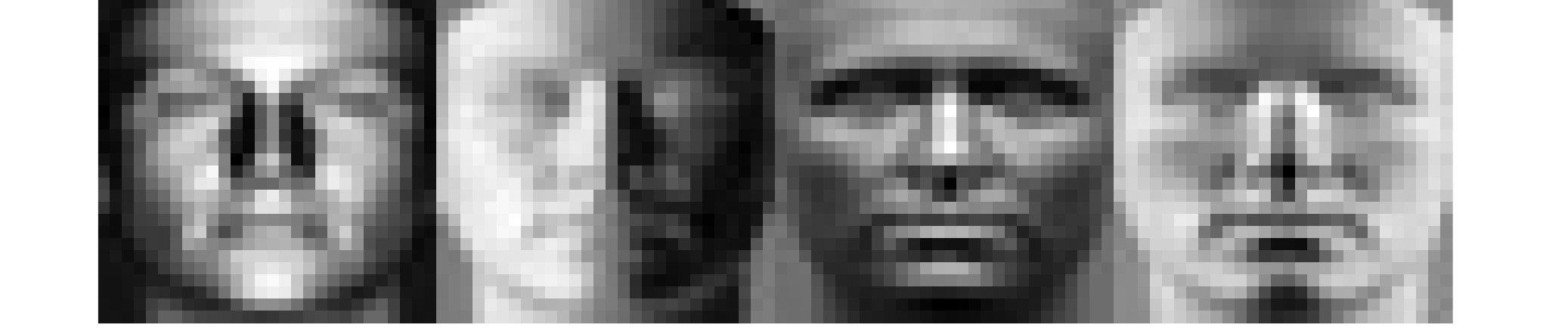}\\
          
          & & $D_G = 0.6225$ & $D_G = 0.3653$ \\
          \midrule
          
         \includegraphics[trim = {18cm, 13cm, 18cm, 6cm}, clip, width=0.11\textwidth]{figures/input_image_6.png} 
          & \includegraphics[trim = {14cm, 14cm, 14cm, 6cm}, clip, width=0.22\textwidth]{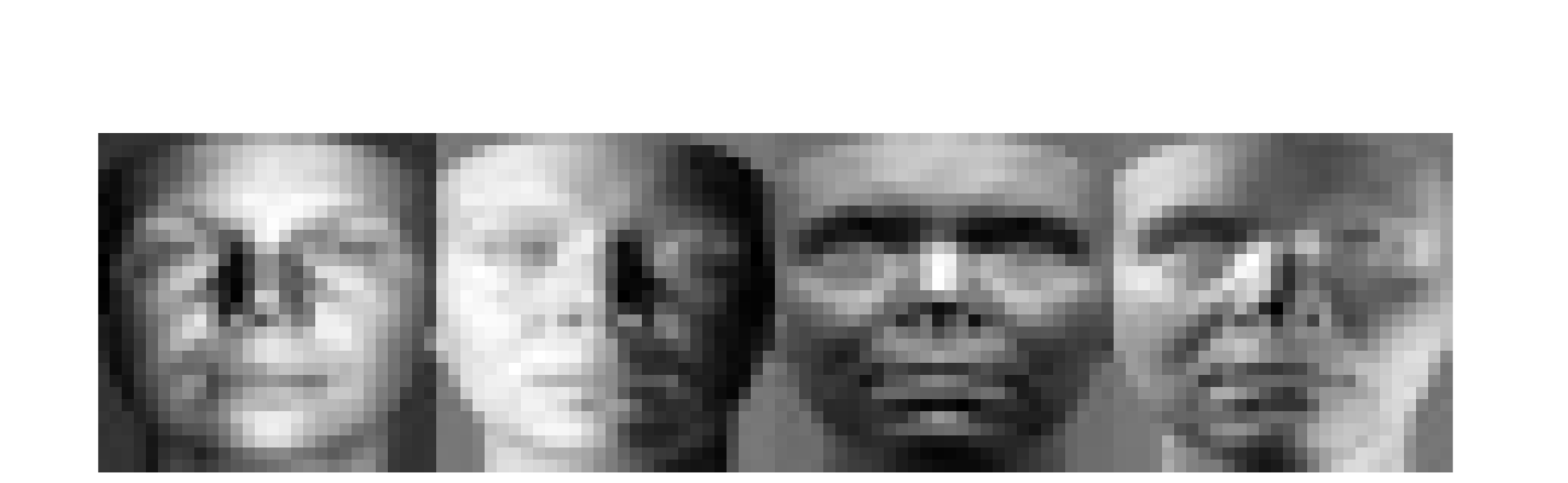} 
          & \includegraphics[trim = {14cm, 14cm, 14cm, 8cm}, clip, width=0.22\textwidth]{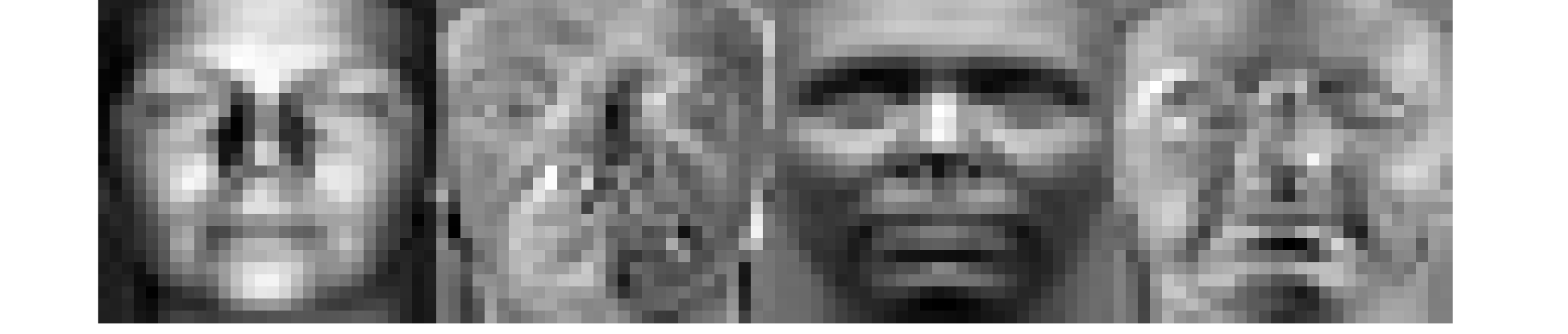}
          & \includegraphics[trim = {14cm, 14cm, 14cm, 8cm}, clip, width=0.22\textwidth]{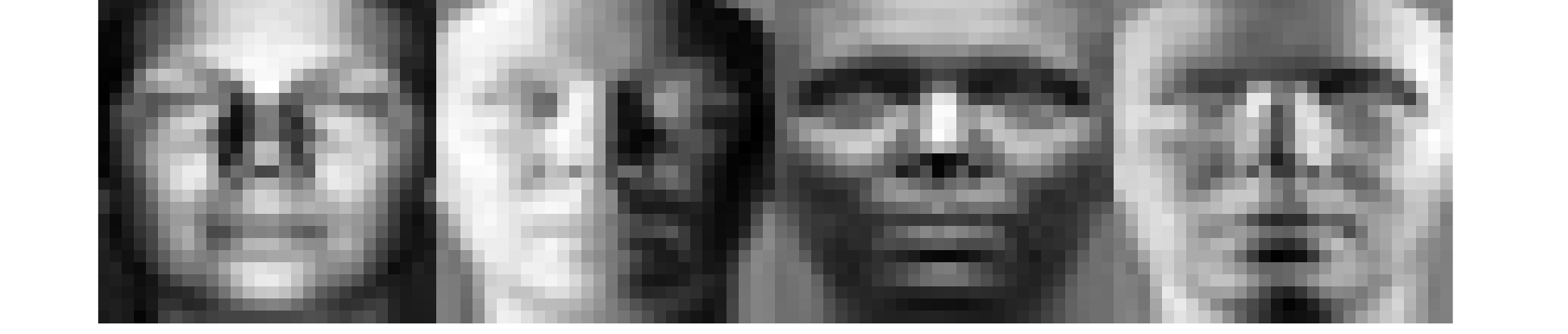}\\
          
          & & $D_G = 1.5900$ & $D_G = 0.9339$ \\
          \bottomrule
    \end{tabular}
    \caption{Test results for two input images using $d=4$. As in the case of $d=5$, GrassmannNet-TS (with pole $\mathbf{U}_{Fr}^d$) framework performs much better than the baseline that attempts to regress directly to the PCs. See Supplementary Material for more results.}\label{table:f2is_results_4}
\end{table*}

\begin{table*}[]
    \centering
    %\bgroup
    %\def\arraystretch{1.5}
    \caption{Test results for six input images using $d=3$. From the figures, we can clearly observe that the GrassmannNet-TS (with the Fr\'{e}chet mean of the training set as the pole) framework performs much better than the baseline that attempts to regress directly to the PC's. The numbers below the output images indicate the subspace distance from the ground truth (lower the better). }\label{table:f2is_results_subspace_dim_3}.
    \begin{tabular}{cccc}
     
        \toprule
        Input & Ground truth PC's & Output of baseline n/w & Output of GrassmannNet-TS \\
        \midrule
        \includegraphics[trim = {18cm, 13cm, 18cm, 6cm}, clip, width=0.11\textwidth]{figures/input_image_1.png}
         &\includegraphics[trim = {14cm, 14cm, 14cm, 6cm}, clip, width=0.22\textwidth]{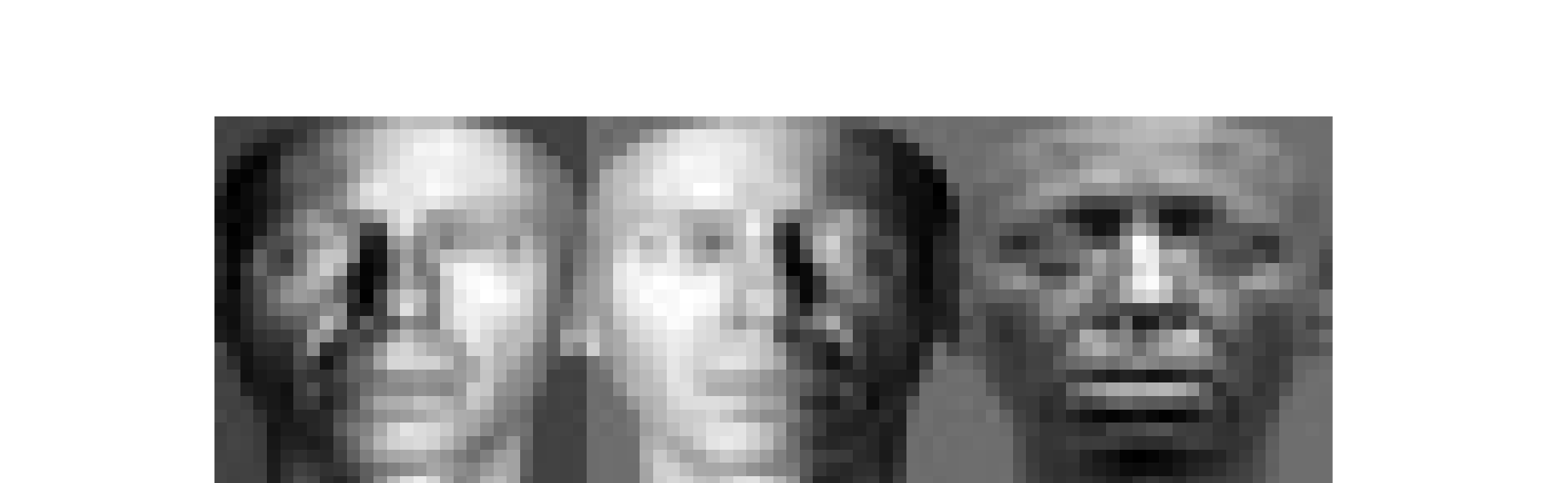}
         & \includegraphics[trim = {14cm, 14cm, 14cm, 8cm}, clip, width=0.22\textwidth]{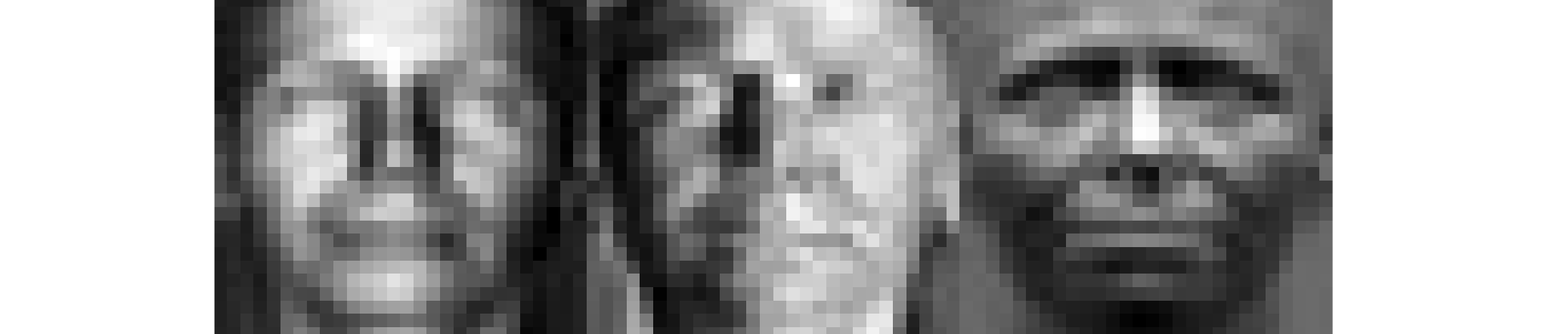}
         & \includegraphics[trim = {14cm, 14cm, 14cm, 8cm}, clip, width=0.22\textwidth]{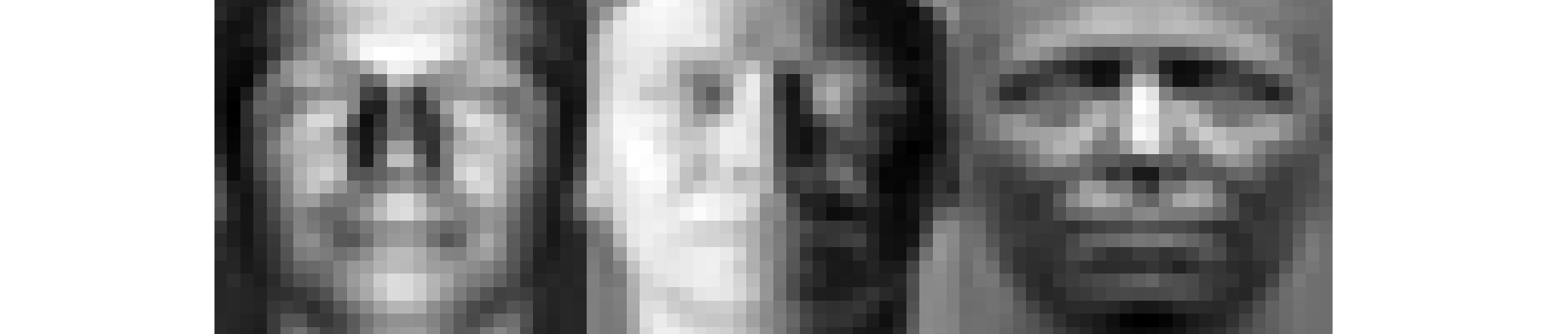}\\
         
         & & $D_G = 0.5766$ & $D_G = 0.4854$ \\
         \midrule
         
        \includegraphics[trim = {18cm, 13cm, 18cm, 6cm}, clip, width=0.11\textwidth]{figures/input_image_2.png} 
         &\includegraphics[trim = {14cm, 14cm, 14cm, 6cm}, clip, width=0.22\textwidth]{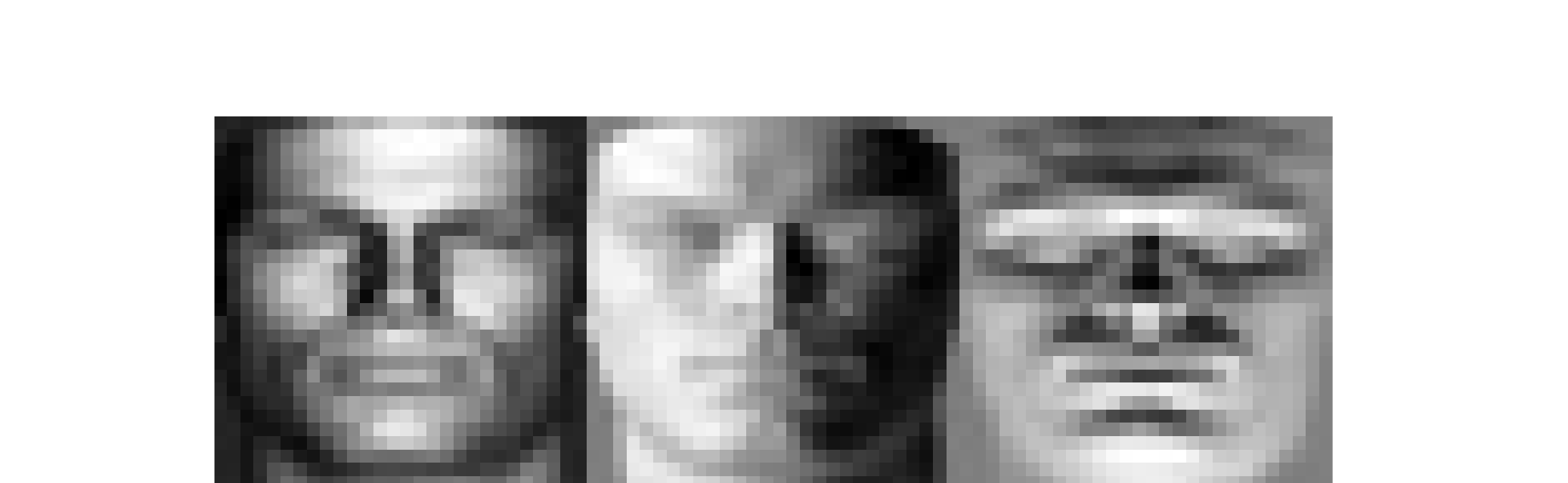}
         & \includegraphics[trim = {14cm, 14cm, 14cm, 8cm}, clip, width=0.22\textwidth]{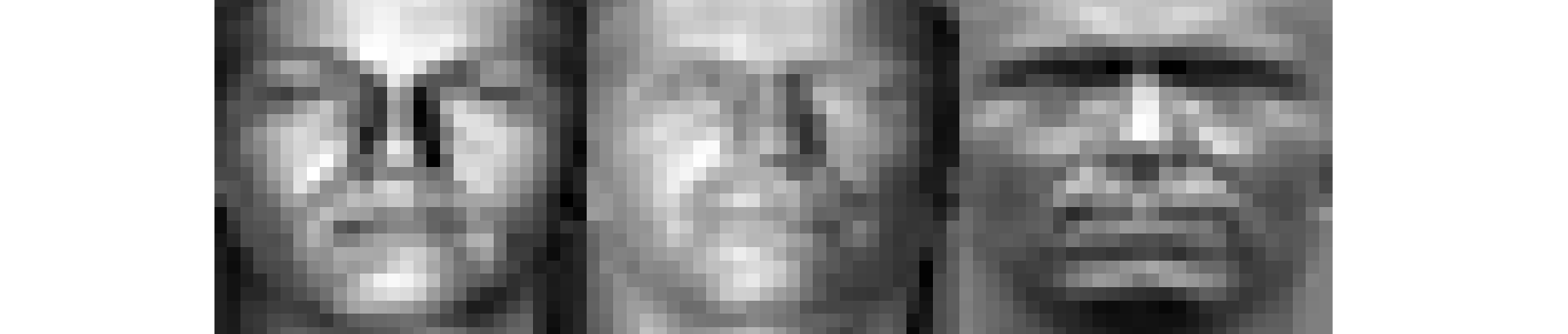}
         & \includegraphics[trim = {14cm, 14cm, 14cm, 8cm}, clip, width=0.22\textwidth]{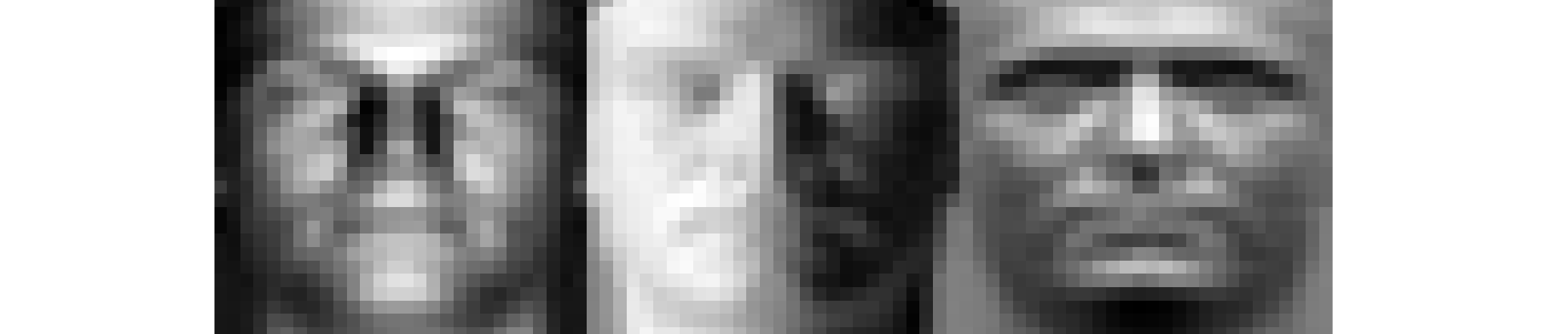}\\
         
         & & $D_G = 0.7095$ & $D_G = 0.4787$ \\
         \midrule
         
        \includegraphics[trim = {18cm, 13cm, 18cm, 6cm}, clip, width=0.11\textwidth]{figures/input_image_3.png}
         & \includegraphics[trim = {14cm, 14cm, 14cm, 6cm}, clip, width=0.22\textwidth]{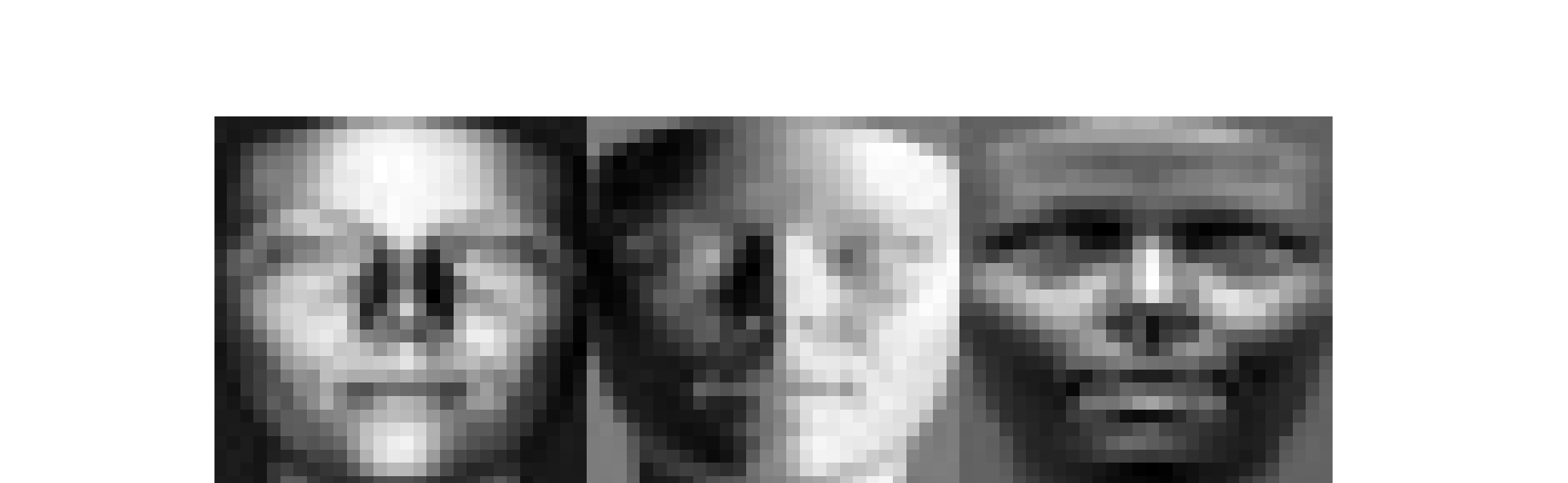}
         & \includegraphics[trim = {14cm, 14cm, 14cm, 8cm}, clip, width=0.22\textwidth]{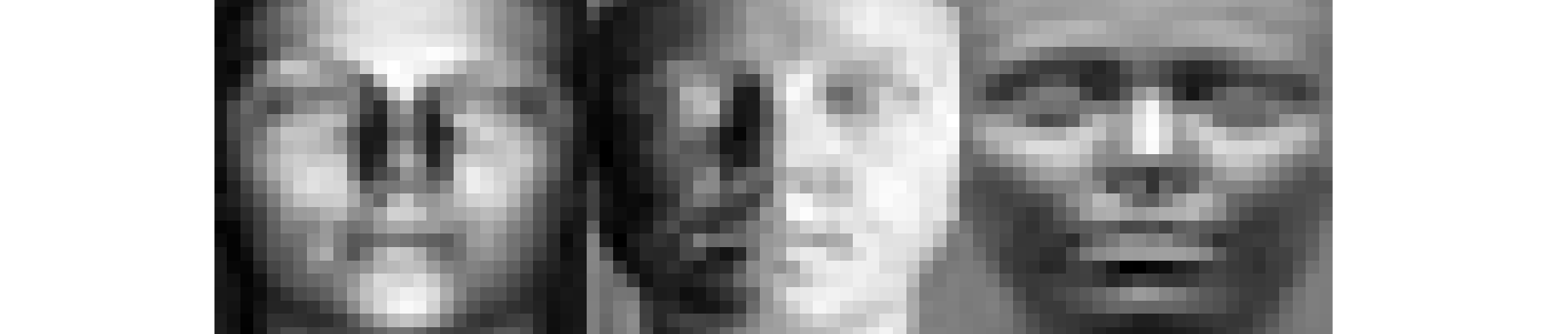}
         & \includegraphics[trim = {14cm, 14cm, 14cm, 8cm}, clip, width=0.22\textwidth]{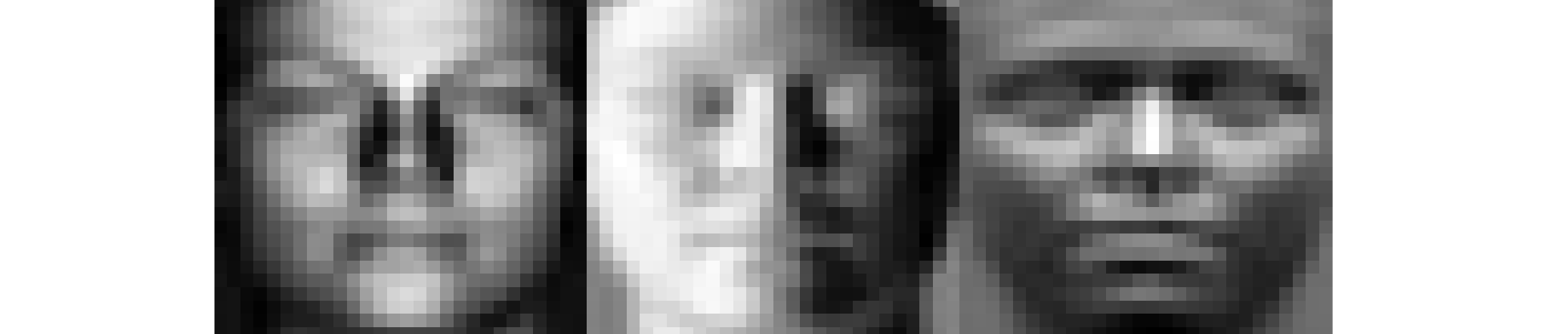}\\
         
         & & $D_G = 0.4429$ & $D_G = 0.4009$ \\
         \midrule
         
         \includegraphics[trim = {18cm, 13cm, 18cm, 6cm}, clip, width=0.11\textwidth]{figures/input_image_4.png}
         &\includegraphics[trim = {14cm, 14cm, 14cm, 6cm}, clip, width=0.22\textwidth]{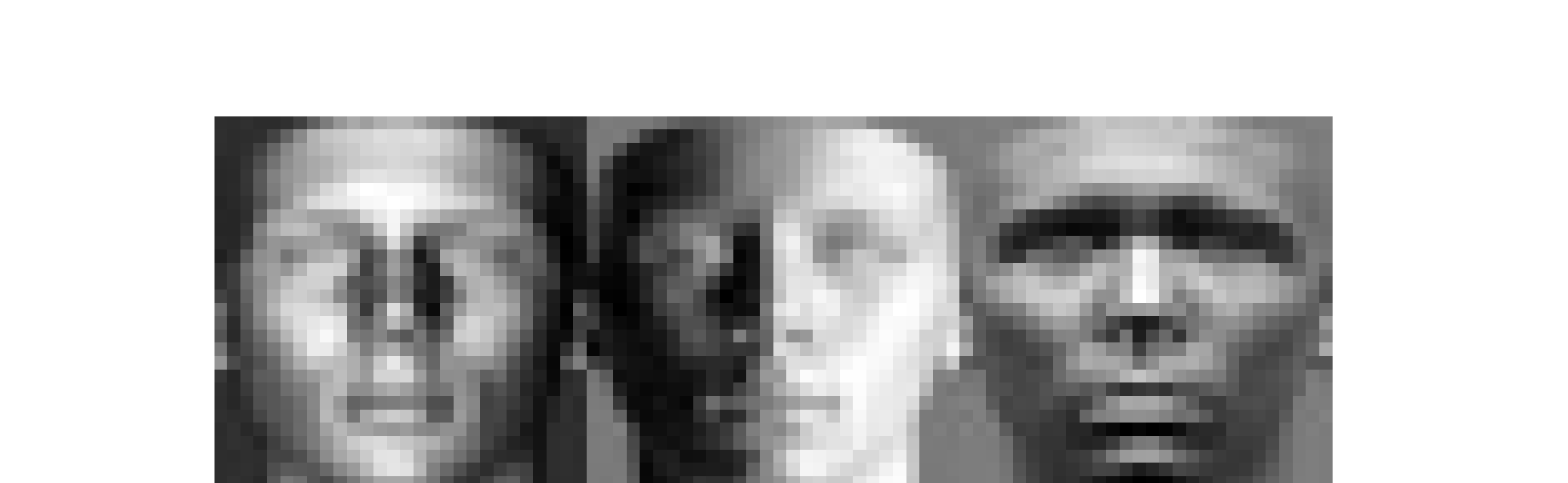}
         & \includegraphics[trim = {14cm, 14cm, 14cm, 8cm}, clip, width=0.22\textwidth]{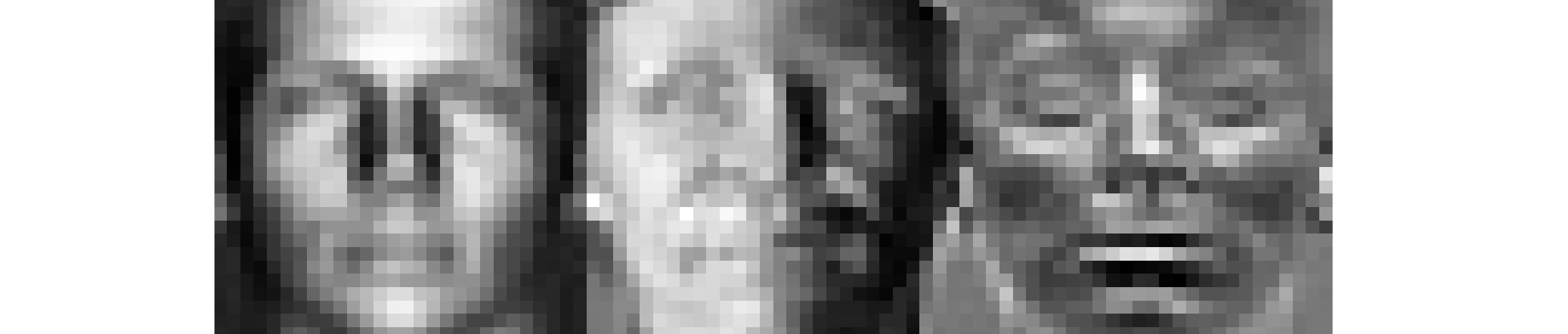}
         & \includegraphics[trim = {14cm, 14cm, 14cm, 8cm}, clip, width=0.22\textwidth]{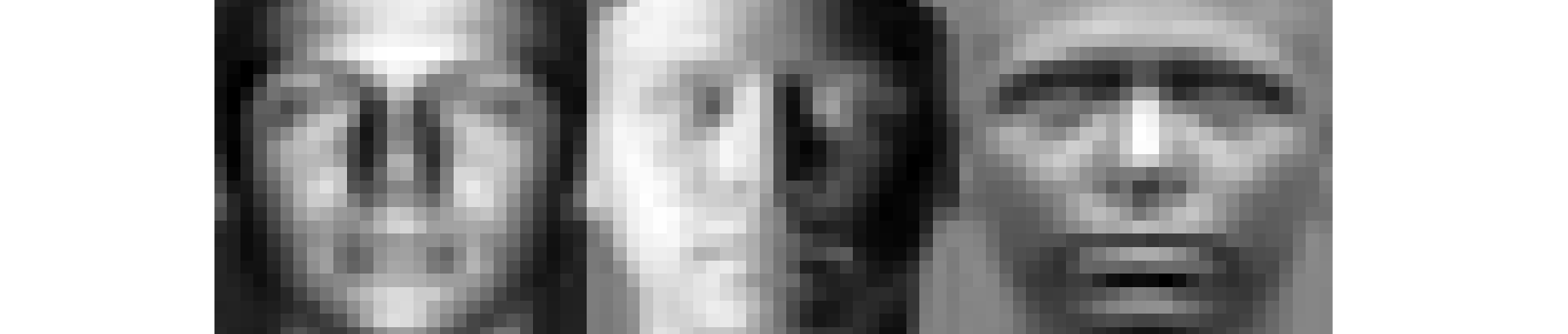}\\
         
         & & $D_G = 1.0904$ & $D_G = 0.3042$ \\
         \midrule
         
         \includegraphics[trim = {18cm, 13cm, 18cm, 6cm}, clip, width=0.11\textwidth]{figures/input_image_5.png}
         & \includegraphics[trim = {14cm, 14cm, 14cm, 6cm}, clip, width=0.22\textwidth]{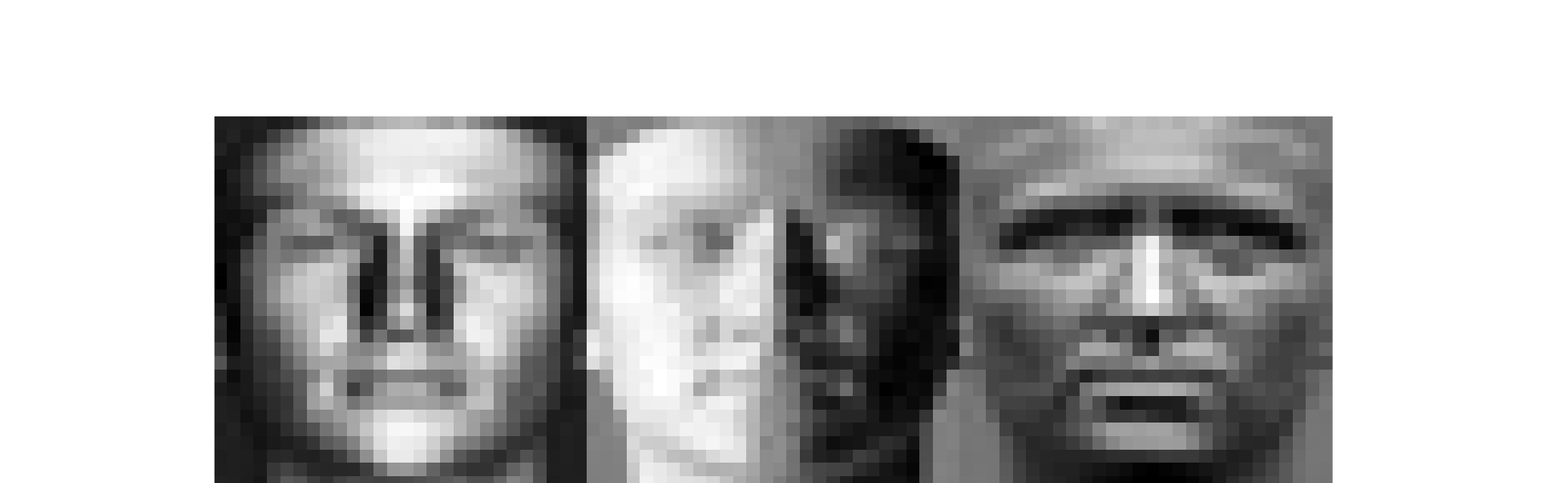}
         & \includegraphics[trim = {14cm, 14cm, 14cm, 8cm}, clip, width=0.22\textwidth]{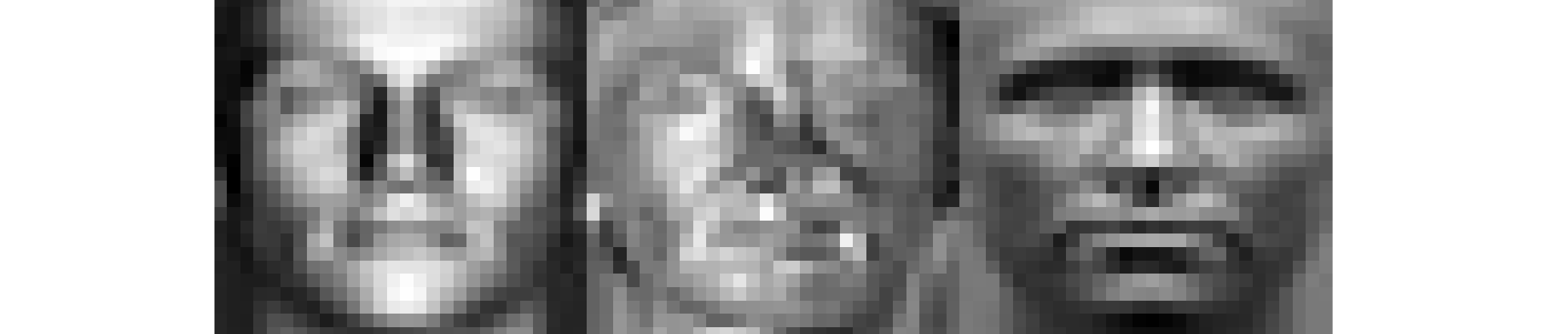}
         & \includegraphics[trim = {14cm, 14cm, 14cm, 8cm}, clip, width=0.22\textwidth]{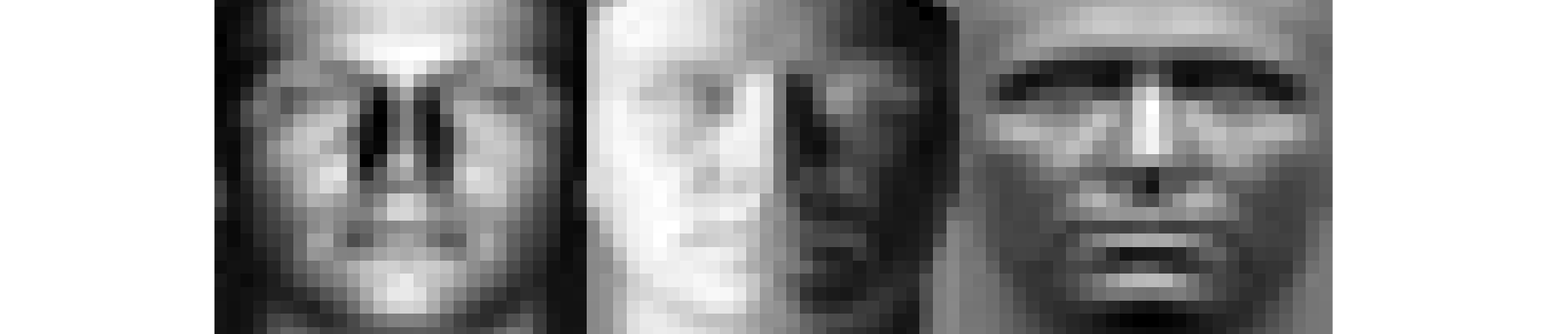}\\
         
         & & $D_G = 0.9012$ & $D_G = 0.3118$ \\
         \midrule

         \includegraphics[trim = {18cm, 13cm, 18cm, 6cm}, clip, width=0.11\textwidth]{figures/input_image_6.png}
         & \includegraphics[trim = {14cm, 14cm, 14cm, 6cm}, clip, width=0.22\textwidth]{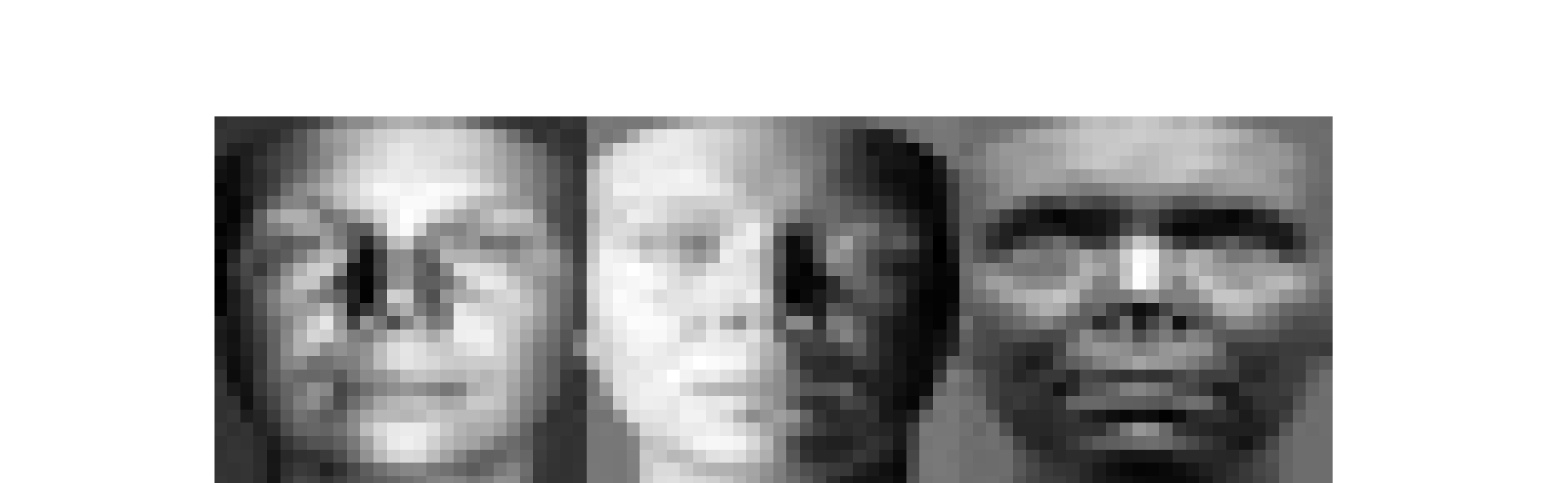}
         & \includegraphics[trim = {14cm, 14cm, 14cm, 8cm}, clip, width=0.22\textwidth]{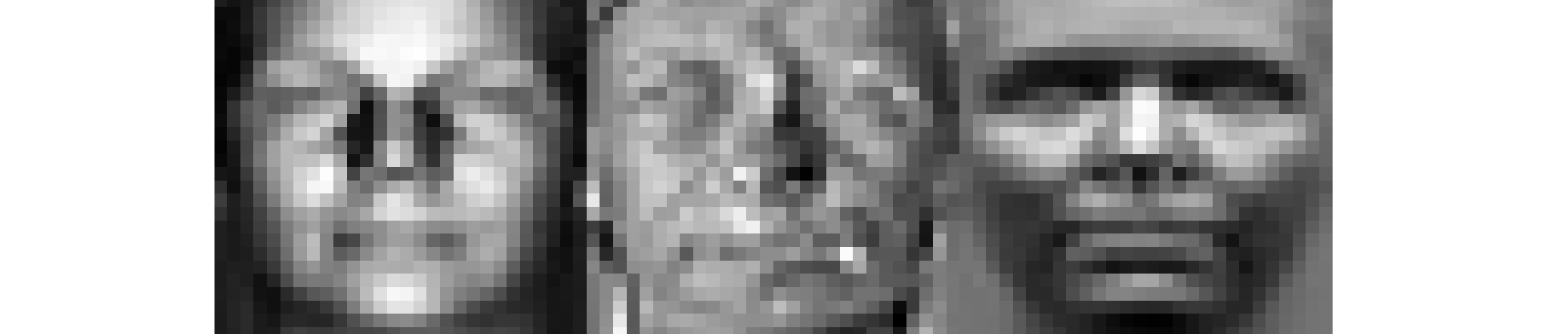}
         & \includegraphics[trim = {14cm, 14cm, 14cm, 8cm}, clip, width=0.22\textwidth]{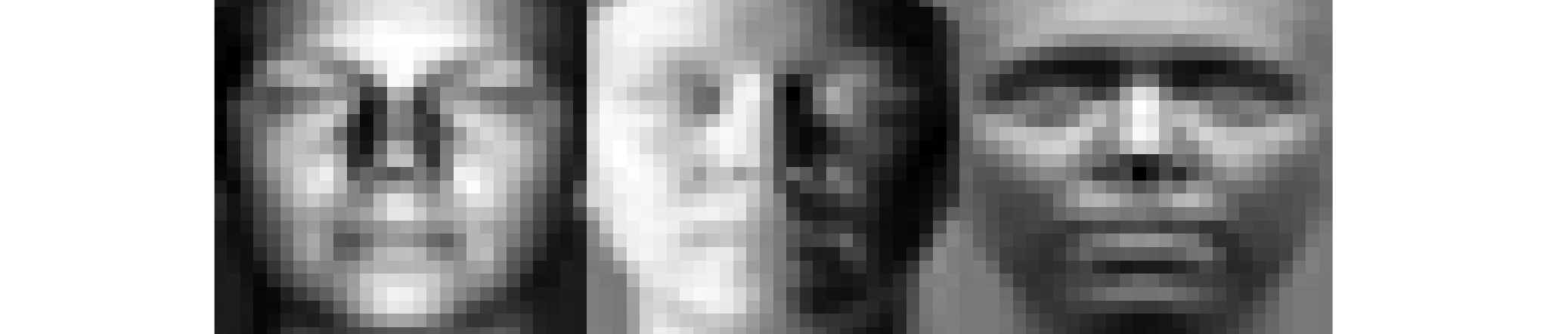}\\
         
         & & $D_G = 0.9244$ & $D_G = 0.3294$ \\
         
         \bottomrule
    \end{tabular}
    %\egroup
\end{table*}

As shown in Section \ref{sec:grassmann}, a tangent at some pole $p$ is given by the matrix $\mathbf{X}$, which in turn is completely specified by the matrix $\mathbf{A} \in \mathbb{R}^{(784-d) \times d}$, a much smaller matrix. Therefore, we design a network to map an input face $\mathbf{F}_i^j$ to the desired matrix $\mathbf{A_i}$. An training pair can be represented as $(\mathbf{F}_i^j, \mathbf{A}_i)$ and the let the output of the network be the vectorized version of a matrix $\hat{\mathbf{A}}_i \in \mathbb{R}^{(784-d) \times d}$. The $\mathbf{A}_i$'s are computed using the Grassmann logarithm map in \cite{srivastava2004bayesian}. We also note that $\mathbf{A}$ does not possess any additional structure to be enforced and thus a neural network can be easily trained to map to this space using just the Euclidean loss between $\mathbf{A}_i$ and the network output $\hat{\mathbf{A}}_i$: $L_G = ||\mathbf{A}_i - \hat{\mathbf{A}}_i||^2_F$.

{\bf Pole of tangent space:} This is a design choice and we conduct experiments with two different poles: \begin{enumerate*}[label=(\arabic*)] 
\item We compute the the illumination subspace of the entire training set. We will denote these PCs by $\mathbf{E}_{Tr}^k, k = 1,2, \dots, d$ and the corresponding matrix representation by $\mathbf{U}_{Tr}^d$, which forms the pole of the Grassmann tangent space. 
\item It is common practice to use the Fr\'{e}chet (also known as geometric or Karcher) mean as the pole of the tangent space. We compute the Fr\'{e}chet mean of the ground-truth subspaces of the training set using the iterative algorithm given by Turaga et al. \cite{turaga2011statistical}. We denote this pole as $\mathbf{U}_{Fr}^d$. 
\end{enumerate*} 

During the testing phase, using the output $\hat{\mathbf{A}}$ matrix, we employ the exponential map for a given pole to find the corresponding point on the Grassmann manifold. This framework of first mapping to the Grassmann tangent space using a network and then to the corresponding subspace using the Grassmann exponential map is referred to as \textbf{GrassmannNet-TS}. See supplementary material for details on visualizing the output Grassmann point.

\subsection{Experimental results on F2IS}\label{sec:experiments_f2is}
For the frameworks described in Sections \ref{sec:frameworks_f2is}, we describe the results on the test set of F2IS here. We compute the distance between predicted and ground-truth subspace as the measure to quantify the efficacy of the proposed frameworks. Various measures exist that quantify this notion based on principal angles between subspaces \cite{hamm2008grassmann}. We use the Grassmann geodesic distance. For two subspaces represented by $\mathbf{U}_1$ and $\mathbf{U}_2 \in \mathcal{G}_{n,d}$, the geodesic distance is given by $D_G(\mathbf{U}_1, \mathbf{U}_2) = \bigg(\sum_{i=1}^d\theta_i^2\bigg)^{\sfrac{1}{2}}$, where $\theta_i$'s are the principal angles obtained by the SVD of $\mathbf{U}_1^T\mathbf{U}_2 = \mathbf{W(\cos\Theta)\mathbf{V}^T}$, where $\cos\Theta = diag(\cos\theta_1,\dots,\cos\theta_d)$. We use the implementation in \cite{knyazev2002principal} for computing the principal angles. We report the arithmetic mean of this distance measure for the entire test set. Note that the maximum value of $D_G(.)$ is $\frac{\pi\sqrt{d}}{2}$. 

\begin{table}
	
	%\resizebox{\textwidth}{!}{%
	\centering
    %\bgroup
    %\def\arraystretch{1.5}
	\begin{tabular}{cccc}
		\hline
		
	    \multirow{2}{*}{\makecell{Subspace \\ Dim $d$}} & \multirow{2}{*}{Baseline} & \multicolumn{2}{c}{GrassmannNet-TS}\\
		\cline{3-4} 
		& & Pole = $\mathbf{U}_{Tr}^d$ & Pole = $\mathbf{U}_{Fr}^d$ \\
		\hline
		3 & 0.6613 & 0.3991 & 0.3953\\
		4 & 1.0997 & 0.5489 & 0.5913\\
		5 & 1.4558 & 0.8694 & 0.6174\\
		\hline
	
	\end{tabular}%}
	%\egroup
	\caption{Mean geodesic distance between predictions and ground-truth on the test set using the proposed frameworks. GrassmannNet-TS expectedly provides excellent results compared to the baseline framework for all subspace dimensions. Note that the max $D_G(.)$ possible for $d=3,4$ and $5$ are $2.72,3.14$ and $3.51$ respectively}
	\label{table:f2is_results}
\end{table}

The results based on the mean subspace distance on the test set for the proposed frameworks for different values of the subspace dimension $d$ are presented in Table \ref{table:f2is_results}. The baseline, as expected, performs poorly. This is because during the training phase, the network received conflicting ground-truth information because of the permutation and sign flips inherently present in the data. On the other hand, GrassmannNet-TS yields excellent performance as it is invariant to these transformations by design. We reiterate that this is possible only in the case of regression directly on the Grassmann tangent space since mapping to the Grassmann manifold is infeasible because of the very large number of variables required to represent the projection matrix. The outputs of the baseline as well as GrassmannNet-TS using the Fr\'{e}chet mean as the pole are also presented visually in Tables \ref{table:f2is_results_5} and \ref{table:f2is_results_4} for two test images for $d=4,5$ respectively, and show similar trends. The choice of the pole does not seem to affect the results significantly except in the case of $d=5$ where the Fr\'{e}chet mean performs better. Additional results are shown in the Supplementary Material.

%$$$$$$$$$$$$$$$$$$$$$$$$$$$$$$$$$$$$$$$$$$$$$$$$$$$$$$$$$$$$$$$$$$$$$

\section{Deep regression on the unit hypersphere for multi-class classification}

\paragraph{Reformulating classification as mapping to the unit hypersphere:}
For multi-class classification problems, deep networks usually output a probability distribution, where one uses the mode of the distribution to predict the class label. What ensures that the output elements form a probability distribution is the "softmax layer". However, by using a square-root parametrization -- replacing each element in the distribution by its square-root -- we can map a probability distribution to a point on the non-negative orthant of a unit hypersphere $\mathcal{S}^C$, where $C$ is now the number of classes. The square-root parameterization reduces the complicated Riemannian metric on the space of probability density functions, the Fisher-Rao metric, to the simpler Euclidean inner product on the tangent space of the unit hypersphere with closed form expressions for the geodesic distance, exponential and logarithm maps \cite{srivastava2007riemannian}. In this work, equipped with the knowledge of differential geometry of the sphere, we propose different loss functions for the tackling the classification problem. We consider two main variants -- learning a network to map to the sphere directly or map to its tangent space. We note that the constraint for a point to be on a sphere or to be a probability distribution is simple and can be easily satisfied by using an appropriate normalization (dividing by its 2-norm or using softmax). However, mapping to the tangent space of the sphere provides a novel perspective to the same problem and is more general since as we showed earlier, it is necessary for the Grassmannian. 

Consider a classification problem with $C$ classes. For a given input vector $\mathbf{x}$, let the ground-truth probability distribution over the class labels be $\mathbf{c}_{pd}$. The corresponding point on $\mathcal{S}^C$ is given by $\mathbf{c}_S$, such that $\mathbf{c}_S(i) = \sqrt{\mathbf{c}_{pd}(i)}, i = 1 \dots C$. The pole $\mathbf{u}_S$ for constructing the tangent space $T_{\mathbf{u}_S}\mathcal{S}^C$ is chosen to be the point on $\mathcal{S}^C$ corresponding to the uniform distribution $\mathbf{u}_{pd}, \mathbf{u}_{pd}(i) = \frac{1}{C}, i = 1 \dots C$. Let $\mathbf{\xi}$ be the desired point on $T_{\mathbf{u}_S}\mathcal{S}^C$ for the input $\mathbf{x}$ and is given by output of the log map $\mathbf{\xi} = \exp^{-1}_{\mathbf{u}_S}(\mathbf{c}_S)$. \textbf{Let the output of the last fully connected layer be denoted by $\hat{\mathbf{o}}$.}

\vspace{-0.1in}
\paragraph{Geometry of the unit hypersphere \cite{absil2009optimization}:} The the $n$-dimensional unit sphere denoted as $\mathcal{S}^n$ and is defined as $\mathcal{S}^n = \{(x_1, x_2,\dots,x_{n+1}) \in \mathbb{R}^{n+1} |\quad \sum_{i=1}^{n+1} x_i^2 = 1\}$. Given any two points $\mathbf{x},\mathbf{y} \in \mathcal{S}^n$, the geodesic distance between $\mathbf{x}$ and $\mathbf{y}$ is calculated using $d(\mathbf{x},\mathbf{y}) = \cos^{-1}\langle\mathbf{x},\mathbf{y}\rangle \label{eq:sphere_geodesic}$. For a given point $\mathbf{x} \in \mathcal{S}^n$, the tangent space of $\mathcal{S}^n$ at $\mathbf{x}$ is given by $ T_\mathbf{x}\mathcal{S}^n = \{\mathbf{\xi} \in \mathbb{R}^n |\quad \mathbf{x}^T\mathbf{\xi} = 0\}$. Since the Riemannian metric (the inner product on the tangent space) is the usual Euclidean inner product, the distance function on the tangent space induced by this inner product is the Euclidean distance. The exponential map $\exp: T_\mathbf{x}\mathcal{S}^n \to \mathcal{S}^n$ is computed using the following formula: $ \exp_\mathbf{x}\mathbf{\xi} =  \cos(||\mathbf{\xi}||)\mathbf{x} + \sin(||\mathbf{\xi}||)\frac{\mathbf{\xi}}{||\mathbf{\xi}||},$ where $\mathbf{\xi} \in T_\mathbf{x}\mathcal{S}^n$. For $\mathbf{x},\mathbf{y} \in \mathcal{S}^n$, the inverse exponential map $\exp^{-1}: \mathcal{S}^n \to T_\mathbf{x}\mathcal{S}^n$ is given by $\exp_\mathbf{x}^{-1}(\mathbf{y}) = \frac{d(\mathbf{x}, \mathbf{y})}{||P_\mathbf{x}(\mathbf{y} - \mathbf{x})||}P_\mathbf{x}(\mathbf{y} - \mathbf{x})$. $P_\mathbf{x}(\mathbf{v})$ is the projection of a vector $\mathbf{v} \in \mathbb{R}^n$ onto $T_\mathbf{x}\mathcal{S}^n$, given by $P_\mathbf{x}(\mathbf{v}) = (\mathbf{I}_n - \mathbf{xx}^T)\mathbf{v}$, $\mathbf{I}_n$ is the $n \times n$ identity matrix.

\subsection{Mapping to the hypersphere directly: SNet-M}
In this case, the network directly outputs points on the sphere and a training pair is represented as $(\mathbf{x}, \mathbf{c}_S)$. We call this framework Snet-M. At test time, the network outputs a point on the sphere and the corresponding probability distribution is computed by squaring the elements of the output. We propose the following loss functions on the sphere. While training, the loss is averaged over the entire batch. In this case, we also employ a normalizing layer as the last layer of the network which guarantees that $\hat{\mathbf{o}}$ lies on $\mathcal{S}^C$. \begin{enumerate}[label=(\arabic*)]
    \item \textbf{Euclidean loss on} $\mathcal{S}^C$ : This simply measures the Euclidean distance between two points on a sphere and does not take into account the non-linear nature of the manifold: $L_{S_{euc}} = ||\mathbf{c}_S - \frac{\hat{\mathbf{o}}}{||\hat{\mathbf{o}}||_2}||^2_2$.
    \item \textbf{Geodesic loss on} $\mathcal{S}^C$ : The ``true" distance between the two points on the sphere is given by $\theta = \cos^{-1}\langle\mathbf{c}_S, \frac{\hat{\mathbf{o}}}{||\hat{\mathbf{o}}||_2}\rangle$. Since minimizing this function directly leads to numerical difficulties, we instead minimize its surrogate, $L_{S_{geo}} = 1 - \cos\theta$.
\end{enumerate}

\begin{table*}[]
	%\resizebox{\textwidth}{!}{%
	\centering
	%\bgroup
    %\def\arraystretch{1.5}
    
	\begin{tabular}{ccccc}
		\hline
		Framework & \makecell{Desired Output of\\Network Lies on} & Loss Function & \makecell{Test Accuracy\\on MNIST (\%)} & \makecell{Test Accuracy \\on CIFAR-10 (\%)}\\
		
		\hline
		Baseline & & Cross Entropy & 99.224 (0.0306) & 78.685 (0.3493)\\
		
		\hline
		\multirow{2}{*}{SNet-M} & \multirow{2}{*}{$\mathcal{S}^C$} & $L_{S_{euc}}$  & 99.263 (0.0479) & 79.738 (0.4009) \\
		& & $L_{S_{geo}}$ & 99.293 (0.0343) & \textbf{80.024} (0.5131)\\
		\hline
		\multirow{3}{*}{SNet-TS} & \multirow{3}{*}{$T_{\mathbf{u_S}}\mathcal{S}^C$} & $L_{T_{euc}}$ & 99.293 (0.0691) & 77.548 (0.5620)\\
		& & $L_{T_{orth}}$ & 99.279 (0.0448) & 77.708 (0.3517)\\
		& & $L_{T_{proj}}$ & \textbf{99.332} (0.0600) & 76.047 (1.6225)\\
		\hline
	\end{tabular}%}
	%\egroup
	\caption{Avg test accuracy (std. dev.) over 10 runs using different loss functions on $S^C$ and $T_{\mathbf{u}_S}S^C$, compared to the cross entropy loss.}\label{table:sphere_results}
\end{table*}

\subsection{Mapping to the hypersphere via its tangent space: SNet-TS}
Here, given an input, the algorithm first produces an intermediate output on $T_{\mathbf{u_S}}\mathcal{S}^C$ and then the exponential map is used to compute the desired point on $\mathcal{S}^C$. The corresponding probability distribution is computed by simply squaring each element of the vector. We refer to this framework as SNet-TS. A training example, then, is of the form $(\mathbf{x}, \mathbf{\xi})$, where $\mathbf{\xi}$ is the desired tangent vector. We propose the following loss functions on $T_{\mathbf{u_S}}\mathcal{S}^C$. \begin{enumerate}[label=(\arabic*)]
    \item \textbf{Euclidean loss on} $T_{\mathbf{u}_S}\mathcal{S}^C$: Measures the Euclidean distance between two points on the tangent space of the sphere : $L_{T_{euc}} = ||\mathbf{\xi} - \hat{\mathbf{o}}||^2_2$. The output, $\hat{\mathbf{o}}$, is however not guaranteed to lie on $T_{\mathbf{u_S}}\mathcal{S}^C$ since a point on $T_{\mathbf{u_S}}S^C$ needs to satisfy the constraint $\mathbf{x}^T\mathbf{\xi} = 0$. Therefore, we first project $\hat{\mathbf{o}}$ to the $T_{\mathbf{u_S}}S^C$ and then use the exponential map.
    
    \item \textbf{Euclidean + Orthogonal loss on} $T_{\mathbf{u}_S}\mathcal{S}^C$: In order to improve the "tangentness" of the output vector, we add the inner product loss that encourages the orthogonality of the output vector relative to the pole, which is the tangent space constraint: $L_{T_{orth}} = ||\mathbf{\xi} - \hat{\mathbf{o}}||^2_2 + \lambda( \hat{\mathbf{o}}^T\mathbf{u}_S)^2$.

    \item \textbf{Projection loss on} $T_{\mathbf{u}_S}\mathcal{S}^C$: Since a closed form expression exists to project an arbitrary vector onto $T_{\mathbf{u}_S}\mathcal{S}^C$, we implement the projection layer as the last layer that guarantees that the output of the projection layer lies on $T_{\mathbf{u}_S}\mathcal{S}^C$. We compute the Euclidean loss between the projected vector and the desired tangent: $L_{T_{proj}} = ||\mathbf{c}_S - P_{\mathbf{u}_S}\mathbf{\hat{o}}||^2_2$.
\end{enumerate}

\subsection{Experiments with image classification}
Image classification problem is a widely studied problem in computer vision and will serve as an example to demonstrate training a network to map to points on a unit hypersphere and its tangent space. We now describe the experiments conducted using MNIST and CIFAR-10 datasets. We train 6 networks with different loss functions. The first network is a baseline using the softmax layer to output a probability distribution directly and employs the well-known cross-entropy loss. The next two networks use the SNet-M framework and $L_{S_{euc}}$ and $L_{S_{geo}}$ as the loss functions. The desired outputs in this case lie on $\mathcal{S}^C$ and the network employs a normalizing layer at the end in order force the output vector to lie on the $\mathcal{S}^C$. The ground-truth output vectors are obtained by using the square-root parametrization. The final 3 networks employ the SNet-TS framework and $L_{T_{euc}}$, $L_{T_{orth}}$ and $L_{T_{proj}}$ as the loss functions. The desired output vector, in this case, should lie on $T_{\mathbf{u}_S}\mathcal{S}^C$. The required logarithm and exponential maps are computed using the Manifold Optimization toolbox \cite{boumal2014manopt}. We note that the purpose of the experiments is to show that for some chosen network architecture, the proposed loss functions that are inspired by the geometry of the hypersphere, perform comparably with the cross-entropy loss function.

\vspace{-0.15in}
\paragraph{MNIST:} The MNIST dataset \cite{lecun1998gradient} consists a total of 60000 images of hand-written digits (0-9). Each image is of size $28 \times 28$ and is in grayscale. The task is to classify each image into one of the 10 classes (0-9). The dataset is split into training and testing sets with 50000 and 10000 images respectively. We use the LeNet-5 architecture as the neural network \cite{lecun1998gradient}. The network consists of 2 convolutional and max-pooling (\texttt{conv}) layers followed by 2 fully-connected (\texttt{fc}) layers. ReLU non-linearity is employed. The filters are of size $5 \times 5$. The first and second \texttt{conv} layers produce 32 and 64 feature maps respectively. The first and second fc layers output 1024 and 10 elements respectively. The networks are trained for 50000 iterations with a batch size of 100 using Adam optimizer \cite{kingma2015adam} with learning rate of $10^{-3}$. 

\vspace{-0.15in}
\paragraph{CIFAR-10:} The CIFAR-10 dataset \cite{krizhevsky2009learning} consists a total of 60000 RGB natural images. Each image is of size $32 \times 32$. The task is to classify each image into one of the 10 classes (Airplane, Automobile, Bird, Cat, Deer, Dog, Frog, Horse, Ship, Truck). The dataset is split into training and testing sets with 50000 and 10000 images respectively. The network consists of 2 \texttt{conv} layers with max-pooling and local response normalization followed by 2 \texttt{fc} layers. ReLU non-linearity is employed. Each input image is mean subtracted and divided by its standard deviation. Data augmentation using 10 $24 \times 24$ random crops per input image is employed to reduce overfitting. At test time, the central $24 \times 24$ region is used as the input to the network, after performing the same normalization as the training inputs. The networks are trained for 500000 iterations with a batch size of 100 using Adam optimizer with learning rate of $10^{-3}$.

For each dataset, we fix the network architecture and train the 6 versions of the network with different loss function as described above. We use $\lambda = 1$ for $L_{T_{orth}}$. The image recognition accuracies obtained on the test set (averaged over 10 runs) are shown in Table \ref{table:sphere_results}.

The results indeed show that some of the proposed loss functions tend to perform better than cross entropy. For both datasets and especially CIFAR-10, SNet-M yields better performance than cross entropy and within this framework, geodesic loss performs better compared to Euclidean loss. SNet-TS shows improvements in accuracy in the case of MNIST, albeit with higher variance in the accuracy.

%$$$$$$$$$$$$$$$$$$$$$$$$$$$$$$$$$$$$$$$$$$$$$$$$$$$$$$$$$$$$$$$$$$$$$
\section{Conclusion}
In this paper, we have studied the problem of learning invariant representations, which are at the heart of many computer vision problems, where invariance to physical factors such as illumination, pose, etc often lead to representations with non-Euclidean geometric properties. %And advances in deep learning do not exploit the knowledge of these geometric constraints and rely purely on data.
We have shown how deep learning architectures can be effectively extended to such non-linear target domains, exploiting the knowledge of data geometry. Through two specific examples -- predicting illumination invariant representations which lie on the Grassmannian, and multi-class classification by mapping to a scale-invariant unit hypersphere representation -- we have demonstrated how the power of deep networks can be leveraged and enhanced by making informed choices about the loss function while also enforcing the required output geometric constraints exactly. Extensions to other geometrically constrained representations, such as symmetric positive-definite matrices are evident. On the theoretical side, extending the current framework to applications where data points may have wider spread from their centroid, and to non-differentiable manifolds which arise in vision remain interesting avenues for the future.

\section*{Acknowledgements}
This work was supported in part by ARO grant number W911NF-17-1-0293 and NSF CAREER award 1451263. We thank Qiao Wang and Rushil Anirudh for helpful discussions. 

\section*{Appendix 1: Concepts from differential geometry}
In this section, we will define some terms from differential geometry that are necessary for the rest of the paper. For a more comprehensive treatment of matrix manifolds considered here, refer Absil et al.\cite{absil2009optimization} and Edelman et al.\cite{edelman1998geometry}.

\textbf{Manifold}: A manifold is a topological space that is locally Euclidean i.e., at every point on the manifold $p \in \mathcal{M}$, there exists an open neighborhood $H$ around $p$, and a mapping $\phi$ such that $\phi(H)$ is an open subset of $\mathbb{R}^n$ where $\phi$ is a diffeomorphism. A differentiable manifold is a manifold that has a differentiable structure associated with it. A smooth manifold can be defined similarly.

\textbf{Tangent and tangent space}: At every point on a differentiable manifold, a linear/vector space, called the tangent space, of the same dimension as the manifold can be constructed. Consider a point $p \in \mathcal{M}$. Consider a curve $\alpha(t)$ on the manifold passing through $p$ such that $\alpha(0)=p$. The derivative of this curve at $p$, $\alpha'(0)$, is the velocity vector, also called the tangent. If one considers all possible curves through this point $\{\alpha_i(t)\}, i=1,2,\dots$, then the set of all velocity vectors $\{\alpha_i'(0)\}$ is the tangent space $T_p\mathcal{M}$, at this point. The point at which the tangents are computed is called the pole of the tangent space.  

\textbf{Riemannian metric}: A Riemannian metric is function that smoothly associates, to each point $p \in \mathcal{M}$, an inner product on the tangent space $T_p\mathcal{M}$. A smooth manifold equipped with a Riemannian metric is called a Riemannian manifold.

\textbf{Geodesic}: Consider a curve on the manifold $\gamma: [a,b] \to \mathcal{M}$ such that $\gamma(a) = x$ and $\gamma(b) = y$. The curve that minimizes the functional $E = \int_a^b ||\gamma'(t)||^2\,dt$  is called the geodesic and locally minimizes the path length between two points on the manifold. The norm $||.||$ is induced by the Riemannian metric at $\gamma(t)$.

\textbf{Exponential Map}: Given that a unique geodesic $\gamma(t)$ exists locally at $p \in \mathcal{M}$ and $\gamma(0) = p$ and $\gamma'(0)=v \in T_p\mathcal{M}$, the exponential map at $p$ is the function $\exp_p: T_p\mathcal{M} \to \mathcal{M}$ given by $\exp_p(v) = \gamma(1)$. For a neighborhood, $U \subset T_p\mathcal{M}$ containing $0$, it can be shown that $\exp_p$ is a diffeomorphism, i.e., it has an inverse which is also continuous. The algorithm for computing the exponential map depends both on the manifold of interest and the pole of the tangent space. 

\textbf{Logarithm Map}: At least in the neighborhood $U \subset T_p\mathcal{M}$ containing $0$, the exponential map has an inverse called the logarithm map $\exp^{-1}_p: \mathcal{M} \to T_p\mathcal{M}$. This also points to the fact that the pole is an important design choice. The algorithm for computing the logarithm map depends both on the manifold of interest and the pole of the tangent space. 

\section*{Appendix 2: Additional information on face $\to$ illumination subspace}
\subsection*{Illumination directions used for network inputs} \label{sec:illum_directions}
The 33 illumination directions used for creating inputs for both the training and test sets are shown in the left column of Table \ref{table:illumination_directions} and are a subset of the illumination directions used in the Extended Yale Face Database B \cite{georghiades2001few}.

\begin{table}
    \centering
    \begin{tabular}{c|c||c|c}
        Azimuth & Elevation & Azimuth & Elevation \\
        \hline
        0 & 0 & 0 & 90 \\
        0 & -20 & -35 & 65 \\
        0 & 20 & 35 & 65 \\
        0 & -35 & -50 & -40 \\
        0 & 45 & 50 & -40 \\
        -5 & -5 & -60 & -20 \\
        -5 & 5 & 60 & -20 \\
        5 & -5 & -70 & -35 \\
        5 & 5 & -70 & 0 \\
        -10 & 0 & -70 & 45 \\
        -10 & -20 & 70 & -35 \\
        10 & 0 & 70 & 0 \\
        10 & 20 & 70 & 45 \\
        15 & 20 & -85 & -20 \\
        -15 & 20 & -85 & 20 \\
        -20 & -10 & 85 & -20 \\
        -20 & 10 & 85 & 20 \\
        -20 & -40 & -95 & 0 \\
        20 & -10 & 95 & 0\\
        20 & 10 & -110 & -20\\
        20 & -40 & -110 & 15\\
        25 & 0 & -110 & 40\\
        -25 & 0 & -110 & 65\\
        -35 & -20 & 110 & -20\\
        -35 & 15 & 110 & 15\\
        -35 & 40 & 110 & 40\\
        35 & -20 & 110 & 65\\
        35 & 15 & -120 & 0\\
        35 & 40 & 120 & 0\\
        -50 & 0 & -130 & 20\\
        50 & 0 & 130 & 20\\
        -60 & 20 & & \\
        60 & 20 & & \\
        
    \end{tabular}
    \caption{Illumination directions (azimuth and elevation in degrees) used to generate illumination subspaces. As network inputs for both training and test sets, we only use the 33 illuminations shown in the left column of the table.}
    \label{table:illumination_directions}
\end{table}

\subsection*{Visualizing output of GrassmannNet-TS}\label{sec:visualizing_grassmann}
\begin{figure}
	\begin{minipage}[c][10cm][t]{0.4\textwidth}
  \centering
  \includegraphics[trim = {14cm, 14cm, 14cm, 8cm}, clip, width=\textwidth]{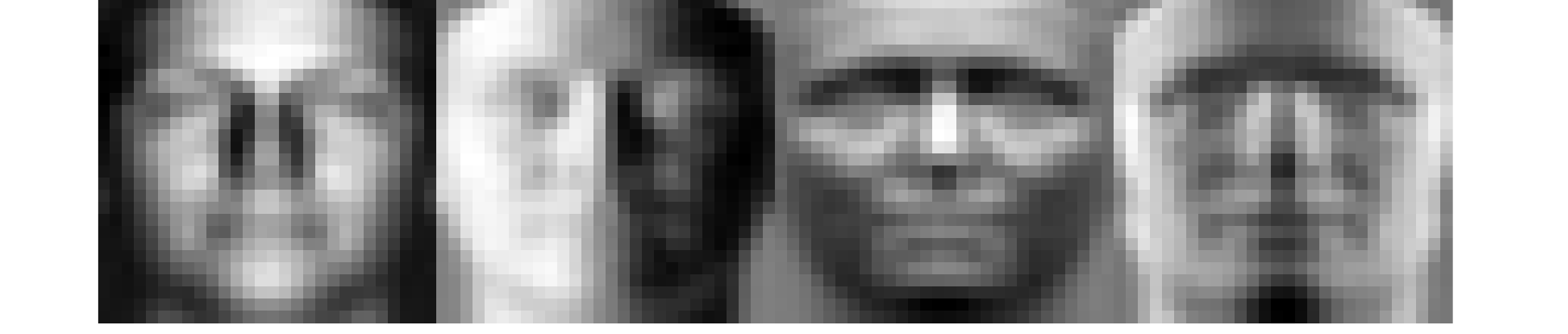}
  \subcaption{}
  \par
  \includegraphics[trim = {14cm, 14cm, 14cm, 8cm}, clip, width=\textwidth]{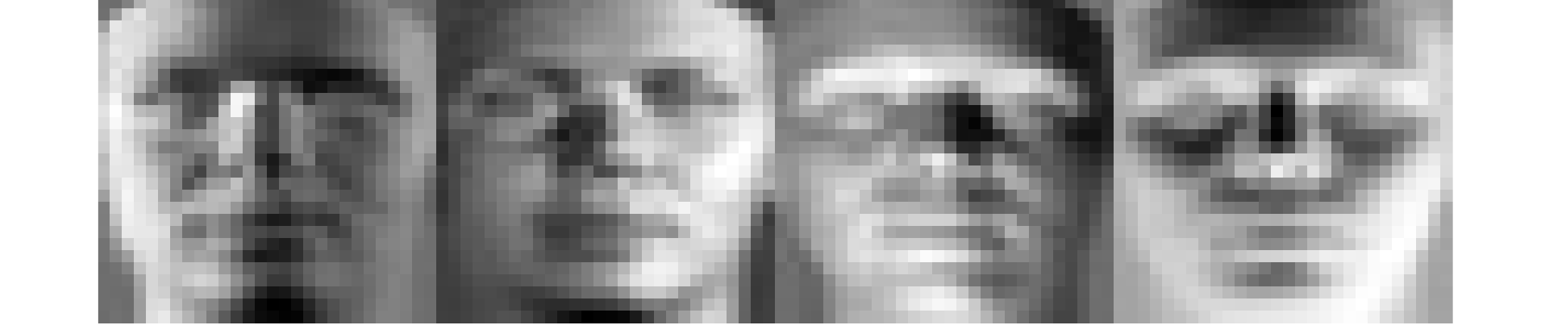}
  \subcaption{}
  \par
  \includegraphics[trim = {14cm, 14cm, 14cm, 8cm}, clip, width=\textwidth]{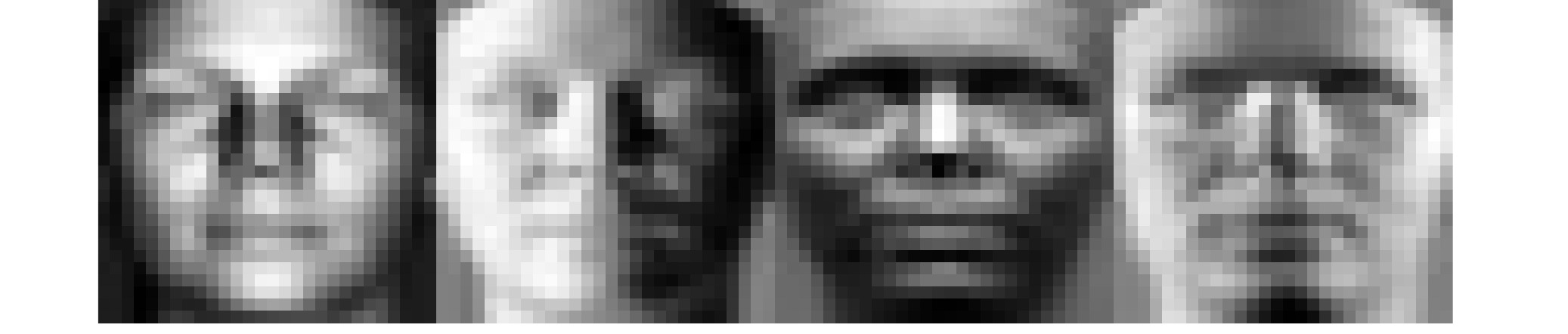}
  \subcaption{}
  \end{minipage}
  \vspace{-1.75in}
\caption{Images illustrating F2IS for $d$ = 4 (a) Top 4 PCs of the training set i.e., $\mathbf{U}_{Tr}^4$. (b) Output of exponential map for GrassmannNet-TS. (c) Output of GrassmannNet-TS after rotation by $\mathbf{Q}^*$.}
\label{fig:f2is_illustration}
\end{figure}

Due to the invariance provided by the Grassmann manifold, the exponential map of the tangent need not return the same point $\mathbf{U}$ whose columns are the principal components (PCs) of the illumination subspace, it is only guaranteed to return a point $\mathbf{R}$ whose columns spans the same subspace as the columns of $\mathbf{U}$. This means that we cannot use the columns of $\mathbf{R}$ immediately for visual comparison with the ground-truth PCs of the subspace. Therefore we first find an orthogonal matrix $\mathbf{Q}^*$ such that $\mathbf{Q}^* = \underset{\mathbf{Q}}\argmin ||\mathbf{U}_{avg} - \mathbf{RQ}||_F$ and then use $\hat{\mathbf{R}}=\mathbf{RQ}^*$ as the new point that can be used for visualization. $\mathbf{Q}^*$ has a simple closed form expression given by $\mathbf{Q}^* = \mathbf{WY}^T$, where $\mathbf{\mathbf{R}^T\mathbf{U}_{avg}} = \mathbf{W\Sigma Y^T}$ is the singular value decomposition (SVD) of $\mathbf{R}^T\mathbf{P}_{avg}$ \cite{golub2012matrix}. See Figure \ref{fig:f2is_illustration} for visual illustration. 

\subsection*{More results on F2IS}
Table \ref{table:f2is_histograms} shows the histograms of the subspace distances obtained on the test set using the two proposed frameworks. These results demonstrate that GrassmannNet-TS performs much better than the baseline.

\begin{table*}[]
    \centering
    
    \begin{tabular}{cccc}
     
        \toprule
        \makecell{Subspace\\ Dimension} & Baseline & GrassmannNet-TS & \makecell{All pairs of training \\and test subspaces} \\
        \midrule
        3 & \raisebox{-.5\height}{\includegraphics[trim = {2cm, 8cm, 4cm, 9cm}, clip, width=0.22\textwidth]{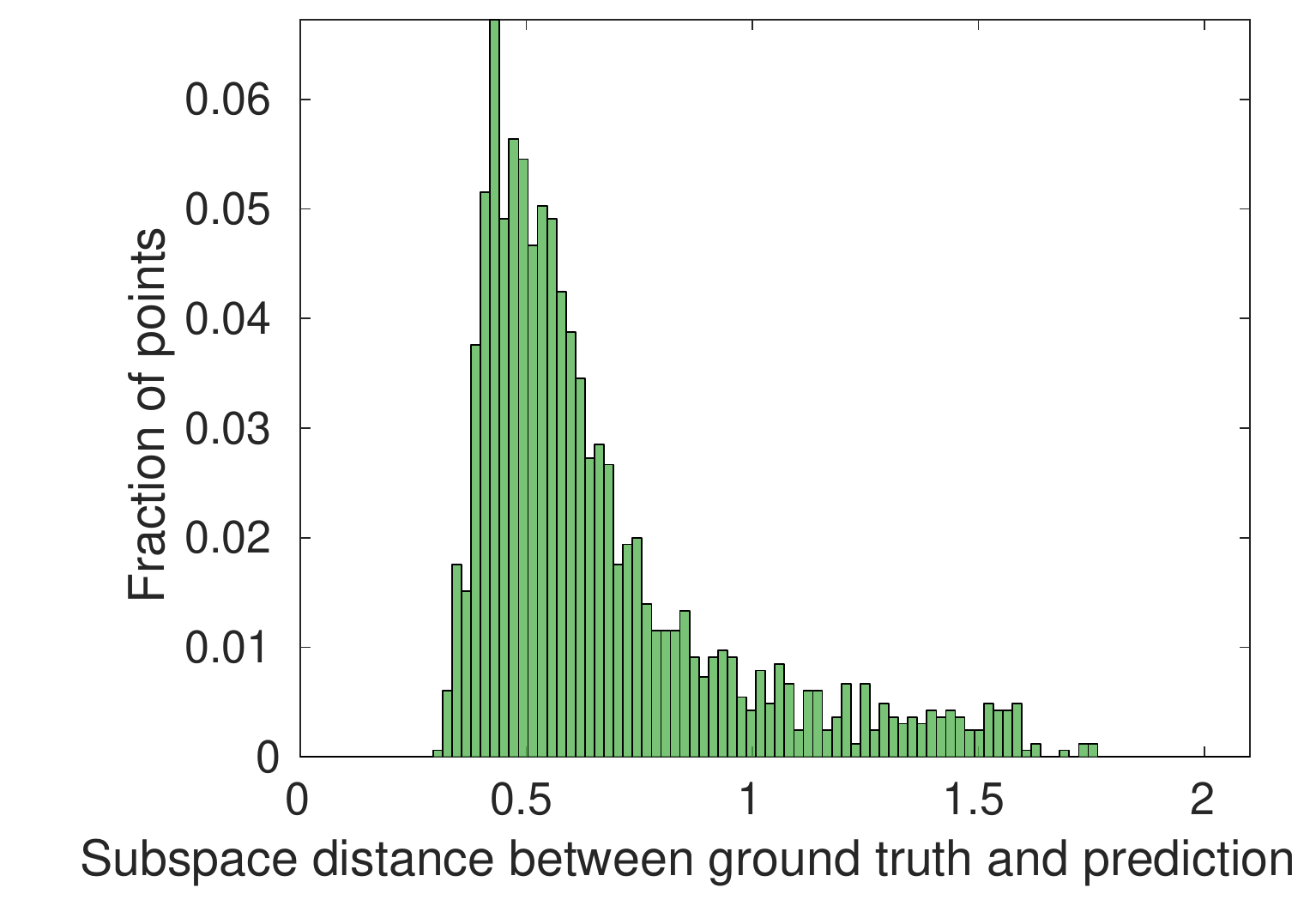}}
         & \raisebox{-.5\height}{\includegraphics[trim = {2cm, 8cm, 4cm, 9cm}, clip, width=0.22\textwidth]{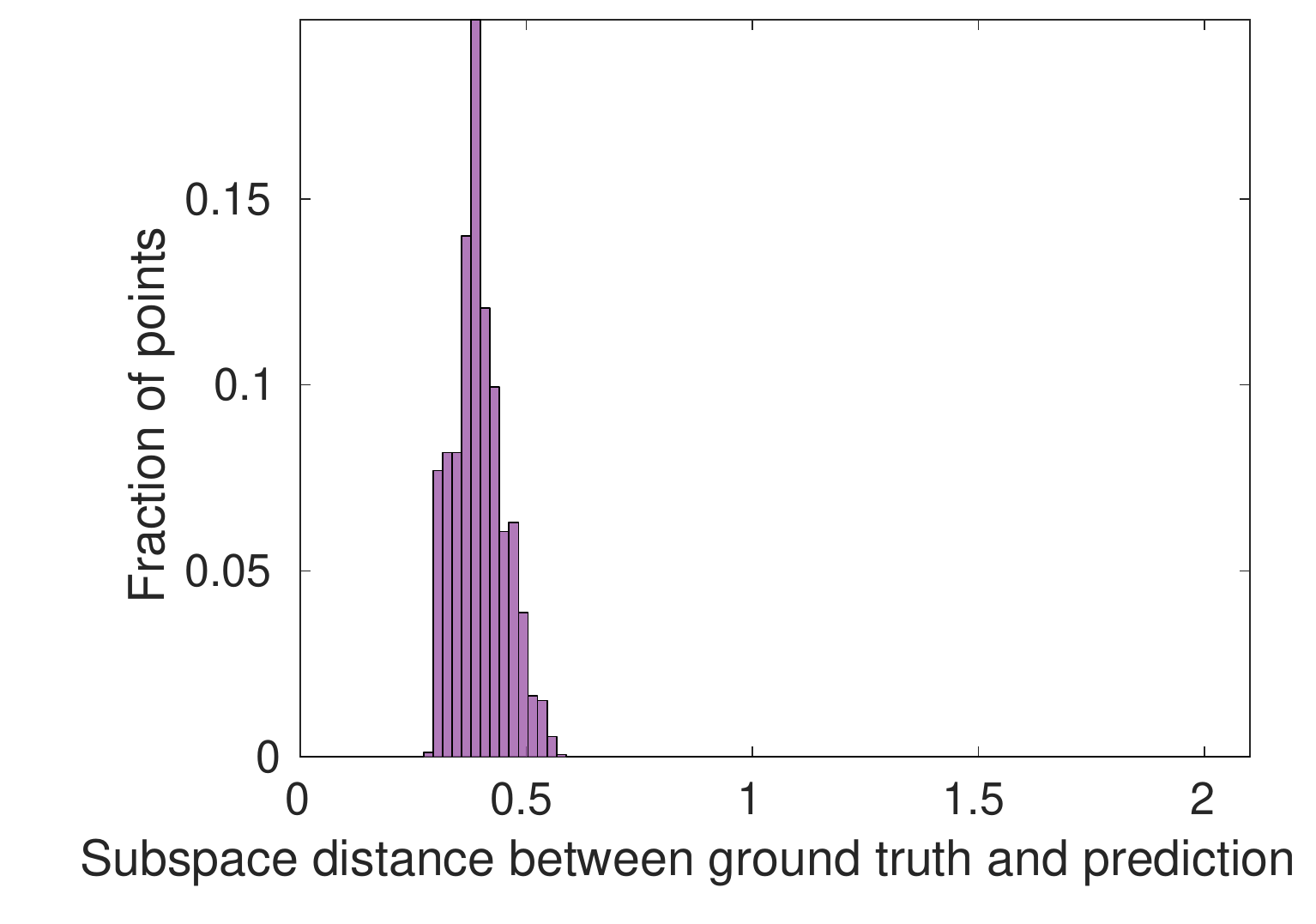}} 
         & \raisebox{-.5\height}{\includegraphics[trim = {2cm, 8cm, 4cm, 9cm}, clip, width=0.22\textwidth]{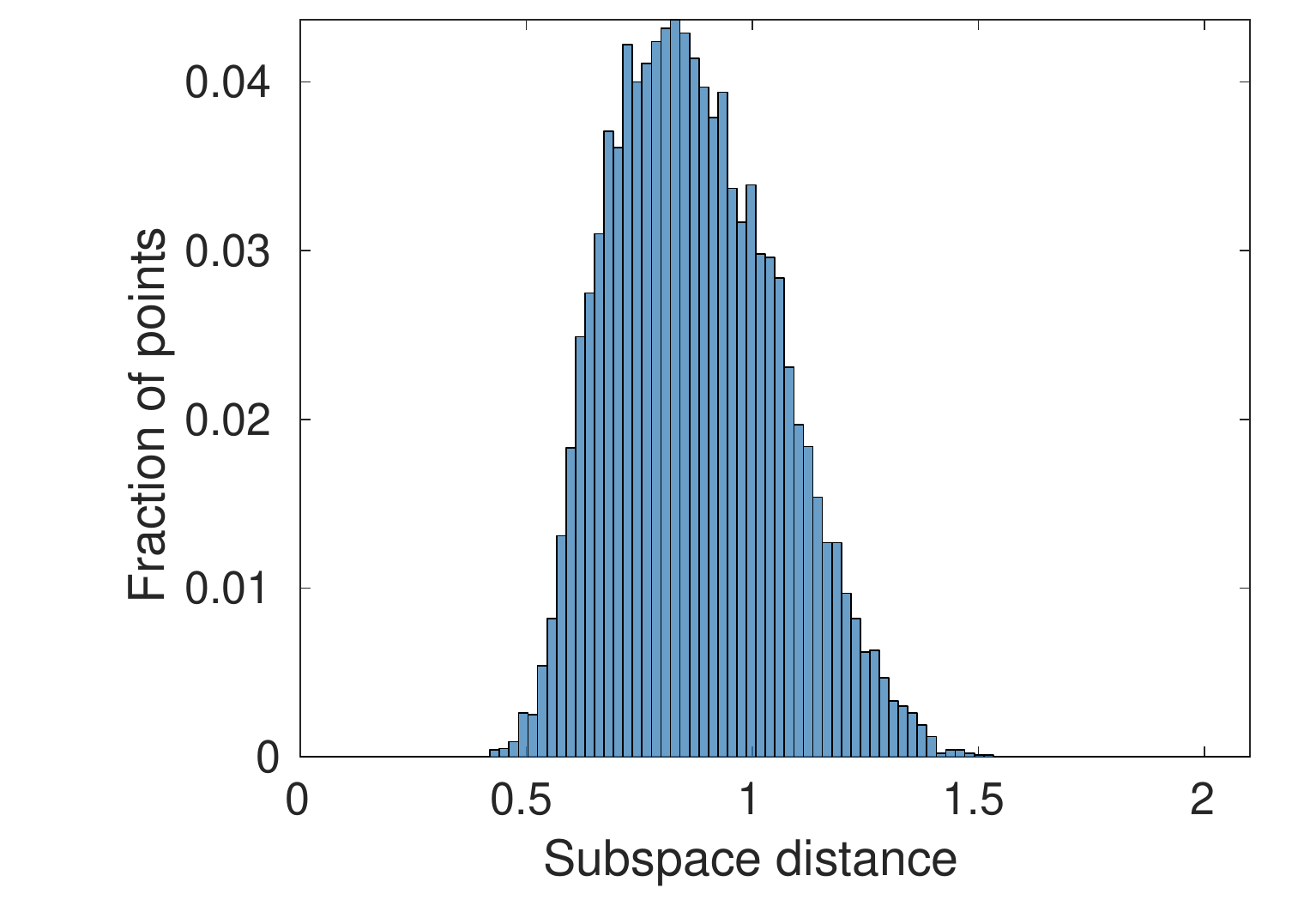}}\\

         4 & \raisebox{-.5\height}{\includegraphics[trim = {2cm, 8cm, 4cm, 9cm}, clip, width=0.22\textwidth]{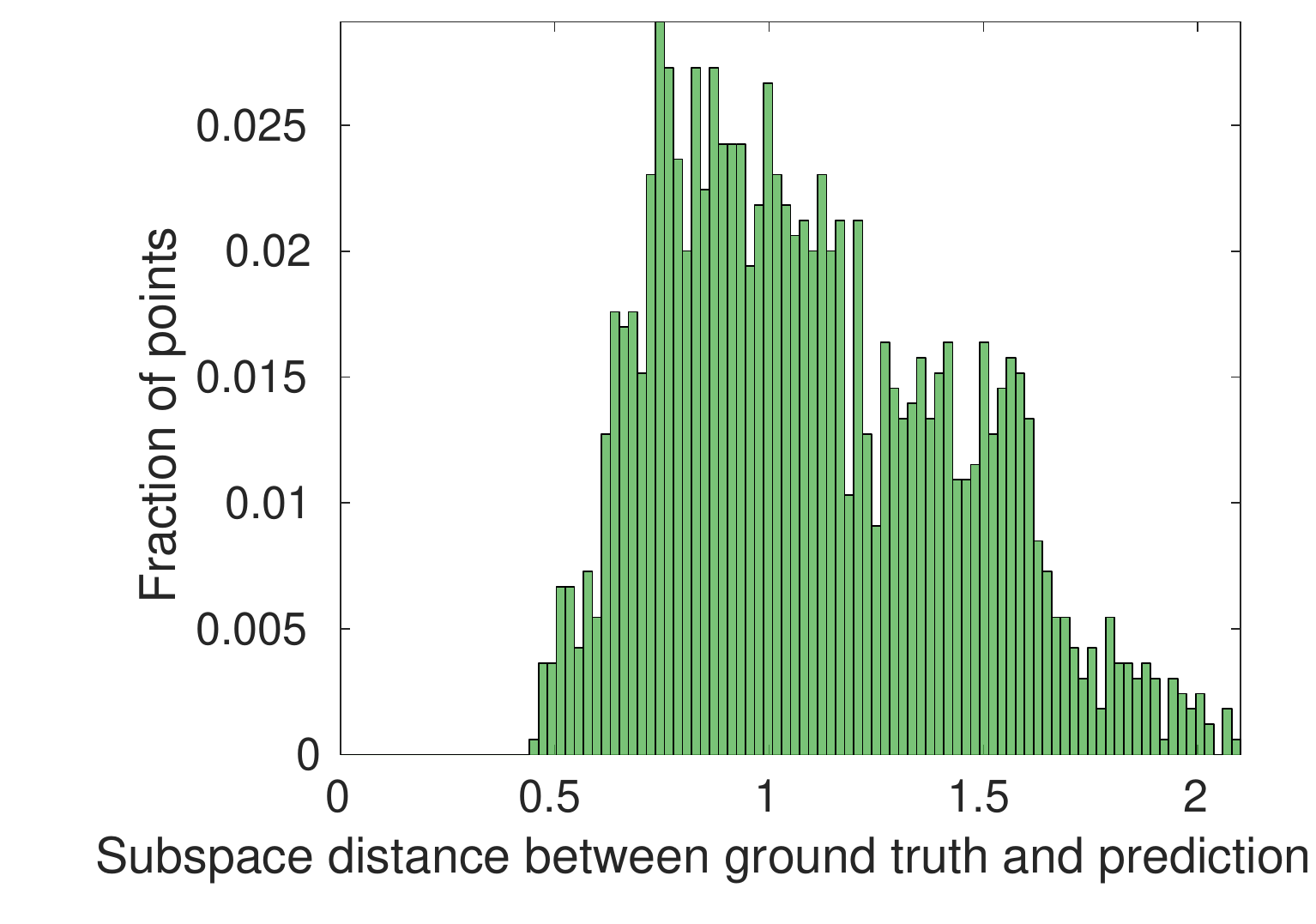}} 
         & \raisebox{-.5\height}{\includegraphics[trim = {2cm, 8cm, 4cm, 9cm}, clip, width=0.22\textwidth]{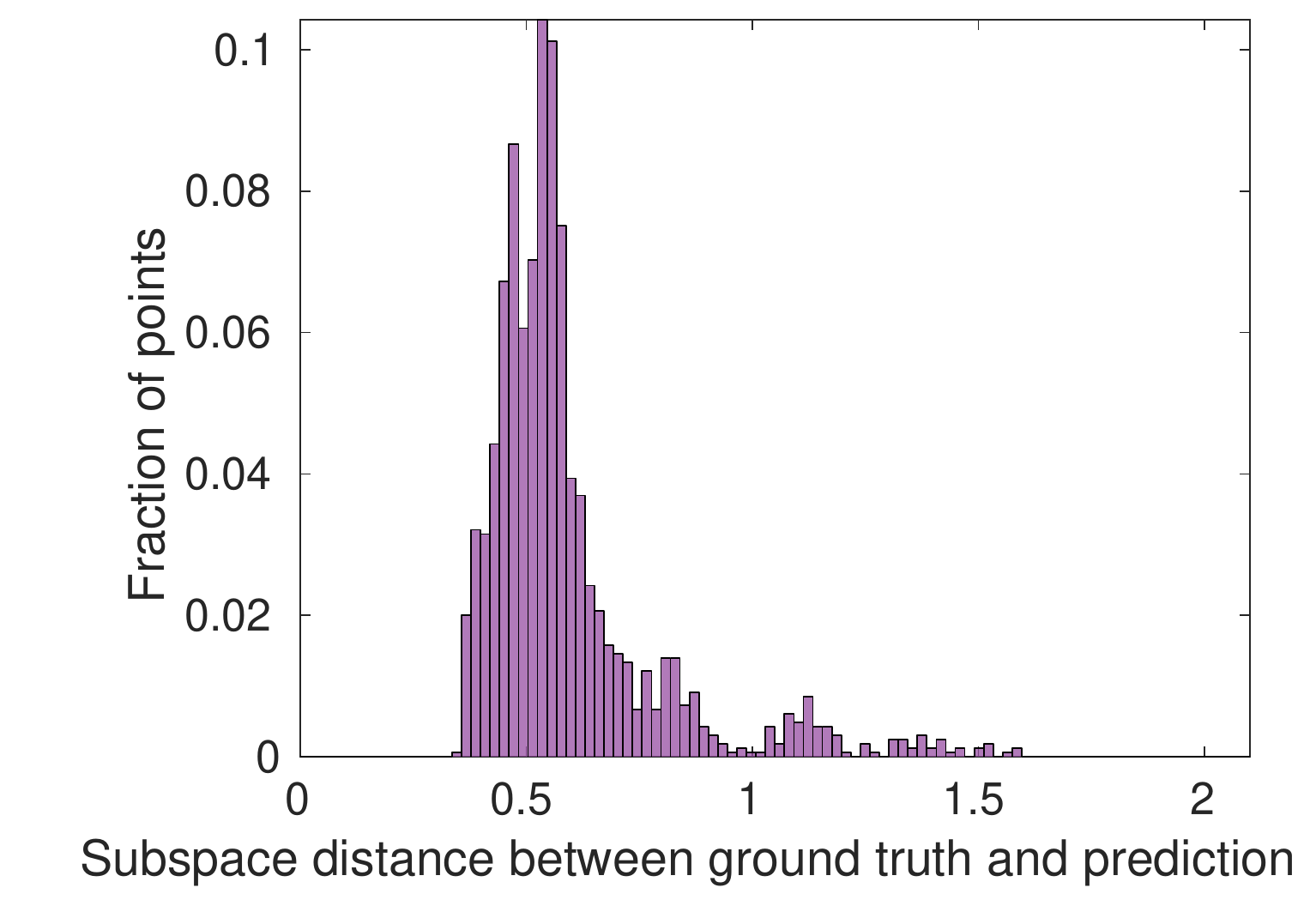}} 
         & \raisebox{-.5\height}{\includegraphics[trim = {2cm, 8cm, 4cm, 9cm}, clip, width=0.22\textwidth]{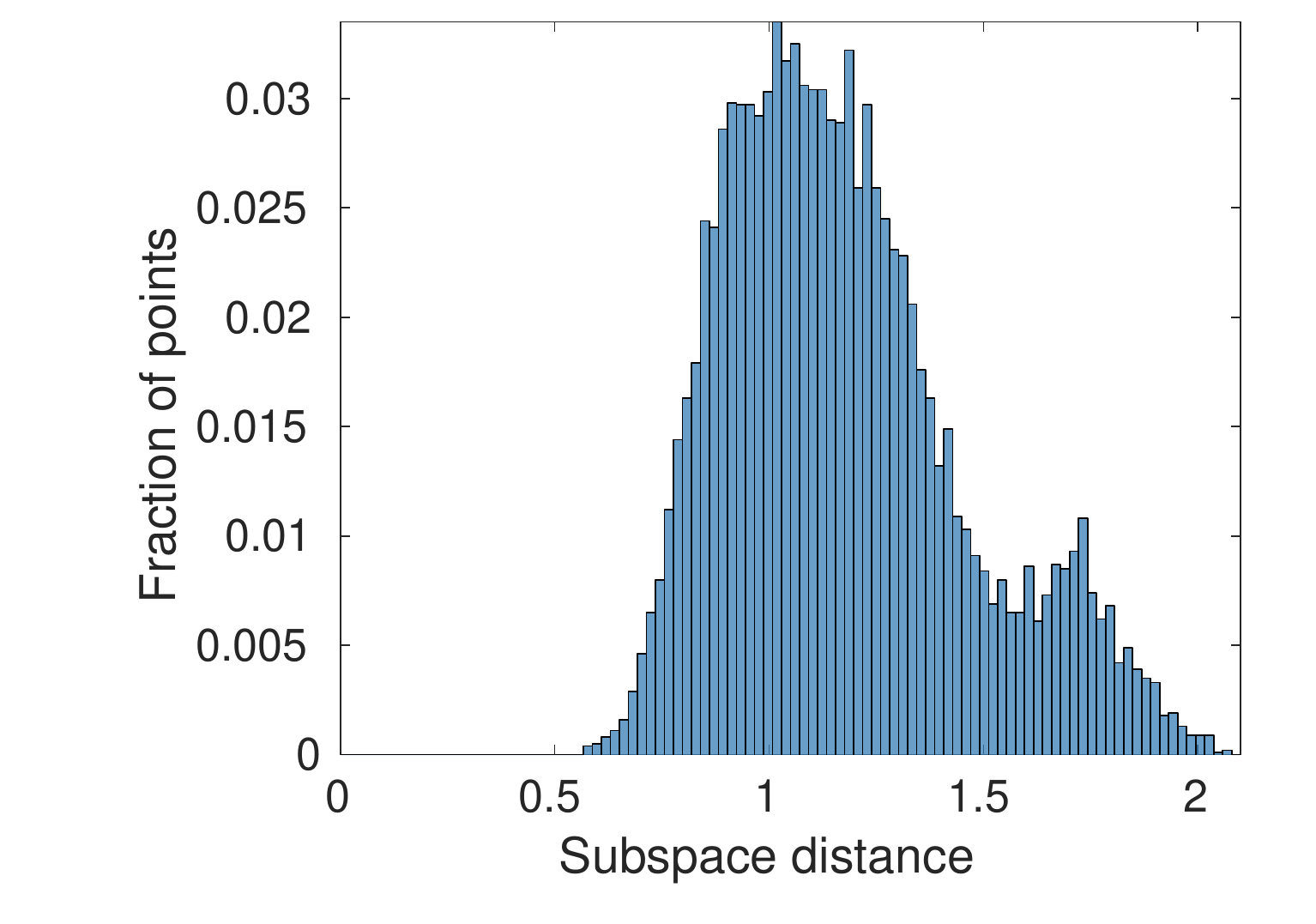}}\\

         5 & \raisebox{-.5\height}{\includegraphics[trim = {2cm, 8cm, 4cm, 9cm}, clip, width=0.22\textwidth]{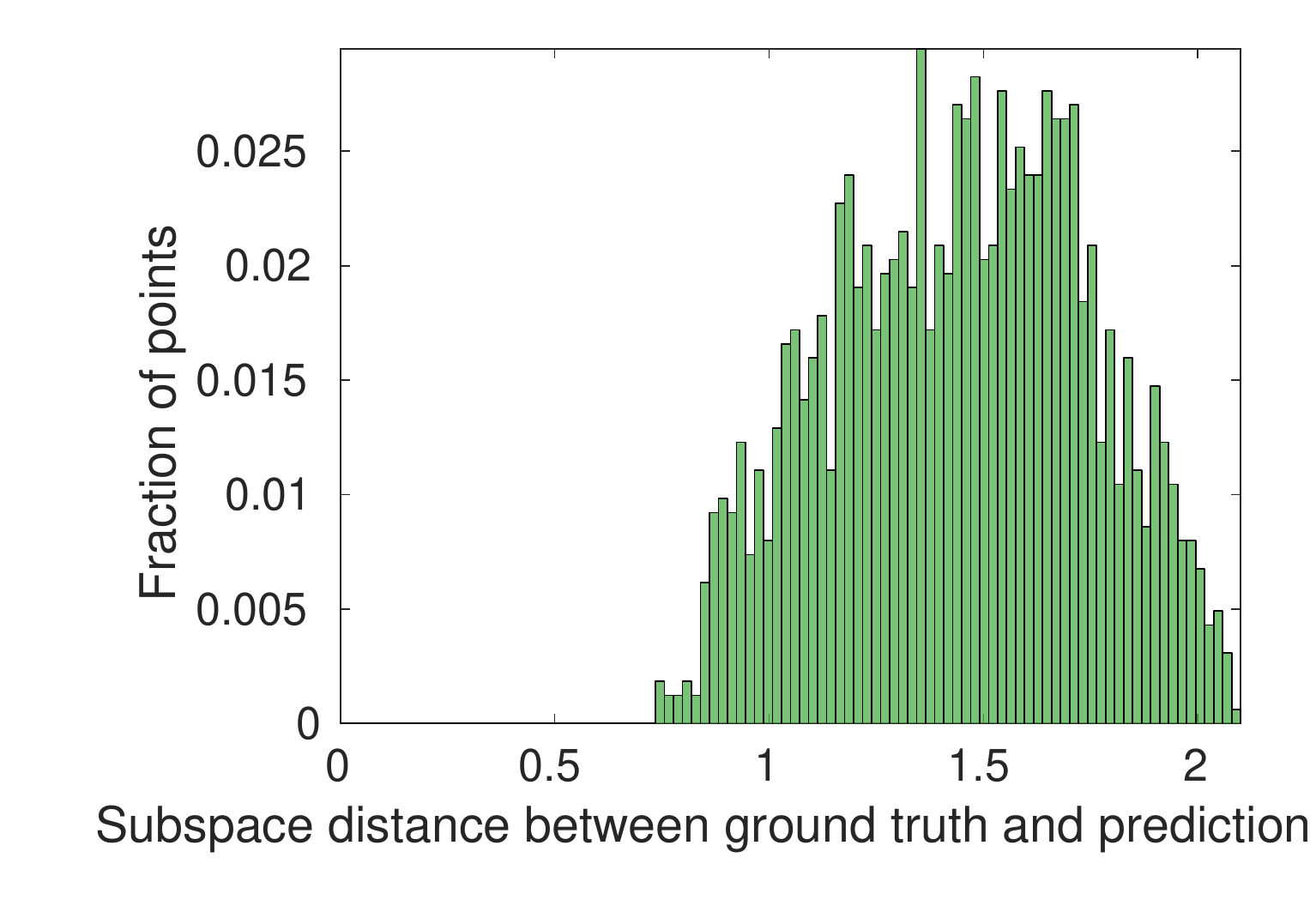}} 
         & \raisebox{-.5\height}{\includegraphics[trim = {2cm, 8cm, 4cm, 9cm}, clip, width=0.22\textwidth]{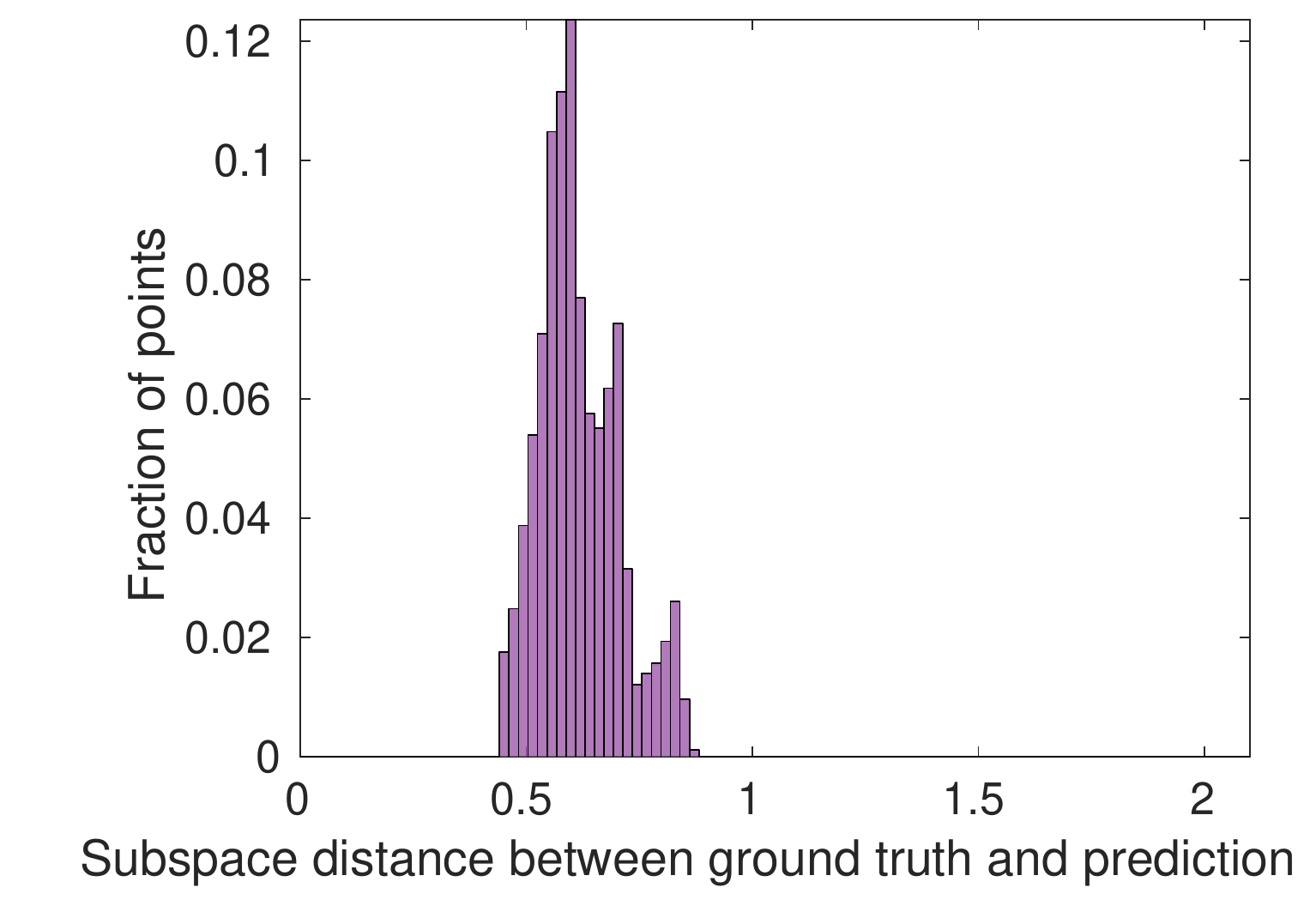}} 
         & \raisebox{-.5\height}{\includegraphics[trim = {2cm, 8cm, 4cm, 9cm}, clip, width=0.22\textwidth]{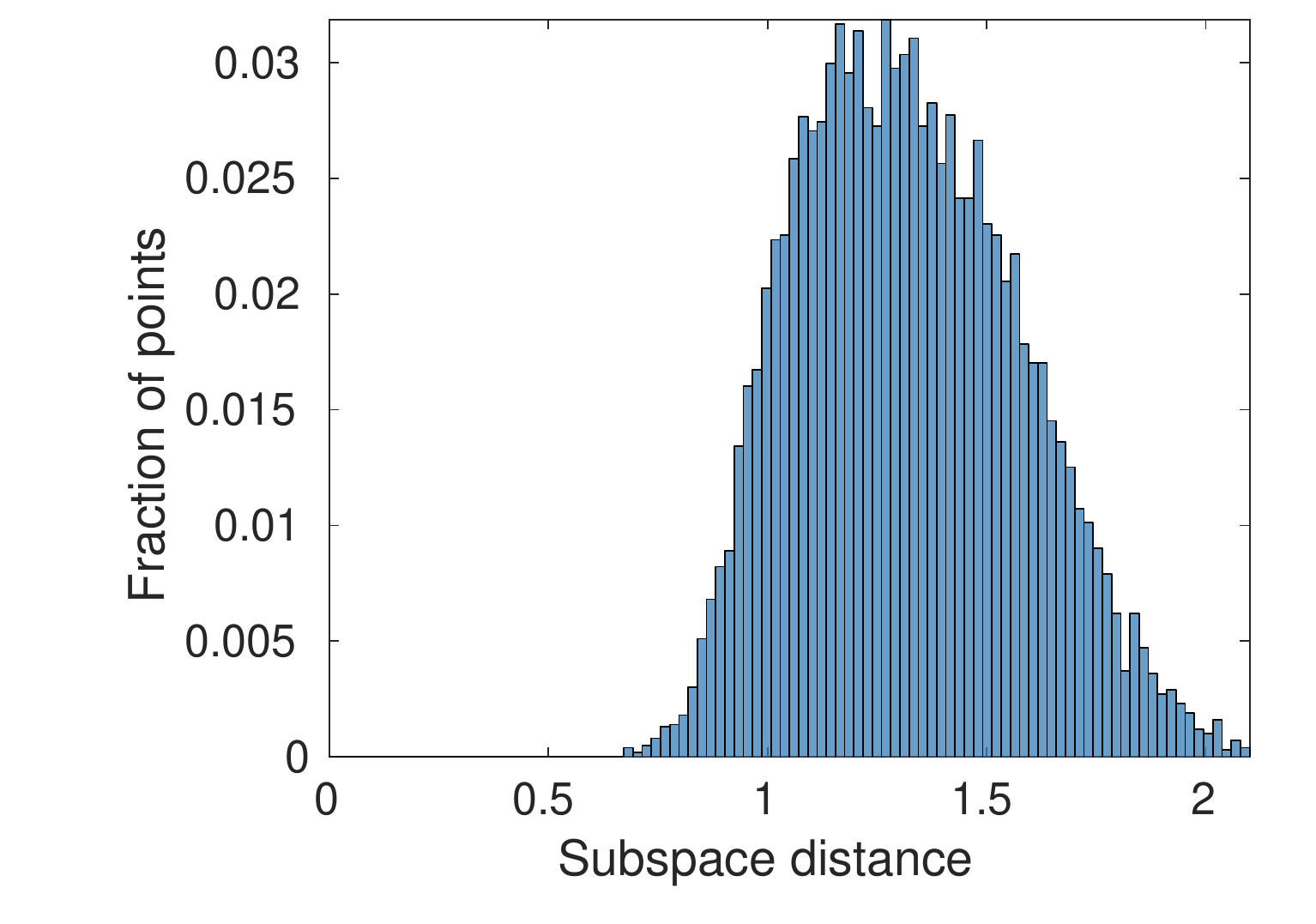}}\\

         \hline
    \end{tabular}
    \caption{Histograms of test results. From the plots, we can clearly observe that the GrassmannNet-TS (with the Fr\'{e}chet mean of the training set as the pole) framework performs much better than the baseline that attempts to regress directly to the PC's. The last column shows the histograms of subspace distances between training and test subspaces.}\label{table:f2is_histograms}
\end{table*}

{\small
\bibliographystyle{ieee}
\bibliography{main}
}

\end{document}